\documentclass{article}

\usepackage{microtype}
\usepackage{graphicx}
\usepackage{subcaption}
\usepackage{tcolorbox}
\usepackage{multirow}
\usepackage{booktabs} 
\usepackage{enumitem}
\usepackage[table]{xcolor}
\usepackage{titletoc}
\usepackage{tabularx}
\newcolumntype{Y}{>{\centering\arraybackslash}X}
\usepackage{hyperref}
\usepackage{listings}

\usepackage[accepted]{icml2026}

\usepackage{amsmath}
\usepackage{amssymb}
\usepackage{mathtools}
\usepackage{amsthm}
\tcbuselibrary{breakable} 

\usepackage[capitalize,noabbrev]{cleveref}

\theoremstyle{plain}

\theoremstyle{definition}

\theoremstyle{remark}

\usepackage{marvosym}
\usepackage[textsize=tiny]{todonotes}

\icmltitlerunning{Can Agents Generalize to the Open World? Unveiling the Fragility of Static Training in Tool Use}

\begin{document}

\twocolumn[ 
  \icmltitle{Can Agents Generalize to the Open World? \\Unveiling the Fragility of Static Training in Tool Use}



\icmlsetsymbol{equal}{*} 

\begin{icmlauthorlist}
\icmlauthor{Song-Lin Lv}{equal,yyy,sch} 
\icmlauthor{Weiming Wu}{equal,yyy}   
\icmlauthor{Rui Zhu}{yyy}
\icmlauthor{Zi-Jian Cheng}{yyy,sch}
\icmlauthor{Lan-Zhe Guo}{yyy,sch}
\end{icmlauthorlist}

\icmlaffiliation{yyy}{School of Intelligence Science and Technology, Nanjing University, Nanjing, China}
\icmlaffiliation{sch}{National Key Laboratory for Novel Software Technology, Nanjing University, Nanjing, China}

\icmlcorrespondingauthor{Lan-Zhe Guo}{guolz@lamda.nju.edu.cn}

  \icmlkeywords{Machine Learning, ICML}

  \vskip 0.3in
]

\printAffiliationsAndNotice{\icmlEqualContribution}

\begin{abstract}
While Large Language Model (LLM) agents demonstrate proficiency in static benchmarks, their deployment in real-world scenarios is hindered by the dynamic nature of user queries, tool sets, and interaction dynamics. 
To address this generalization gap, we formalize \textbf{OpenAgent} (Tool-Use Agent in Open-World), a problem setting characterized by distributional shifts across query, action, observation, and domain dimensions.
To systematically diagnose its impact, we  construct a controlled sandbox environment where we define fine-grained environmental shifts across a four-tier hierarchy, \emph{Perception}, \emph{Interaction}, \emph{Reasoning}, and \emph{Internalization}, and conduct a comprehensive series of experiments.
Our analysis yields a series of key insights, demonstrating that agents trained via both Supervised Fine-Tuning(SFT) and Reinforcement Learning suffer from varying degrees of performance degradation when confronting open environmental shifts.
Building on these insights, we propose Perturbation-Augmented Fine-Tuning, a disturbance-based intervention strategy for SFT that lays the foundation for enhancing agent robustness and utility in realistic environments. Our code will be released at: \url{https://github.com/LAMDA-NeSy/OpenAgent}.
\end{abstract}

\section{Introduction}

\begin{figure}
    \centering
    \includegraphics[width=\linewidth]{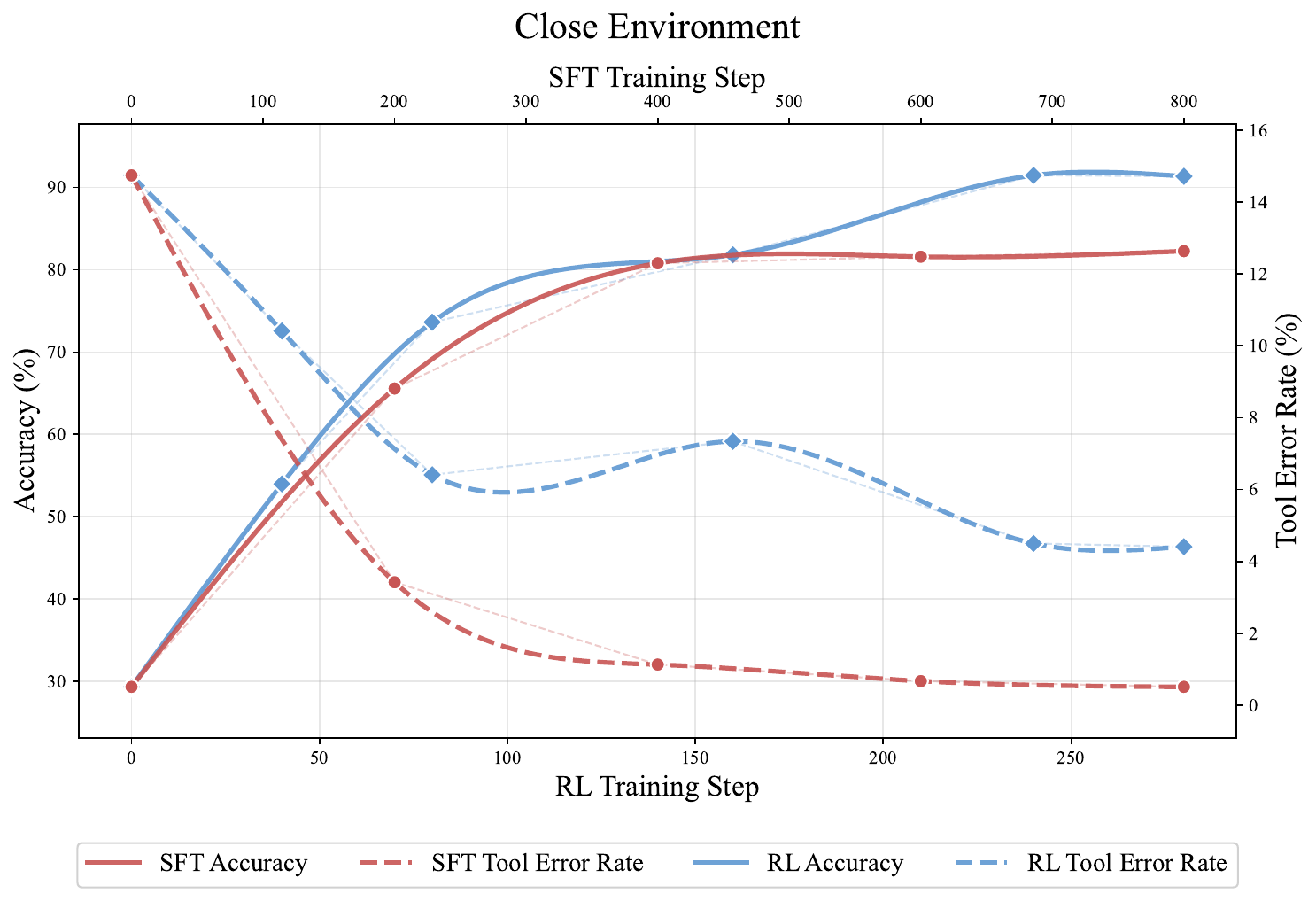}
    \caption{Performance evolution of SFT and RL agents in the closed setting. Both approaches show stable and continuous improvement throughout the training process.}
    \label{fig:closeset}
\end{figure}

The integration of Tool Learning and the Model Context Protocol (MCP)~\cite{anthropic2024mcp} has catalyzed a paradigm shift in Large Language Model (LLM) agents, enabling them to navigate external environments to solve complex, multi-step tasks~\cite{stabletoolbench,survey1,survey2}. Optimized via Supervised Fine-Tuning (SFT) and Reinforcement Learning (RL), recent open-source models~\cite{tool1, tool3,qwen2.5,qwen2.5vl} have achieved tool invocation proficiency that rivals proprietary frontiers like GPT-4~\cite{gpt-4} and Gemini~\cite{gemini}. As illustrated in Figure~\ref{fig:closeset}, under the prevailing \textit{static world assumption}, where the distribution of tools, schemas, and interaction logic remains consistent between training and inference~\cite{huggingfacegpt,toolvqa, gentool}, both SFT and RL paradigms demonstrate stable and continuous performance gains, eventually converging on near-perfect success rates.

However, this stability is often an artifact of the closed-set nature of current benchmarks. Real-world deployment is fundamentally non-stationary: APIs are deprecated, tool schemas evolve, and user instructions are frequently ambiguous. When these environmental dynamics diverge from training priors, the apparent mastery observed in static benchmarks often proves fragile. This discrepancy raises a fundamental research question: 
\textbf{\textit{Do current training paradigms enable agents to generalize to the open world?}}

To rigorously address this, we formally define \textbf{OpenAgent} (Tool-Use Agent in Open-World), a problem setting characterizing shifts across four dimensions: \textit{User Queries} ($\Delta \mathcal{Q}$), \textit{Tool Sets} ($\Delta \mathcal{A}$), \textit{Interaction Dynamics} ($\Delta \mathcal{O}$) and \textit{Task Domain} ($\Delta \mathcal{D}$). To isolate the impact of these shifts from the inherent instability and transient noise of real-world APIs, we establish a controlled sandbox environment. This setup allows for controlled probing, enabling us to maintain a pure closed-set baseline while systematically injecting open-world perturbations across a four-tier diagnostic framework: \textit{Perception}, \textit{Interaction}, \textit{Reasoning}, and \textit{Internalization}. 

Our comprehensive evaluations reveal varying degrees of generalization and adaptability in SFT and RL models under open-world settings, while further identifying their respective structural weaknesses. We find that \textbf{both} paradigms suffer from performance degradation in open environments, albeit through distinct failure modes. While SFT agents are prone to trajectory overfitting and brittle symbolic anchoring, RL agents, despite demonstrating better semantic grounding, remain vulnerable to boundary blindness due to a teleological bias in their reward structures.

Building on these insights, we introduce Perturbation-Augmented Fine-Tuning (PAFT), a data-centric intervention strategy for SFT. By injecting controlled observation anomalies and symbolic noise into training trajectories, PAFT decouples models from shallow pattern memorization and restores robust generalization in open worlds.

In summary, the primary threefold contributions of this work are as follows:
\begin{itemize}[leftmargin=*]
    \item \textbf{Formalization of Open Environment in Tool Use:} We formally define the OpenAgent setting, characterizing the shifts in queries, tool sets, interaction dynamics and task domain inherent to real-world deployments.
    \item \textbf{Systematic Evaluation in Open Environments:} We establish a controlled sandbox environment and conduct a comprehensive series of experiments to evaluate the robustness of SFT and RL agents across varying levels of environmental shifts.
    \item \textbf{Key Insights and Methods:} We identify critical failure modes in current paradigms, specifically symbolic anchoring in SFT and boundary blindness in RL, and propose PAFT to effectively mitigate performance degradation.
\end{itemize}
\paragraph{Conflict of Interest Disclosure.}
The authors declare no conflicts of interest.
\section{Related Works}

\begin{figure}[t]
    \centering
    \includegraphics[width=\linewidth]{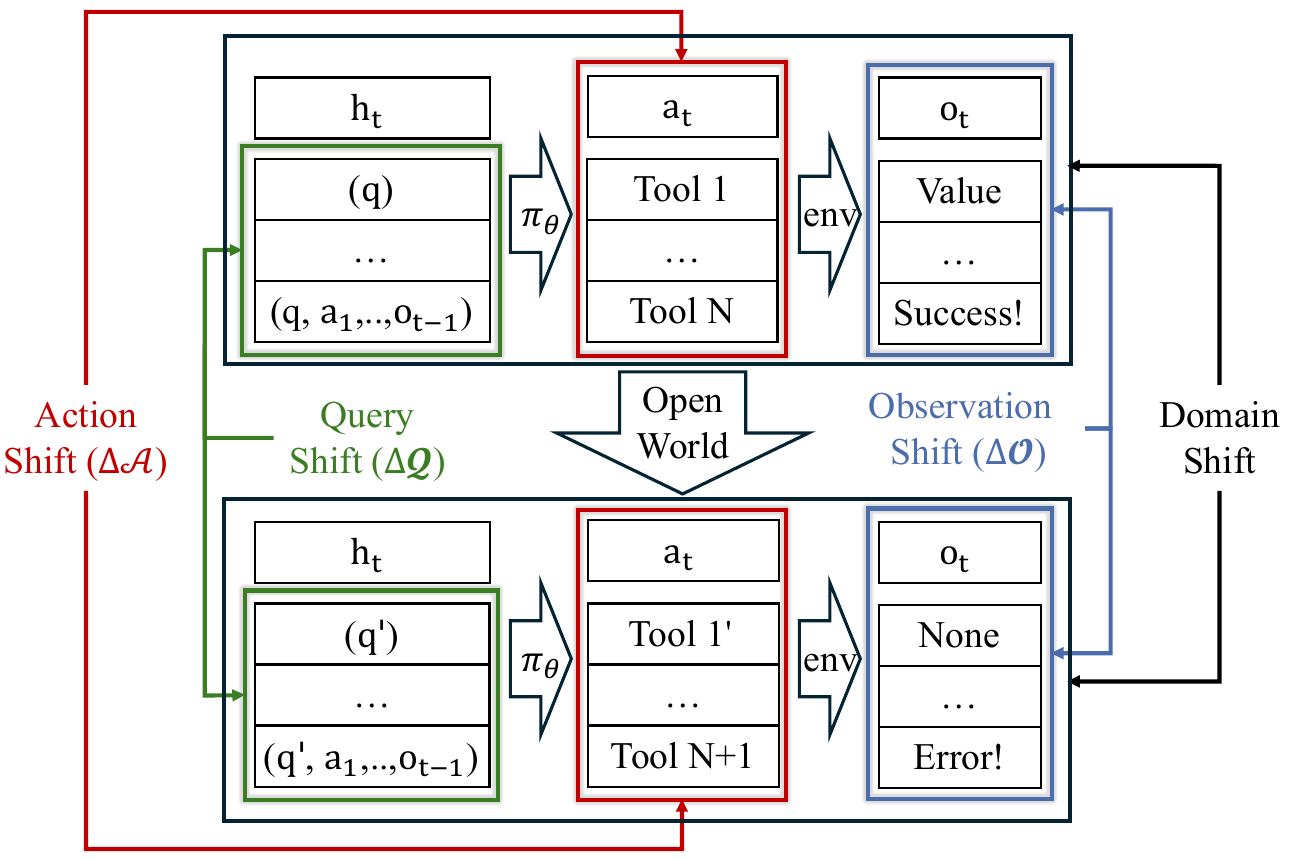}
    \caption{Challenges in OpenAgent setting, including query, action, observation and domain shifts.}
    \vspace{-0.5cm}
    \label{fig:opentool}
\end{figure}
\begin{figure*}
    \centering
    \includegraphics[width=\linewidth]{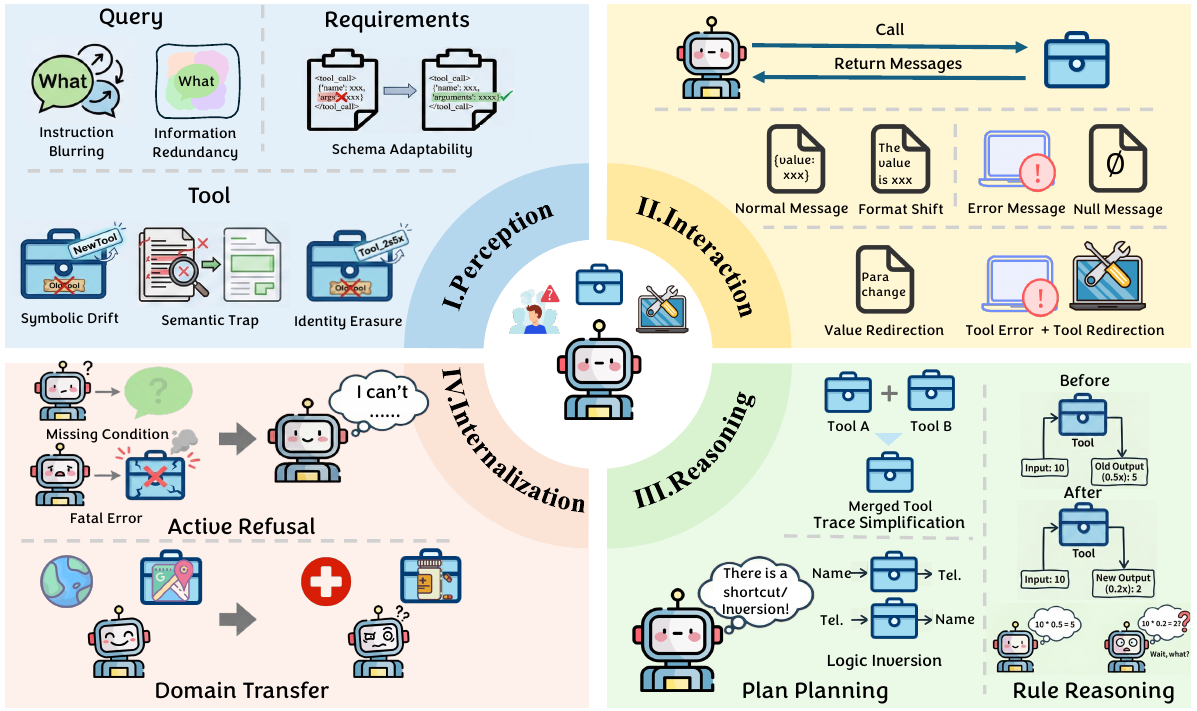}
    \caption{Architecture Diagram of the Evaluation Task. We partition this evaluation task architecture diagram into four levels from shallow to deep: Perception, Interaction, Reasoning, and Internalization.}
    \label{fig:placeholder}
\end{figure*}
\paragraph{Post-training Paradigms for Agents}
Current methodologies are categorized into three methods. SFT-based works like GenTool~\cite{gentool}, TOUCAN~\cite{toucan}, and others~\cite{tool6,gentool,tool5,hammer,mo2026open} standardize tool invocation via token prediction. Conversely, RL frameworks such as ToolRL~\cite{toolrl}, DeepEyes~\cite{deepeyes}, and others~\cite{pvpo,dapo,toolrl} utilize reward mechanisms to drive robust decision-making. Hybrid paradigms including ReTool~\cite{retool}, DeepEyesV2~\cite{deepeyesv2}, and others~\cite{openthinkimg,li2026flow} combine SFT initialization with RL exploration. Following recent closed-source advances models like Deepseek-V3.2, GLM-4.5~\cite{glm45}, and Kimi K2~\cite{kimik2}, we select full-parameter SFT and RL as baselines. 
\vspace{-10pt}
\paragraph{Evaluation Paradigms for Agents}
Agent capabilities are evaluated via local tools~\cite{gta,stabletoolbench,metatool,tooleyes} and remote protocols~\cite{apibank,benchmark3,mcpradar,mcpbench}. Existing benchmarks mainly prioritize task success~\cite{gaia,mcpeval,bfcl,shao2026chinatravel}, while robustness studies focus on inference-time perturbations~\cite{rotbench,butterfly}. Recently, agent benchmarks further move toward realistic  environments~\cite{agentbench,webarena,workarena,swebench,visualwebarena,appworld,agentboard,osworld,theagentcompany}. Unlike open-environment learning for passive models~\citep{guo2020safessl,li2021safewsl,openml_zhou,guo2022classimb,jia2024realisticssl,guo2025robustssl,tabfsbench}, OpenAgent studies shifts in multi-step. While prior work~\cite{SFTRL} shows RL's advantage over SFT, it lacks tool-use contexts; we address this gap by analyzing training-to-testing transfer under open-environment shifts.
\section{Preliminaries}
\label{sec:prelims}

\subsection{Problem Formulation}

We model the agent as a sequential decision-making process
$\mathcal{M} = \langle \mathcal{Q}, \mathcal{A}, \mathcal{O}, \pi \rangle$,
comprising user queries $\mathcal{Q}$, tool space $\mathcal{A}$, and
observation space $\mathcal{O}$. Given a query $q \in \mathcal{Q}$, at each
step $t$, the agent uses policy $\pi_\theta(a_t \mid h_t)$ to select action
$a_t \in \mathcal{A}$ based on history
$h_t = (q, a_1, o_1, \dots, a_{t-1}, o_{t-1})$. The environment returns
observation $o_t \in \mathcal{O}$, which updates the history to
$h_{t+1}=(h_t,a_t,o_t)$ until task completion. Our core challenge is the
generalization of the model when test-time distributions diverge
from training priors.

\subsection{The OpenAgent Setting: Distributional Shifts}

Unlike distributional shifts in passive models, where each input is perturbed
independently, tool-use agents face \emph{cascading} shifts along the
agent-environment loop. A perturbation at step $t$ may alter observation $o_t$,
which enters the history $h_{t'}$ and conditions all subsequent decisions
$\pi_\theta(a_{t'} \mid h_{t'})$ for $t'>t$, causing errors to compound along
the trajectory. As illustrated in Figure~\ref{fig:opentool}, OpenAgent covers
query shifts ($q \rightarrow q'$), action-space shifts including perturbed
existing tools and newly introduced tools, observation shifts such as \textit{None} or \textit{Error}, and
joint domain shifts where all interaction elements change together. We
decompose this interactive shift space into four dimensions.

\textbf{Intent Shift in Query Space ($\Delta \mathcal{Q}$).}
The intent distribution $P(q)$ shifts across both semantic directives and
linguistic surface forms. Unlike passive input perturbations,
query misinterpretation in tool-use agents propagates through the trajectory:
an initial misparse may seed a misaligned tool chain whose errors compound
across subsequent steps.

\textbf{Structural Shift in Action Space ($\Delta \mathcal{A}$).}
The tool space becomes non-stationary
($\mathcal{A}_{\mathrm{test}} \neq \mathcal{A}_{\mathrm{train}}$), spanning
multiple levels of change: \emph{surface drift}, which renames identifiers while
preserving functionality; \emph{semantic conflict}, which retains familiar
names but alters documented behavior; and \emph{structural reconfiguration},
which merges, splits, or inverts tool dependencies. The agent must distinguish
cosmetic changes from fundamental reconfigurations and ground actions in the
current tool semantics.

\textbf{Dynamics Shift in Observation Space ($\Delta \mathcal{O}$).}
The feedback channel becomes non-stationary: test-time observations may include
return formats, error states, null values, or corrective instructions that are
absent or rare in training. Unlike static outputs, these observations directly
update the agent's belief state and reshape the subsequent action space.
Crucially, the agent must discriminate \emph{informative} anomalies that require
policy adjustment from \emph{terminal} failures that require task refusal, a
distinction absent in passive settings where each input simply calls for a
prediction.

\textbf{Compositional Domain Shift ($\Delta \mathcal{D}$).}
This shift jointly changes $\mathcal{Q}$, $\mathcal{A}$, and $\mathcal{O}$ into
a new domain while preserving a latent problem-solving structure
$\mathcal{G}$. Despite distinct surface semantics and interfaces, the agent
must transfer the underlying reasoning topology by re-grounding its nodes and
dependencies in the target domain, rather than relying on domain-specific
lexical or procedural patterns.
\section{Evaluation Tasks}
\label{sec:experiments}

\subsection{Experimental Setup}

\paragraph{Controlled Sandbox Environment.}
We utilize a synthetic sandbox environment to enable controlled probing of agent capabilities. Unlike real-world APIs which introduce uncontrollable transient noise, a sandbox allows us to precisely isolate environmental variables and establish a pure closed environment baseline to measure open environment shifts. We synthesized a diverse set of tool-use tasks focused on simple geographical POI queries and calculation problems to test reasoning boundaries, the full specifications of which are detailed in the Appendix~\ref{app:sandbox}.

\vspace{-10pt}
\paragraph{Datasets and Protocols.}
The dataset comprises 6,050 training samples and 880 evaluation samples. To ensure rigorous evaluation, we enforced a strict separation between the datasets where no variation patterns from the test set appear during the training phase. This guarantees that the reported performance reflects genuine robustness rather than information leakage or pattern memorization.
\vspace{-10pt}
\paragraph{Training Implementation.}
We employ Qwen2.5-7B-Instruct~\cite{qwen2.5} as the backbone model. The SFT stage uses full-parameter fine-tuning, while the RL stage adopts the GRPO framework with sparse answers and format rewards. To further ensure a fair comparison with SFT, the RL environment provides only basic error codes upon failure without detailed guidance. We trained both models under comparable computational budgets until they achieved convergence in the closed environment as illustrated in Figure \ref{fig:closeset}. The Training details are provided in Appendix~\ref{app:setup}.

\textbf{Metrics.}
We employ four metrics to assess performance across correctness, efficiency, robustness, and adaptability:
\begin{itemize}[leftmargin=*]
    \setlength{\itemsep}{2pt}
    \item \textbf{Tool Error Rate (TER):} Measures \textit{robustness} via the ratio of invalid invocations to total calls ($TER = N_{error}/N_{total}$). A lower TER indicates precise adherence.

    \item \textbf{Active Exploration Score (AES):} Quantifies \textit{adaptability} by penalizing ``cognitive inertia'' when tools fail.
    \begin{equation}
        AES = e^{-(n_{old}-1)}
    \end{equation}
    where $n_{old}$ is the count of repeated invalid attempts. A score of $1.0$ signifies immediate adaptation.

    \item \textbf{Average Tool Chain Length (ATCL):} Gauges \textit{efficiency} by averaging tool invocations on solvable tasks. Lower values reflect concise reasoning topologies.

    \item \textbf{Refusal Rate (RR):} Evaluates \textit{boundary awareness} in unsolvable scenarios by measuring the ratio of correctly identified unsolvable cases ($RR = N_{refusal}/N_{unsolvable}$). A higher RR indicates superior active boundary detection.
\end{itemize}

\subsection{Tier 1: Perception Generalization}
\label{sec:tier1_perception}

In the context of agentic interaction, perception acts as the cognitive gateway grounding raw observations into actionable representations. To successfully initiate a task, an agent must traverse three representational alignments: deciphering unstructured user intents, aligning with rigid interface protocols, and comprehending functional tool definitions. We systematically evaluate the generalization boundaries of these capabilities through the following three dimensions.

\textbf{Instruction Robustness ($\Delta \mathcal{Q}$).
}Real-world user instructions are rarely canonical. We evaluate whether the agent can extract invariant intents from non-standard inputs by introducing two distinct types of variations.
\begin{itemize}[nosep, leftmargin=*]
    \item \textbf{Instruction Blurring:} We simulate scenarios where user intent is obfuscated by syntactic variance (e.g., colloquialisms, rhetorical questions) or semantic ambiguity (e.g., implicit references, omitted entities). This tests the model's ability to reconstruct exact intents without relying on explicit keyword matching.
    \item \textbf{Information Redundancy:} We test the robustness of the model's attention mechanism against information overload. By embedding the core instruction within irrelevant context or verbose background details, we evaluate the agent's capability to filter out red herrings and lock onto functional constraints.
\end{itemize}

\textbf{Schema Adaptability ($\Delta \mathcal{A}$).}
Agents often overfit to specific API formats seen during training. In this experiment, we introduce protocol shifts by modifying the reserved keywords in the tool invocation schema. For instance, we alter the parameter dictionary key from the standard \texttt{arguments} to variants like \texttt{args} or \texttt{parameters}. This verifies whether the model comprehends the abstract concept of parameter passing or merely performs rigid template completion based on memorized structural tokens.

\textbf{Semantic Grounding ($\Delta \mathcal{A}$).}
Current agents often rely on Symbolic Anchoring with rigid dependence on memorized identifiers rather than true understanding of functional definitions. To decouple these factors, we investigate whether models can adapt to dynamic documentation shifts through a Drift-Trap-Erasure trilogy:
\begin{itemize}[nosep, leftmargin=*]
    \item \textbf{Symbolic Drift:} We test tolerance to  perturbations via two sub-variants: \textit{Noise Injection} (appending non-semantic suffixes) and \textit{Synonymous Rewriting} (using LLMs to paraphrase). This verifies generalization across semantically equivalent but symbolically distinct labels.
    \item \textbf{Semantic Trap:} We engineer a conflict between symbol and semantics with adversarial attack.by retaining a familiar tool name but modify its description to explicitly forbid usage. If the agent invokes it, it proves that symbolic overrides semantic understanding.
    \item \textbf{Identity Erasure:} We perform the ultimate test of grounding by replacing meaningful tool names with random identifiers (e.g., Tool\_2A3D). This removes symbolic shortcuts, forcing the agent to rely exclusively on reading comprehension of the documentation to solve the task.
\end{itemize}
\subsection{Tier 2: Interaction Generalization}
\label{sec:tier2_interaction}

Interaction generalization evaluates whether an agent can use environmental
feedback to refine subsequent actions. In open environments, observations may
indicate format changes, execution failures, or new operational constraints
rather than merely confirming previous tool calls. We examine three capabilities:
\textit{Format Adaptability} for parsing diverse observation structures,
\textit{Error Correction} for recovering from explicit or silent failures, and
\textit{Dynamic Alignment} for adjusting actions according to feedback-provided
rules.

\paragraph{Format Adaptability ($\Delta \mathcal{O}$).}
Agents typically overfit to standard JSON formats for parsing observations. This experiment tests the fundamental capability of the agent to parse non-standard feedback. We introduce Format Shift by altering the structure of tool return values from structured JSON to unstructured formats such as Markdown tables or raw text logs. This evaluates whether the agent extracts key information based on semantic content rather than relying on rigid syntactic parsing templates.

\paragraph{Error Correction ($\Delta \mathcal{O}$).}
Open environments may expose agents to execution failures that interrupt the
planned trajectory. We evaluate whether the agent can diagnose such
failures and recover by revising its invocation plan or selecting alternative
tools. We consider two types of failure:
\begin{itemize}[nosep, leftmargin=*]
    \item \textbf{Error Return:} The primary tool explicitly returns an error
    code or traceback message after invocation. This tests whether the agent can
    interpret explicit failure signals and switch to a redundant tool to
    resolve the blockage.
    \item \textbf{Null Return:} The primary tool returns a null value or an
    empty string without any warning. This tests whether the agent can detect
    an implicit anomaly where the return is syntactically valid but
    semantically uninformative, and infer that corrective action is required.
\end{itemize}
\paragraph{Dynamic Alignment ($\Delta \mathcal{O}$).}
\vspace{-7pt}
Real-world environments are non-stationary and may provide corrective
instructions within feedback to steer the agent away from obsolete assumptions.
We evaluate whether the agent can perform feedback-conditioned policy
adjustment through two active guidance mechanisms:
\begin{itemize}[nosep, leftmargin=*]
    \item \textbf{Value Redirection:} The environment notifies the agent that the valid parameter logic has shifted (e.g., ``ID 1024 is deprecated, use ID 2048''). We verify if the agent can grasp this dynamic constraint and reissue the tool call with corrected values.
    \item \textbf{Tool Redirection:} The environment signals that a specific tool schema is obsolete and explicitly suggests an alternative one. This tests whether the agent can abandon its pre-planned trajectory and strictly adhere to the new operational manual provided in the feedback.
\end{itemize}
\subsection{Tier 3: Reasoning Generalization}
\label{sec:tier3_reasoning}

Beyond perception and interaction, \textit{Reasoning} determines whether an
agent can compose individual tool calls into a valid execution plan. Instead
of triggering APIs sequentially according to memorized trajectories, a robust
agent should infer the current rules governing tool behavior and construct an
execution graph that satisfies the task dependencies. We evaluate this ability
through two complementary dimensions: \textit{Rule Reasoning}, which tests
whether the agent can adapt to changed functional definitions, and
\textit{Path Planning}, which tests whether it can reorganize tool dependencies
under altered execution structures.

\paragraph{Rule Reasoning ($\Delta \mathcal{A}$).}
Models may memorize training-time rules instead of
performing deduction from the current tool description. We introduce a
\textit{Calculation Shift} by modifying the functional semantics of selected
tools while keeping their interfaces unchanged. For example, a tool originally
defined to return a value multiplied by $0.5$ is redefined in the test setting
to use a factor of $0.1$. This experiment verifies whether the agent can
derive the correct result from the updated rule description, or instead
reverts to the memorized rule from the training distribution.

\paragraph{Path Planning ($\Delta \mathcal{A}$).}
This dimension evaluates whether the agent can construct valid execution
graphs under changed tool dependencies. We test whether it genuinely reasons
over the current causal structure among tools, rather than overfitting to fixed
tool chains observed during training, through two complementary
sub-experiments:
\begin{itemize}[nosep, leftmargin=*]
    \item \textbf{Trace Simplification:} We test whether the agent can improve
    execution efficiency when a shortcut becomes available. Specifically, we
    introduce a composite tool that functionally replaces a multi-step tool
    chain used during training. The agent should recognize the new affordance
    and select the efficient composite tool, rather than rigidly following the
    previously memorized multi-step trajectory.

    \item \textbf{Logic Inversion:} We test robustness to causal reversal in
    the tool dependency graph. At test time, we invert the documented
    information-acquisition order, for example by making the output of tool
    $B$ a prerequisite for invoking tool $A$. This verifies whether the agent
    can resolve preconditions from the current dependency graph, or instead
    adheres to the fixed topological order learned from training trajectories.
\end{itemize}

\subsection{Tier 4: Internalization Generalization}
\label{sec:tier4_internalization}

Internalization generalization evaluates whether an agent has learned
task-level principles that go beyond local tool invocation patterns. At this
level, a robust agent should not only execute valid tool-use trajectories, but
also recognize when a task is infeasible under the current constraints and
transfer abstract problem-solving structures across domains. We examine this
through two capabilities: \textit{Active Refusal}, which tests solvability
boundary recognition, and \textit{Domain Transfer}, which tests whether
learned reasoning topologies can be reused in semantically different
environments.
\paragraph{Active Refusal ($\Delta \mathcal{Q} \cup \Delta \mathcal{O}$).}
A robust agent should determine whether a task is solvable given the available
information and executable tools, rather than forcing a plausible but
unsupported answer. We evaluate this boundary-awareness ability through two
insufficient-condition scenarios:
\begin{itemize}[nosep, leftmargin=*]
    \item \textbf{Missing Condition:} We remove necessary conditions from the
    user query through a human-verified process, making the task theoretically
    unsolvable due to information deficiency. This tests whether the agent can
    detect the logical incompleteness of the request and refuse to answer,
    instead of fabricating missing information to force completion.

    \item \textbf{Fatal Error:} We simulate an unrecoverable
    failure where a necessary tool returns a fatal error and no alternative
    tool is available. This tests whether the agent can recognize the resulting
    operational deadlock and conclude that the task is currently unexecutable,
    rather than repeatedly retrying or hallucinating a successful result.
\end{itemize}
\vspace{-7pt}
\paragraph{Domain Transfer ($\Delta \mathcal{D}$).}
This dimension evaluates whether the agent can transfer internalized
problem-solving structure across domains. We introduce a
\textit{Holistic Domain Shift}, where user queries, tool definitions, and
environmental observations change jointly. For example, we migrate the testing
environment from an address-query scenario to a medical-registration scenario.
Although the surface semantics and interfaces change substantially, the
underlying causal structure is preserved as a latent structural invariance.
This experiment tests whether the agent can re-ground the same abstract
reasoning topology, such as a search-select-confirm chain, in a novel target
domain rather than overfitting to domain-specific lexical or procedural cues.

\section{Key Observations}
\label{sec:observations}
\begin{figure*}[h]
    \centering
    \includegraphics[width=0.96\linewidth]{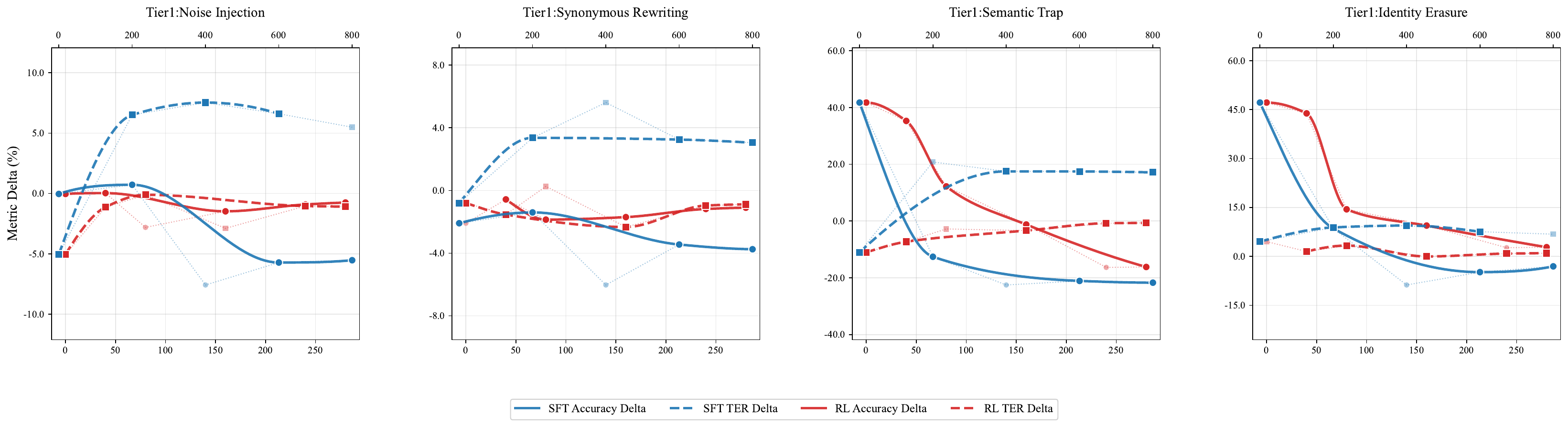}
   \caption{Accuracy Delta and TER in Tier-1 Perception. SFT exhibits brittle symbolic anchoring and underperforms compared to RL when tool semantics shift. For detailed case analyses, refer to Appendix \ref{case_study}. Delta represents the performance gap relative to the closed-set baseline. Setup and absolute values are provided in Appendix \ref{implementation} and \ref{detailed_results}.}
    \label{fig:tier1-delta}
\end{figure*}
\begin{figure*}[h]
    \centering
    \includegraphics[width=1\linewidth]{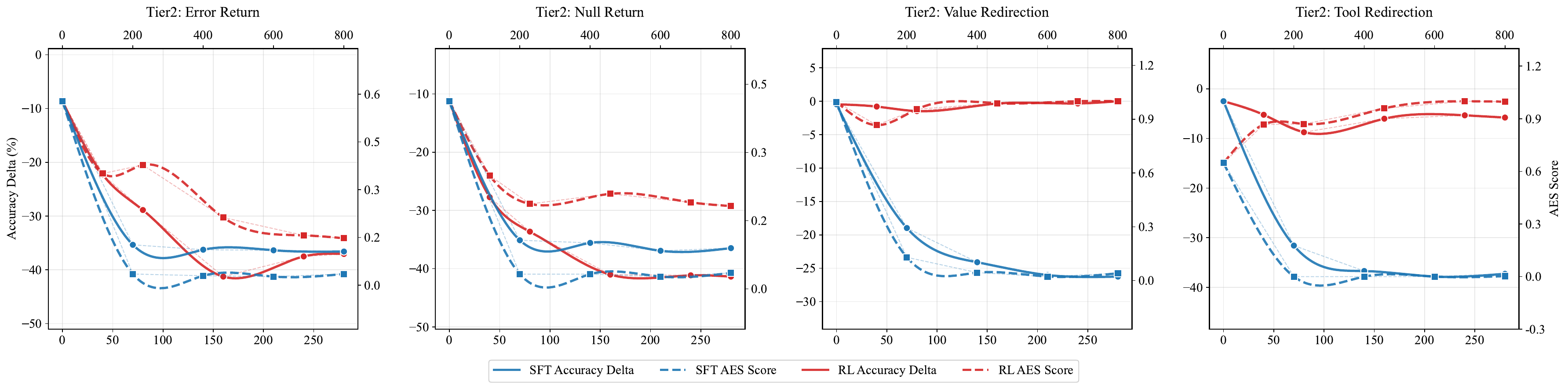}
\caption{Accuracy Delta and AES Score in Tier-2 Interaction. Both RL and SFT degrade under ambiguous feedback, but RL maintains superior resilience under explicit guidance while SFT fails to adapt. Delta represents the performance gap relative to the closed-set baseline. Setup and absolute values are provided in Appendix \ref{implementation} and \ref{detailed_results}.}
    \label{fig:tier2-delta}
\end{figure*}

Synthesizing the results across the four tiers of generalization, we distill observations regarding the behavioral divergence between SFT and RL in open-world agentic tasks.

\begin{tcolorbox}[
  colback={rgb,255:red,230; green,240; blue,250},
  colframe={rgb,255:red,42; green,110; blue,187},
  boxrule=1pt,
  arc=3mm,
  left=6pt, right=6pt, top=4pt, bottom=4pt,
  title={\small Tier 1: Tool Perception}
]
RL achieves semantic grounding through interaction while SFT relies on brittle symbolic anchoring that degrades as fitting increases.
\end{tcolorbox}

\begin{figure*}[h]
    \centering
    \includegraphics[width=1\linewidth]{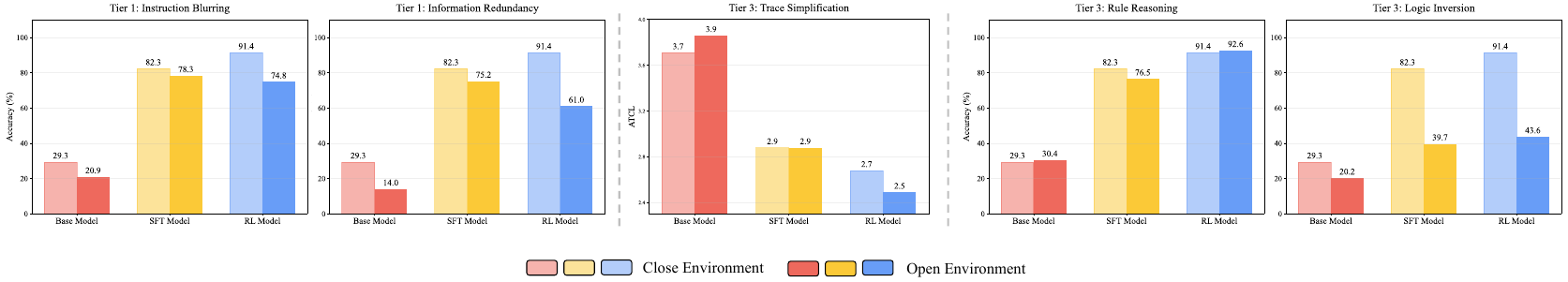}
\caption{ Accuracy comparison on Tier 1-Instruction Robustness (left) and Tier 3-Rule Reasoning \& Path Planning (right), evaluated at the stable training phase. While SFT models exhibit degradation across all perturbations, RL models show drops primarily under logic inversion and query variations. Detailed setups and full training dynamics are provided in Appendices \ref{implementation} and \ref{detailed_results}.}
    \label{fig:tier3-delta}
\end{figure*}

As illustrated in Figure \ref{fig:tier1-delta}, we observe a consistent degradation pattern where robustness declines as training steps increase. Specifically, in the Semantic Trap and Identity Erasure experiments, the SFT model displays a sharp increase in TER accompanied by a significant drop in accuracy. The dashed blue line indicates that SFT frequently invokes the wrong tools when surface-level names are perturbed. Conversely, while RL also experiences some performance decay due to overfitting, it maintains a consistently lower error rate and higher accuracy retention compared to SFT. The RL agent demonstrates superior resistance to adversarial descriptions and effectively ignores misleading name cues.

This disparity stems from SFT's reliance on \textit{symbolic anchoring}, where the model acts as a superficial pattern matcher that collapses once surface tokens are decoupled from their semantics. In contrast, the penalty-driven nature of RL necessitates \textit{semantic grounding}, compelling the agent to scrutinize functional descriptions rather than memorizing rigid name-context associations.

\begin{tcolorbox}[
  colback={rgb,255:red,230; green,240; blue,250},
  colframe={rgb,255:red,42; green,110; blue,187},
  boxrule=1pt,
  arc=3mm,
  left=6pt, right=6pt, top=4pt, bottom=4pt,
  title={\small Tier 2: Interaction with Guidance}
]
RL leverages explicit guidance for dynamic policy adaptation, while SFT exhibits trajectory inertia and often hallucinates outcomes under corrective feedback.
\end{tcolorbox}

As illustrated in Figure \ref{fig:tier2-delta}, we observe a distinct behavioral divergence dependent on feedback clarity. Under ambiguous signals like null returns, both models exhibit performance degradation. Crucially, this failure mode follows a clear trend where the inability to adapt intensifies as training progresses. As the models become more fitted to the training distribution, their capacity to actively explore under uncertainty progressively declines. A significant disparity emerges under explicit tool redirection. Qualitative case studies reveal that SFT tends to interpret observations as successful confirmations, even when the feedback explicitly redirects the agent. Consequently, it ignores the guidance and hallucinates successful observations to force the continuation of its pre-planned trajectory. In contrast, RL maintains high adaptability by parsing the instruction and migrating to the new tool immediately, and then continues reasoning.

We hypothesize that this behavioral divergence stems from differences in training paradigms. SFT is optimized on clean demonstrations where observations consistently validate actions. This likely conditions the model to disregard the causal link between feedback content and the subsequent policy, thereby making its behavior resemble an open-loop execution policy. Conversely, RL encourages a more closed-loop behavior because ignoring error signals is penalized during interaction. This interaction mechanism potentially forces the agent to treat environmental observations as critical state updates, ensuring that explicit environmental instructions are prioritized to maximize reward.

\begin{tcolorbox}[
  colback={rgb,255:red,230; green,240; blue,250},
  colframe={rgb,255:red,42; green,110; blue,187},
  boxrule=1pt,
  arc=3mm,
  left=6pt, right=6pt, top=4pt, bottom=4pt,
  title={\small Tier 3: Reasoning Generalization}
]
RL demonstrates generalization in local adaptation, yet both paradigms suffer substantial degradation under global dependency inversion.
\end{tcolorbox}

As illustrated in Figure \ref{fig:tier3-delta}, RL exhibits superior generalization in local optimization tasks. In \textit{Trace Simplification}, the RL model achieves the lowest ATCL, indicating it successfully abandons the memorized training-time sequence to utilize efficient composite tools. Similarly, in \textit{Rule Reasoning}, RL outperforms SFT by dynamically adhering to updated calculation logic, whereas SFT shows a stronger tendency to regress to memorized training constants.

However, a critical shared failure emerges in \textit{Logic Inversion}. When the causal dependency between tools is reversed (e.g., shifting from $A \rightarrow B$ to $B \rightarrow A$), both SFT and RL suffer precipitous performance drops. Qualitative analysis in Appendix \ref{case_study} reveals that despite the documentation explicitly stating the new prerequisite order, both models frequently hallucinate the existence of the training-set topology. This suggests that while RL generalizes well to local rule shifts, both paradigms overfit to the temporal order of action sequences, treating the dependency graph as a fixed action order rather than a dynamically specified logical structure.

\begin{tcolorbox}[
  colback={rgb,255:red,230; green,240; blue,250},
  colframe={rgb,255:red,42; green,110; blue,187},
  boxrule=1pt,
  arc=3mm,
  left=6pt, right=6pt, top=4pt, bottom=4pt,
  title={\small Tier 4: Boundary Awareness}
]
Both paradigms struggle with boundary awareness in unsolvable states, but exhibit distinct failure patterns: SFT often fails to perceive fatal feedback, whereas RL perceives the failure but still favors forced completion.
\end{tcolorbox}
 
\begin{figure*}[h]
    \centering
    \includegraphics[width=1\linewidth]{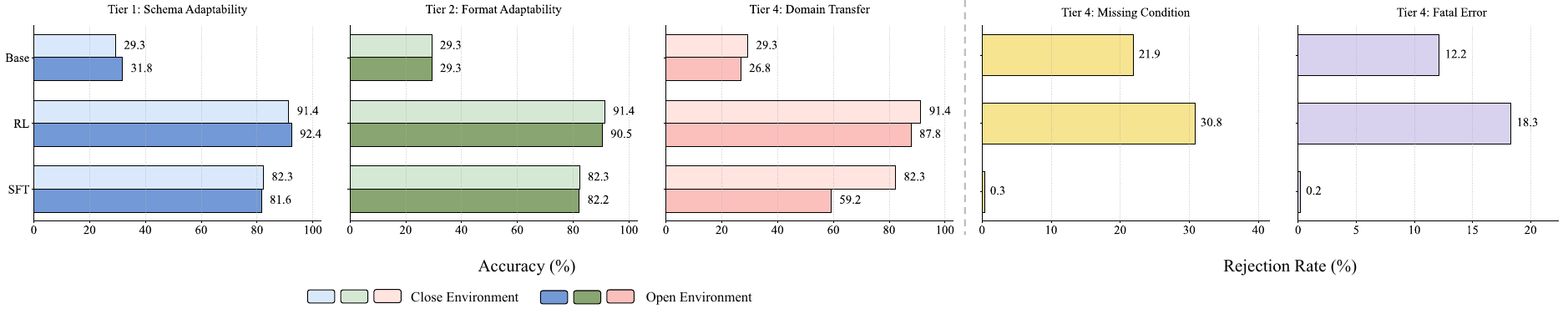}
\caption{Accuracy in Tier 1-Schema Adaptability, Tier 2-Format Adaptability, and Tier 4-Domain Transfer (left) and Refusal Rate in Tier 4-Active Refusal (right), evaluated at the stable training phase. Both SFT and RL models are robust to simple format changes, whereas SFT shows significant degradation under domain transfer. Both methods demonstrate limited active refusal for unsolvable queries. Detailed setups and full training dynamics are provided in Appendices \ref{implementation} and \ref{detailed_results}.}
\label{fig:tier4-delta}
\end{figure*}
 
As illustrated in Figure \ref{fig:tier4-delta}, both models struggle to identify unsolvable queries, leading to low refusal success rates when tools are unavailable. However, a fine-grained analysis of non-refusal cases under \textit{Fatal Error} scenarios reveals that SFT and RL fail through qualitatively distinct mechanisms. Case studies in Appendix~\ref{app:Tier4} show that upon receiving an error, SFT directly outputs an unsupported answer without acknowledging the error at all, effectively treating the error observation as if it were a normal return value. In contrast, RL explicitly recognizes the failure (e.g., ``It seems the tool is not available''), demonstrating a notably high error awareness rate, yet it still proceeds to fabricate an answer rather than issuing a refusal.
 
This divergence is consistent with the open-loop versus closed-loop dichotomy observed in earlier tiers. SFT's failure is perceptual: its open-loop policy does not parse environmental feedback, so the error signal is simply ignored. RL's failure, however, is more decisional: it perceives the error but still favors completion, which is consistent with a \textit{teleological bias} induced by outcome-oriented rewards. Because training rewards presuppose that every problem is solvable, the policy learns to provide an answer at all costs, prioritizing completion over honest refusal. This is distinct from mere overconfidence; it reflects a structural limitation of outcome-based reward design that lacks an explicit incentive to abstain when the task is infeasible. Such persistent forced-completion behavior raises critical agent safety concerns, as models may sacrifice truthfulness for helpfulness in unpredictable open-world deployments.
 
\vspace{-5pt}
\section{Perturbation-Augmented Fine-Tuning}
\label{sec:method}

Our analysis reveals that SFT's generalization decay stems from a distributional mismatch: standard SFT optimizes $\mathcal{L} = -\mathbb{E}_{(h,a) \sim \mathcal{D}_{\text{clean}}}[\log \pi_\theta(a|h)]$, where all training histories $h$ are drawn exclusively from successful executions. This confines the learned policy to a narrow manifold $\mathcal{H}_{\text{clean}}$ that excludes error states, ambiguous feedback, and symbolic variations. To bridge this gap, we propose \textbf{Perturbation-Augmented Fine-Tuning (PAFT)}, a data intervention strategy that augments $\mathcal{D}_{\text{clean}}$ with controlled trajectory-level perturbations:
\begin{equation}
    \mathcal{D}_{\text{PAFT}} = (1-\alpha)\,\mathcal{D}_{\text{clean}} \;\cup\; \alpha \cdot \mathcal{G}(\mathcal{D}_{\text{clean}})
\end{equation}
where $\mathcal{G}$ applies perturbation operators that expand the support of $h$ into regions containing anomalous observations, unsolvable states, and symbolic variants. Unlike standard augmentation that perturbs inputs in isolation, PAFT operates at the trajectory level, injecting perturbations mid-trajectory and supervising the corrective continuations. 

\subsection{Method}

\textbf{Environmental Feedback Perturbation (EFP).}
To combat the open-loop trajectory inertia, we extend successful sub-trajectories by injecting stochastic observation anomalies that simulate unexpected environmental feedback. A standard chain is transformed via a generator $\mathcal{G}_{env}$ as:
\begin{equation}
    \tau_{\text{orig}} = \{a_i, o_i\} \xrightarrow{\mathcal{G}_{env}} \tau' = \{a_i, o_{\text{change}}, a'_i, o_i\}
\end{equation}
Here, $o_{\text{change}}$ represents a feedback anomaly (e.g., tool redirection or value deprecation). The model is supervised to generate a corrective action $a'_i$ that recovers from the disturbance, rather than blindly assuming success. This converts the open-loop pattern into a closed-loop recovery behavior.

\textbf{Solvability Boundary Perturbation (SBP).}
To address the boundary blindness identified in Tier~4, we construct negative samples that enforce explicit refusal behavior. A refusal generator injects unrecoverable errors into the observation, modifying the trajectory to:
\begin{equation}
    \tau_{\text{orig}} = \{a_i, o_i, a_{i+1}\} \xrightarrow{\mathcal{G}_{\text{refusal}}} \tau' = \{a_i, o_{\text{fatal}}, a_{\text{end}}\}
\end{equation}
where $o_{\text{fatal}}$ signifies a fatal error with no alternative tools available, and $a_{\text{end}}$ represents the explicit refusal action. This breaks the implicit ``every problem is solvable'' assumption in $\mathcal{D}_{\text{clean}}$ and teaches the model to halt execution rather than hallucinate.

\textbf{Symbolic Representation Perturbation (SRP).}
To decouple the symbolic anchoring identified in Tier~1, we apply surface-level augmentations to the tool definition space. An auxiliary agent performs synonymous rewriting and noise injection on tool names and documentation content. This forces the model to ground its decision-making in invariant functional semantics rather than memorized token sequences. To ensure evaluation fairness, all perturbation patterns are automatically generated without any prior knowledge of the specific variation types used in the test set, reducing the risk of information leakage.

\subsection{Experimental Results}

We evaluate PAFT on the most challenging tasks from each tier: \textit{Semantic Trap} (Tier-1), \textit{Tool Redirection} (Tier-2), \textit{Logic Inversion} (Tier-3), and \textit{Fatal Error} (Tier-4). For Tiers~1--3, we report the Accuracy Delta ($\Delta$) relative to the closed-set baseline; for Tier-4, we report the absolute RR. We set $\alpha = 0.3$ as the default perturbation ratio.

\begin{table}[t]
    \centering
    \footnotesize
    \renewcommand{\arraystretch}{1.05}
    \setlength{\tabcolsep}{3pt}
    
    \caption{PAFT results across open environment tiers. }
    \label{tab:open_env_full}
    
    \begin{tabular}{l cccc}
        \toprule
        \multirow{2}{*}{\textbf{Model}} & \textbf{Tier-1} & \textbf{Tier-2} & \textbf{Tier-3} & \textbf{Tier-4} \\
        & Acc $\Delta${\scriptsize$\uparrow$} & Acc $\Delta${\scriptsize$\uparrow$} & Acc $\Delta${\scriptsize$\uparrow$} & RR{\scriptsize$\uparrow$} \\
        \midrule
        \multicolumn{5}{l}{\textit{Performance across training stages}} \\
        \midrule
        Base & -29.8 & -8.5 & -8.5 & 12.2 \\
        \midrule
        SFT-200 & -67.7 & -48.2 & -39.9 & 0.3 \\
        \quad + PAFT & \cellcolor{blue!6}\textbf{+28.6} & \cellcolor{blue!6}\textbf{+26.5} & \cellcolor{blue!6}\textbf{+22.7} & \cellcolor{blue!6}\textbf{99.3} \\
        SFT-400 & -53.9 & -45.4 & -32.5 & 0.0 \\
        \quad + PAFT & \cellcolor{blue!6}\textbf{+5.6} & \cellcolor{blue!6}\textbf{+4.9} & \cellcolor{blue!6}-2.8 & \cellcolor{blue!6}\textbf{97.8} \\
        SFT-600 & -51.3 & -46.4 & -33.0 & 0.1 \\
        \quad + PAFT & \cellcolor{blue!6}-2.5 & \cellcolor{blue!6}-2.9 & \cellcolor{blue!6}-10.7 & \cellcolor{blue!6}\textbf{99.6} \\
        SFT-800 & -50.4 & -45.3 & -28.0 & 0.2 \\
        \quad + PAFT & \cellcolor{blue!6}-4.1 & \cellcolor{blue!6}-5.3 & \cellcolor{blue!6}-9.8 & \cellcolor{blue!6}\textbf{99.6} \\
        \midrule
        \multicolumn{5}{l}{\textit{Ablation: perturbation ratio $\alpha$ (at 800 steps)}} \\
        \midrule
        $\alpha=0.2$ & -6.9 & -7.7 & -15.8 & 90.4 \\
        $\alpha=0.3$ (default) & \textbf{-4.1} & \textbf{-5.3} & \textbf{-9.8} & \textbf{99.6} \\
        $\alpha=0.4$ & -5.3 & -7.8 & -11.1 & 96.4 \\
        \midrule
        \multicolumn{5}{l}{\textit{Ablation: individual components (at 800 steps, $\alpha=0.3$)}} \\
        \midrule
        EFP only  & -30.1 & \textbf{-3.2} & -27.0 & 1.7 \\
        SBP only  & -38.9 & -36.5 & -31.7 & \textbf{99.2} \\
        SRP only  & \textbf{-14.9} & -34.2 & \textbf{-10.6} & 2.1 \\
        \bottomrule
    \end{tabular}
    \vspace{-15pt}

\end{table}
\textbf{Performance.}
As shown in the upper portion of Table~\ref{tab:open_env_full}, PAFT yields consistent improvements across all training stages and tiers. For early-stage models, PAFT transforms severe degradation into positive deltas across Tiers~1--3, indicating that perturbation-augmented training can preemptively prevent overfitting to clean trajectories.

\textbf{Ablation Study.}
The lower portion of Table~\ref{tab:open_env_full} reveals two key findings.
 
\textit{(1)~Perturbation ratio.}
Increasing $\alpha$ from 0.2 to 0.3 yields clear gains across all tiers, while further increasing to 0.4 provides marginal or no additional benefit.
We attribute this diminishing-return pattern to a trade-off between distributional coverage and signal quality: at low $\alpha$, the model is exposed to too few perturbed trajectories to learn robust recovery behaviors; as $\alpha$ grows, the augmented data progressively covers more failure regions, but beyond a saturation point, an excessive proportion of perturbed samples dilutes the clean supervision signal and introduces conflicting gradients between normal task completion and anomaly handling, leading to no further improvement or even slight regression.
The result suggests that $\alpha{=}0.3$ achieves a near-optimal balance between robustness and task performance.
 
\textit{(2)~Component contributions.}
EFP improves Tier-2 interaction adaptability, as the injected observation anomalies teach the model to parse and react to unexpected feedback rather than executing blindly along pre-planned trajectories. Notably, EFP contributes minimally to Tier-4 refusal, indicating that error adaptation and unsolvable-state recognition capture distinct capabilities. SBP almost exclusively restores Tier-4 refusal capability by introducing fatal error samples that break the implicit solvability assumption, while leaving other tiers largely unaffected, which suggests that, in our setting, boundary blindness is strongly tied to data coverage problem where the model never learns to refuse simply because $\mathcal{D}_{\text{clean}}$ contains no refusal demonstrations. SRP primarily enhances Tier-1 perception robustness and Tier-3 reasoning generalization, as symbolic perturbations force the model to attend to functional descriptions rather than surface-level identifiers, mitigating both symbolic anchoring and the associated downstream reasoning errors. This targeted effectiveness, where each component is necessary yet individually insufficient, confirms that PAFT is a structured, diagnosis-driven intervention rather than generic data augmentation.
\vspace{-2pt}
\section{Conclusion}
In this paper, we formalize the \textbf{OpenAgent} setting and leverage a controlled sandbox to  diagnose the open-world adaptability of LLM agents. Our systemic analysis reveals that both SFT and RL paradigms exhibit vulnerabilities when facing non-stationary environmental shifts. Furthermore, we propose PAFT, a disturbance-based intervention strategy designed to mitigate trajectory overfitting in SFT.

\section*{Acknowledgements}
This work was supported by the Key Program of Jiangsu Science Foundation (BK20243012), the NSFC (62306133), and the 111 Center (No.~B26023).
\section*{Impact Statement}
This work studies robustness and failure modes of tool-use agents under open-world shifts. Its primary positive impact is to improve the reliability and safety of deployed agents by identifying conditions under which models ignore feedback, misuse tools, or hallucinate answers. Potential risks include misuse of stronger tool-use agents in automated decision-making systems. We mitigate this by emphasizing refusal behavior, boundary awareness, and controlled evaluation rather than unrestricted deployment.

\bibliography{cited}

@article{qwen2.5vl,
  title={{Qwen2.5-VL} Technical Report},
  author={Bai, Shuai and Chen, Keqin and Liu, Xuejing and Wang, Jialin and Ge, Wenbin and Song, Sibo and Dang, Kai and Wang, Peng and Wang, Shijie and Tang, Jun and Zhong, Humen and Zhu, Yuanzhi and Yang, Mingkun and Li, Zhaohai and Wan, Jianqiang and Wang, Pengfei and Ding, Wei and Fu, Zheren and Xu, Yiheng and Ye, Jiabo and Zhang, Xi and Xie, Tianbao and Cheng, Zesen and Zhang, Hang and Yang, Zhibo and Xu, Haiyang and Lin, Junyang},
  journal={arXiv preprint arXiv:2502.13923},
  year={2025}
}

@article{qwen2.5,
  title={Qwen2.5 Technical Report},
  author={Qwen and Yang, An and Yang, Baosong and Zhang, Beichen and Hui, Binyuan and Zheng, Bo and Yu, Bowen and Li, Chengyuan and Liu, Dayiheng and Huang, Fei and Wei, Haoran and Lin, Huan and Yang, Jian and Tu, Jianhong and Zhang, Jianwei and Yang, Jianxin and Yang, Jiaxi and Zhou, Jingren and Lin, Junyang and Dang, Kai and Lu, Keming and Bao, Keqin and Yang, Kexin and Yu, Le and Li, Mei and Xue, Mingfeng and Zhang, Pei and Zhu, Qin and Men, Rui and Lin, Runji and Li, Tianhao and Tang, Tianyi and Xia, Tingyu and Ren, Xingzhang and Ren, Xuancheng and Fan, Yang and Su, Yang and Zhang, Yichang and Wan, Yu and Liu, Yuqiong and Cui, Zeyu and Zhang, Zhenru and Qiu, Zihan},
  journal={arXiv preprint arXiv:2412.15115},
  year={2024}
}

@inproceedings{mo2026open,
title={{Open-World} {LLM} Logical Reasoning},
author={Ye Mo and Chuan Zhou and Fengxiang Cheng and Jialin Yu and Liangming Pan and Fenrong Liu and Sheng Zhou and Haoxuan Li and Zhouchen Lin and Philip Torr},
booktitle={Forty-third International Conference on Machine Learning},
year={2026}
}

@article{gemini,
  title={Gemini: a family of highly capable multimodal models},
  author={Team, Gemini and Anil, Rohan and Borgeaud, Sebastian and Alayrac, Jean-Baptiste and Yu, Jiahui and Soricut, Radu and Schalkwyk, Johan and Dai, Andrew M and Hauth, Anja and Millican, Katie and others},
  journal={arXiv preprint arXiv:2312.11805},
  year={2023}
}

@inproceedings{rotbench,
  title={Rotbench: A multi-level benchmark for evaluating the robustness of large language models in tool learning},
  author={Ye, Junjie and Wu, Yilong and Gao, Songyang and Huang, Caishuang and Li, Sixian and Li, Guanyu and Fan, Xiaoran and Zhang, Qi and Gui, Tao and Huang, Xuan-Jing},
  booktitle={Proceedings of the 2024 Conference on Empirical Methods in Natural Language Processing},
  year={2024}
}

@inproceedings{retool,
  title={{ReTool}: Reinforcement Learning for Strategic Tool Use in {LLMs}},
  author={Feng, Jiazhan and Huang, Shijue and Qu, Xingwei and Zhang, Ge and Qin, Yujia and Zhong, Baoquan and Jiang, Chengquan and Chi, Jinxin and Zhong, Wanjun},
  booktitle={The Fourteenth International Conference on Learning Representations},
  year={2026}
}

@inproceedings{SFTRL,
  author       = {Tianzhe Chu and
                  Yuexiang Zhai and
                  Jihan Yang and
                  Shengbang Tong and
                  Saining Xie and
                  Dale Schuurmans and
                  Quoc V. Le and
                  Sergey Levine and
                  Yi Ma},
  title        = {{SFT} Memorizes, {RL} Generalizes: {A} Comparative Study of Foundation
                  Model Post-training},
 booktitle = {Proceedings of the 42nd International Conference on Machine Learning},
year         = {2025},
}

@inproceedings{toolrl,
  title={{ToolRL}: Reward is All Tool Learning Needs},
  author={Qian, Cheng and Acikgoz, Emre Can and He, Qi and Wang, Hongru and Chen, Xiusi and Hakkani-T{\"u}r, Dilek and Tur, Gokhan and Ji, Heng},
  booktitle={Advances in Neural Information Processing Systems},
  volume={38},
  year={2025}
}

@article{tool1,
	title        = {Tool documentation enables zero-shot tool-usage with large language models},
	author       = {Hsieh, Cheng-Yu and Chen, Si-An and Li, Chun-Liang and Fujii, Yasuhisa and Ratner, Alexander and Lee, Chen-Yu and Krishna, Ranjay and Pfister, Tomas},
	year         = 2023,
	journal      = {ArXiv:2308.00675}
}

@inproceedings{tool3,
  title={{COLT}: Towards Completeness-Oriented Tool Retrieval for Large Language Models},
  author={Qu, Changle and Dai, Sunhao and Wei, Xiaochi and Cai, Hengyi and Wang, Shuaiqiang and Yin, Dawei and Xu, Jun and Wen, Ji-Rong},
  booktitle={Proceedings of the 33rd ACM International Conference on Information and Knowledge Management},
  year={2024}
}

@inproceedings{tool5,
  title={Easytool: Enhancing llm-based agents with concise tool instruction},
  author={Yuan, Siyu and Song, Kaitao and Chen, Jiangjie and Tan, Xu and Shen, Yongliang and Ren, Kan and Li, Dongsheng and Yang, Deqing},
  booktitle={Proceedings of the 2025 Conference of the Nations of the Americas Chapter of the Association for Computational Linguistics: Human Language Technologies (Volume 1: Long Papers)},
  year={2025}
}

@inproceedings{tool6,
  author       = {Zhengliang Shi and
                  Shen Gao and
                  Xiuyi Chen and
                  Yue Feng and
                  Lingyong Yan and
                  Haibo Shi and
                  Dawei Yin and
                  Pengjie Ren and
                  Suzan Verberne and
                  Zhaochun Ren},
  editor       = {Yaser Al{-}Onaizan and
                  Mohit Bansal and
                  Yun{-}Nung Chen},
  title        = {Learning to Use Tools via Cooperative and Interactive Agents},
  booktitle    = {Findings of the Association for Computational Linguistics: {EMNLP}},
  series       = {Findings of {ACL}},
  volume       = {{EMNLP} 2024},
  year         = {2024},
}

@inproceedings{
shao2026chinatravel,
title={ChinaTravel: An Open-Ended Travel Planning Benchmark with Compositional Constraint Validation for Language Agents},
author={Jie-Jing Shao and Bo-Wen Zhang and Xiao-Wen Yang and Baizhi Chen and Siyu Han and Pang Jinghao and Wen-Da Wei and Guohao Cai and Zhenhua Dong and Lan-Zhe Guo and Yu-Feng Li},
booktitle={The Fourteenth International Conference on Learning Representations},
year={2026}
}

@inproceedings{huggingfacegpt,
  title={{HuggingGPT}: Solving {AI} Tasks with {ChatGPT} and its Friends in {Hugging Face}},
  author={Shen, Yongliang and Song, Kaitao and Tan, Xu and Li, Dongsheng and Lu, Weiming and Zhuang, Yueting},
  booktitle={Advances in Neural Information Processing Systems},
  volume={36},
  year={2023}
}

@inproceedings{toolvqa,
  title={Toolvqa: A dataset for multi-step reasoning vqa with external tools},
  author={Yin, Shaofeng and Lei, Ting and Liu, Yang},
  booktitle={Proceedings of the IEEE/CVF International Conference on Computer Vision},
  year={2025}
}

@inproceedings{deepeyes,
  title={{DeepEyes}: Incentivizing ``Thinking with Images'' via Reinforcement Learning},
  author={Zheng, Ziwei and Yang, Michael and Hong, Jack and Zhao, Chenxiao and Xu, Guohai and Yang, Le and Shen, Chao and Yu, Xing},
  booktitle={The Fourteenth International Conference on Learning Representations},
  year={2026}
}

@inproceedings{deepeyesv2,
  title={{DeepEyesV2}: Toward Agentic Multimodal Model},
  author={Hong, Jack and Zhao, Chenxiao and Zhu, ChengLin and Lu, Weiheng and Xu, Guohai and Yu, Xing},
  booktitle={The Fourteenth International Conference on Learning Representations},
  year={2026}
}

@inproceedings{mcpbench,
  title={{MCP-Bench}: Benchmarking Tool-Using {LLM} Agents with Complex Real-World Tasks via {MCP} Servers},
  author={Wang, Zhenting and Chang, Qi and Patel, Hemani and Biju, Shashank and Wu, Cheng-En and Liu, Quan and Ding, Aolin and Rezazadeh, Alireza and Shah, Ankit and Bao, Yujia and Siow, Eugene},
  booktitle={The Fourteenth International Conference on Learning Representations},
  year={2026}
}

@inproceedings{stabletoolbench,
  author       = {Zhicheng Guo and
                  Sijie Cheng and
                  Hao Wang and
                  Shihao Liang and
                  Yujia Qin and
                  Peng Li and
                  Zhiyuan Liu and
                  Maosong Sun and
                  Yang Liu},
  editor       = {Lun{-}Wei Ku and
                  Andre Martins and
                  Vivek Srikumar},
  title        = {StableToolBench: Towards Stable Large-Scale Benchmarking on Tool Learning
                  of Large Language Models},
  booktitle    = {Findings of the Association for Computational Linguistics},
  year         = {2024},
}

@inproceedings{gaia,
  title={Gaia: a benchmark for general ai assistants},
  author={Mialon, Gr{\'e}goire and Fourrier, Cl{\'e}mentine and Wolf, Thomas and LeCun, Yann and Scialom, Thomas},
  booktitle={The Twelfth International Conference on Learning Representations},
  year={2024}
}

@inproceedings{gentool,
  author       = {Jie He and
                  Jennifer Neville and
                  Mengting Wan and
                  Longqi Yang and
                  Hui Liu and
                  Xiaofeng Xu and
                  Xia Song and
                  Jeff Z. Pan and
                  Pei Zhou},
  editor       = {Wanxiang Che and
                  Joyce Nabende and
                  Ekaterina Shutova and
                  Mohammad Taher Pilehvar},
  title        = {GenTool: Enhancing Tool Generalization in Language Models through
                  Zero-to-One and Weak-to-Strong Simulation},
  booktitle    = {Findings of the Association for Computational Linguistics},
  year         = {2025},
}

@inproceedings{li2026flow,
    title = {In-the-Flow Agentic System Optimization for Effective Planning and Tool Use},
    author = {Li, Zhuofeng and Zhang, Haoxiang and Han, Seungju and Liu, Sheng and Xie, Jianwen and Zhang, Yu and Choi, Yejin and Zou, James and Lu, Pan},
  booktitle={The Fourteenth International Conference on Learning Representations},
  year={2026}
}

@article{openthinkimg,
  title={Openthinkimg: Learning to think with images via visual tool reinforcement learning},
  author={Su, Zhaochen and Li, Linjie and Song, Mingyang and Hao, Yunzhuo and Yang, Zhengyuan and Zhang, Jun and Chen, Guanjie and Gu, Jiawei and Li, Juntao and Qu, Xiaoye and others},
  journal={arXiv preprint arXiv:2505.08617},
  year={2025}
}

@article{toucan,
  title={Toucan: Synthesizing 1.5 m tool-agentic data from real-world mcp environments},
  author={Xu, Zhangchen and Soria, Adriana Meza and Tan, Shawn and Roy, Anurag and Agrawal, Ashish Sunil and Poovendran, Radha and Panda, Rameswar},
  journal={arXiv preprint arXiv:2510.01179},
  year={2025}
}

@article{pvpo,
  title={PVPO: Pre-Estimated Value-Based Policy Optimization for Agentic Reasoning},
  author={Feng, Wenfeng and Zhao, Penghong and Jiang, Guochao and Hao, Chuzhan and Zhang, Yuewei and Liu, Guohua and Wang, Hao},
  journal={arXiv preprint arXiv:2508.21104},
  year={2025}
}

@inproceedings{gta,
  title={{GTA}: A Benchmark for General Tool Agents},
  author={Wang, Jize and Zerun, Ma and Li, Yining and Zhang, Songyang and Chen, Cailian and Chen, Kai and Le, Xinyi},
  booktitle={Advances in Neural Information Processing Systems},
  volume={37},
  year={2024}
}

@inproceedings{mcpeval,
    title = "{MCPE}val: Automatic {MCP}-based Deep Evaluation for {AI} Agent Models",
    author = "Liu, Zhiwei  and
      Qiu, Jielin  and
      Wang, Shiyu  and
      Zhang, Jianguo  and
      Liu, Zuxin  and
      Ram, Roshan  and
      Chen, Haolin  and
      Yao, Weiran  and
      Heinecke, Shelby  and
      Savarese, Silvio  and
      Wang, Huan  and
      Xiong, Caiming",
    booktitle = "Proceedings of the 2025 Conference on Empirical Methods in Natural Language Processing: System Demonstrations",
    year = "2025",
}

@article{mcpradar,
  title={Mcp-radar: A multi-dimensional benchmark for evaluating tool use capabilities in large language models},
  author={Gao, Xuanqi and Xie, Siyi and Zhai, Juan and Ma, Shiqing and Shen, Chao},
  journal={arXiv preprint arXiv:2505.16700},
  year={2025}
}

@misc{anthropic2024mcp,
  title        = {Introducing the Model Context Protocol},
  author       = {{Anthropic}},
  year         = {2024},
  howpublished = {\url{https://www.anthropic.com/news/model-context-protocol}},
}

@article{butterfly,
  title={Butterfly effects in toolchains: A comprehensive analysis of failed parameter filling in llm tool-agent systems},
  author={Xiong, Qian and Huang, Yuekai and Jiang, Ziyou and Chang, Zhiyuan and Zheng, Yujia and Li, Tianhao and Li, Mingyang},
  journal={arXiv preprint arXiv:2507.15296},
  year={2025}
}

@article{glm45,
  title={Glm-4.5: Agentic, reasoning, and coding (arc) foundation models},
  author={Zeng, Aohan and Lv, Xin and Zheng, Qinkai and Hou, Zhenyu and Chen, Bin and Xie, Chengxing and Wang, Cunxiang and Yin, Da and Zeng, Hao and Zhang, Jiajie and others},
  journal={arXiv preprint arXiv:2508.06471},
  year={2025}
}

@article{kimik2,
  title={Kimi k2: Open agentic intelligence},
  author={Team, Kimi and Bai, Yifan and Bao, Yiping and Chen, Guanduo and Chen, Jiahao and Chen, Ningxin and Chen, Ruijue and Chen, Yanru and Chen, Yuankun and Chen, Yutian and others},
  journal={arXiv preprint arXiv:2507.20534},
  year={2025}
}

@inproceedings{dapo,
  title={{DAPO}: An Open-Source {LLM} Reinforcement Learning System at Scale},
  author={Yu, Qiying and Zhang, Zheng and Zhu, Ruofei and Yuan, Yufeng and Zuo, Xiaochen and Yue, Yu and Dai, Weinan and Fan, Tiantian and Liu, Gaohong and Liu, Lingjun and others},
  booktitle={Advances in Neural Information Processing Systems},
  volume={38},
  year={2025}
}

@inproceedings{apibank,
  author       = {Minghao Li and
                  Yingxiu Zhao and
                  Bowen Yu and
                  Feifan Song and
                  Hangyu Li and
                  Haiyang Yu and
                  Zhoujun Li and
                  Fei Huang and
                  Yongbin Li},
  editor       = {Houda Bouamor and
                  Juan Pino and
                  Kalika Bali},
  title        = {API-Bank: {A} Comprehensive Benchmark for Tool-Augmented LLMs},
  booktitle    = {Proceedings of the 2023 Conference on Empirical Methods in Natural
                  Language Processing},
  year         = {2023},
}

@inproceedings{metatool,
  author       = {Yue Huang and
                  Jiawen Shi and
                  Yuan Li and
                  Chenrui Fan and
                  Siyuan Wu and
                  Qihui Zhang and
                  Yixin Liu and
                  Pan Zhou and
                  Yao Wan and
                  Neil Zhenqiang Gong and
                  Lichao Sun},
  title        = {MetaTool Benchmark for Large Language Models: Deciding Whether to
                  Use Tools and Which to Use},
  booktitle    = {The Twelfth International Conference on Learning Representations},
  year         = {2024},
}

@article{gpt-4,
  title={{GPT}-4 technical report},
  author={Achiam, Josh and Adler, Steven and Agarwal, Sandhini and Ahmad, Lama and Akkaya, Ilge and Aleman, Florencia Leoni and Almeida, Diogo and Altenschmidt, Janko and Altman, Sam and Anadkat, Shyamal and others},
  journal={arXiv preprint arXiv:2303.08774},
  year={2023}
}

@inproceedings{tooleyes,
  author       = {Junjie Ye and
                  Guanyu Li and
                  Songyang Gao and
                  Caishuang Huang and
                  Yilong Wu and
                  Sixian Li and
                  Xiaoran Fan and
                  Shihan Dou and
                  Tao Ji and
                  Qi Zhang and
                  Tao Gui and
                  Xuanjing Huang},
  title        = {ToolEyes: Fine-Grained Evaluation for Tool Learning Capabilities of
                  Large Language Models in Real-world Scenarios},
  booktitle    = {Proceedings of the 31st International Conference on Computational
                  Linguistics},
  year         = {2025},
}

@inproceedings{benchmark3,
  author       = {Xingyao Wang and
                  Zihan Wang and
                  Jiateng Liu and
                  Yangyi Chen and
                  Lifan Yuan and
                  Hao Peng and
                  Heng Ji},
  title        = {{MINT:} Evaluating LLMs in Multi-turn Interaction with Tools and Language
                  Feedback},
  booktitle    = {The Twelfth International Conference on Learning Representations},
  year         = {2024},
}

@article{survey1,
  author       = {Lei Wang and
                  Chen Ma and
                  Xueyang Feng and
                  Zeyu Zhang and
                  Hao Yang and
                  Jingsen Zhang and
                  Zhiyuan Chen and
                  Jiakai Tang and
                  Xu Chen and
                  Yankai Lin and
                  Wayne Xin Zhao and
                  Zhewei Wei and
                  Jirong Wen},
  title        = {A survey on large language model based autonomous agents},
  journal      = {Frontiers Comput. Sci.},
  volume       = {18},
  year         = {2024},
}

@article{survey2,
  author       = {Changle Qu and
                  Sunhao Dai and
                  Xiaochi Wei and
                  Hengyi Cai and
                  Shuaiqiang Wang and
                  Dawei Yin and
                  Jun Xu and
                  Ji{-}Rong Wen},
  title        = {Tool learning with large language models: a survey},
  journal      = {Frontiers Comput. Sci.},
  volume       = {19},
  year         = {2025},
}

@inproceedings{agentbench,
  title     = {{AgentBench}: Evaluating {LLM}s as Agents},
  author    = {Liu, Xiao and Yu, Hao and Zhang, Hanchen and Xu, Yifan and Lei, Xuanyu and Lai, Hanyu and Gu, Yu and Ding, Hangliang and Men, Kaiwen and Yang, Kejuan and Zhang, Shudan and Deng, Xiang and Zeng, Aohan and Du, Zhengxiao and Zhang, Chenhui and Shen, Sheng and Zhang, Tianjun and Su, Yu and Sun, Huan and Huang, Minlie and Dong, Yuxiao and Tang, Jie},
  booktitle = {The Twelfth International Conference on Learning Representations},
  year      = {2024}
}

@inproceedings{webarena,
  title     = {{WebArena}: A Realistic Web Environment for Building Autonomous Agents},
  author    = {Zhou, Shuyan and Xu, Frank F. and Zhu, Hao and Zhou, Xuhui and Lo, Robert and Sridhar, Abishek and Cheng, Xianyi and Ou, Tianyue and Bisk, Yonatan and Fried, Daniel and Alon, Uri and Neubig, Graham},
  booktitle = {The Twelfth International Conference on Learning Representations},
  year      = {2024}
}

@inproceedings{workarena,
  title     = {{WorkArena}: How Capable Are Web Agents at Solving Common Knowledge Work Tasks?},
  author    = {Drouin, Alexandre and Gasse, Maxime and Caccia, Massimo and Laradji, Issam H. and Del Verme, Manuel and Marty, Tom and Boisvert, L{\'e}o and Thakkar, Megh and Cappart, Quentin and Vazquez, David and Chapados, Nicolas and Lacoste, Alexandre},
  booktitle = {Proceedings of the 41st International Conference on Machine Learning},
  year      = {2024}
}

@inproceedings{swebench,
  title     = {{SWE}-bench: Can Language Models Resolve Real-World {GitHub} Issues?},
  author    = {Jimenez, Carlos E. and Yang, John and Wettig, Alexander and Yao, Shunyu and Pei, Kexin and Press, Ofir and Narasimhan, Karthik},
  booktitle = {The Twelfth International Conference on Learning Representations},
  year      = {2024}
}

@inproceedings{visualwebarena,
  title     = {{VisualWebArena}: Evaluating Multimodal Agents on Realistic Visual Web Tasks},
  author    = {Koh, Jing Yu and Lo, Robert and Jang, Lawrence and Duvvur, Vikram and Lim, Ming Chong and Huang, Po-Yu and Neubig, Graham and Zhou, Shuyan and Salakhutdinov, Ruslan and Fried, Daniel},
  booktitle = {Proceedings of the 62nd Annual Meeting of the Association for Computational Linguistics},
  year      = {2024}
}

@inproceedings{appworld,
  title     = {{AppWorld}: A Controllable World of Apps and People for Benchmarking Interactive Coding Agents},
  author    = {Trivedi, Harsh and Khot, Tushar and Hartmann, Mareike and Manku, Ruskin and Dong, Vinty and Li, Edward and Gupta, Shashank and Sabharwal, Ashish and Balasubramanian, Niranjan},
  booktitle = {Proceedings of the 62nd Annual Meeting of the Association for Computational Linguistics},
  year      = {2024}
}

@inproceedings{agentboard,
  title     = {{AgentBoard}: An Analytical Evaluation Board of Multi-turn {LLM} Agents},
  author    = {Ma, Chang and Zhang, Junlei and Zhu, Zhihao and Yang, Cheng and Yang, Yujiu and Jin, Yaohui and Lan, Zhenzhong and Kong, Lingpeng and He, Junxian},
  booktitle = {Advances in Neural Information Processing Systems},
  volume    = {37},
  year      = {2024}
}

@inproceedings{osworld,
  title     = {{OSWorld}: Benchmarking Multimodal Agents for Open-Ended Tasks in Real Computer Environments},
  author    = {Xie, Tianbao and Zhang, Danyang and Chen, Jixuan and Li, Xiaochuan and Zhao, Siheng and Cao, Ruisheng and Hua, Toh Jing and Cheng, Zhoujun and Shin, Dongchan and Lei, Fangyu and Liu, Yitao and Xu, Yiheng and Zhou, Shuyan and Savarese, Silvio and Xiong, Caiming and Zhong, Victor and Yu, Tao},
  booktitle = {Advances in Neural Information Processing Systems},
  volume    = {37},
  year      = {2024}
}

@inproceedings{theagentcompany,
  title     = {{TheAgentCompany}: Benchmarking {LLM} Agents on Consequential Real-World Tasks},
  author    = {Xu, Frank F. and Song, Yufan and Li, Boxuan and Tang, Yuxuan and Jain, Kritanjali and Bao, Mengxue and Wang, Zora Z. and Zhou, Xuhui and Guo, Zhitong and Cao, Murong and Yang, Mingyang and Lu, Hao Yang and Martin, Amaad and Su, Zhe and Maben, Leander and Mehta, Raj and Chi, Wayne and Jang, Lawrence and Xie, Yiqing and Zhou, Shuyan and Neubig, Graham},
  booktitle = {Advances in Neural Information Processing Systems},
  volume    = {38},
  year      = {2025}
}

@article{openml_zhou,
  title={Open-environment machine learning},
  author={Zhou, Zhi-Hua},
  journal={National Science Review},
  volume={9},
  year={2022}
}

@article{li2021safewsl,
  title={Towards Safe Weakly Supervised Learning},
  author={Li, Yu-Feng and Guo, Lan-Zhe and Zhou, Zhi-Hua},
  journal={IEEE Transactions on Pattern Analysis and Machine Intelligence},
  volume={43},
  year={2021}
}

@inproceedings{guo2020safessl,
  title={Safe Deep Semi-Supervised Learning for Unseen-Class Unlabeled Data},
  author={Guo, Lan-Zhe and Zhang, Zhen-Yu and Jiang, Yuan and Li, Yu-Feng and Zhou, Zhi-Hua},
  booktitle={Proceedings of the 37th International Conference on Machine Learning},
  year={2020}
}

@inproceedings{guo2022classimb,
  title={Class-Imbalanced Semi-Supervised Learning with Adaptive Thresholding},
  author={Guo, Lan-Zhe and Li, Yu-Feng},
  booktitle={Proceedings of the 39th International Conference on Machine Learning},
  year={2022}
}

@article{guo2025robustssl,
  title={Robust Semi-Supervised Learning in Open Environments},
  author={Guo, Lan-Zhe and Jia, Lin-Han and Shao, Jie-Jing and Li, Yu-Feng},
  journal={Frontiers of Computer Science},
  volume={19},
  year={2025}
}

@inproceedings{jia2024realisticssl,
  title={Realistic Evaluation of Semi-Supervised Learning Algorithms in Open Environments},
  author={Jia, Lin-Han and Guo, Lan-Zhe and Zhou, Zhi and Li, Yu-Feng},
  booktitle={The Twelfth International Conference on Learning Representations},
  year={2024}
}

@inproceedings{tabfsbench,
  title={{TabFSBench}: Tabular Benchmark for Feature Shifts in Open Environment},
  author={Cheng, Zi-Jian and Jia, Zi-Yi and Zhou, Zhi and Li, Yu-Feng and Guo, Lan-Zhe},
  booktitle={Proceedings of the 42nd International Conference on Machine Learning},
  year={2025}
}

@inproceedings{hammer,
  title={Hammer: Robust Function-Calling for On-Device Language Models via Function Masking},
  author={Lin, Qiqiang and Wen, Muning and Peng, Qiuying and Nie, Guanyu and Liao, Junwei and Wang, Jun and Mo, Xiaoyun and Zhou, Jiamu and Cheng, Cheng and Zhao, Yin and Wang, Jun and Zhang, Weinan},
  booktitle={The Thirteenth International Conference on Learning Representations},
  year={2025}
}

@inproceedings{bfcl,
  title={The {Berkeley} Function Calling Leaderboard ({BFCL}): From Tool Use to Agentic Evaluation of Large Language Models},
  author={Patil, Shishir G. and Mao, Huanzhi and Yan, Fanjia and Ji, Charlie Cheng-Jie and Suresh, Vishnu and Stoica, Ion and Gonzalez, Joseph E.},
  booktitle={Proceedings of the 42nd International Conference on Machine Learning},
  year={2025}
}
\bibliographystyle{icml2026}

\newpage
\appendix
\onecolumn
\startcontents[appendix]
\printcontents[appendix]{}{1}{\section*{Appendix Contents}}
\section{Sandbox Environment}
    \label{app:sandbox}

Since live APIs and MCP-based tool ecosystems are often unstable, version-dependent, and difficult to control during evaluation, we build a local code-running sandbox to simulate tool calls. Specifically, we create a Python sandbox environment class that includes database, tools implementation, and tool call module. Detailed information is as follows:

\subsection{Database}
We built our experimental database using public Point of Interest (POI) data from Beijing. Each POI entry has five key attributes: Phone Number, Address Name, Adname (Administrative Name), Coordinates (Latitude/Longitude), and Category.
To strictly prevent data leakage and ensure the model relies entirely on tool usage rather than parameter memory to answer, we adopted an anonymization protocol. Specifically, all Address Names and Phone Numbers are replaced with unique, randomly generated alphanumeric strings and non-repeating number sequences, respectively. In addition, we strictly separate the training and test databases to avoid the model guessing the answers correctly by chance. Examples of the database data are shown in Table~\ref{tab:poi_samples}.
\begin{table}[H]
    \centering
    \caption{Samples of the Anonymized POI Knowledge Base}
    \label{tab:poi_samples}
    \begin{tabular}{ccccc}
        \toprule
        \textbf{Address Name} & \textbf{Phone Number} & \textbf{Adname} & \textbf{Coordinates} & \textbf{Category} \\ 
        \midrule
        exotlwnk & 00000000 & chaoyang & 116.413486, 39.992582 & C1 \\ 
        cqhecbvi & 00000001 & haidian  & 116.318014, 40.029475 & C2 \\ 
        rjztnhhg & 00000002 & changping & 116.303367, 40.160994 & C2 \\ 
        \bottomrule
    \end{tabular}
\end{table}

\begin{figure}[H]
    \centering
    \includegraphics[width=.4\linewidth]{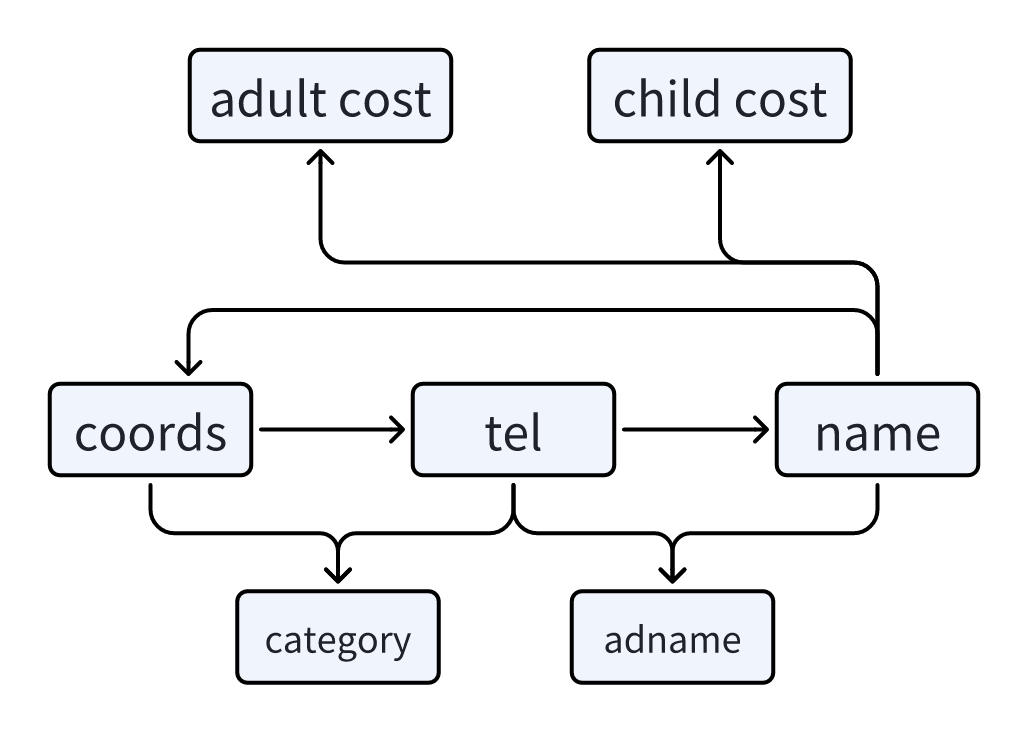}
    \caption{Logic Diagram of Tools. Each line represents one tool. The start of an arrow denotes the input, while the end of an arrow denotes the output.}
    \label{fig:logic}
\end{figure}

\subsection{Logical Architecture and Function Classification of Tools}
We designed the logical architecture of the tools in accordance with the logic diagram (Figure~\ref{fig:logic}) and the following rules:
\begin{enumerate}
    \item For key-value (coords, tel, name) query functions: we only support the counterclockwise query method by default, i.e., coords can only be used to query tel;
    \item For the fare query function: we only allow queries by name;
    \item For the distance calculation function: we only allow latitude and longitude as input parameters; 
    \item For non-key information queries: we allow any three key information to be used as input. 
\end{enumerate}
The above rules also restrict the input type of each tool, and we finally obtained 12 tools, which are shown in Table~\ref{tab:tool-definitions}.

\begin{table*}[ht]
    \centering
    \caption{Definitions of training set tools with categorized input and output types.}
    \label{tab:tool-definitions}
    \small 
    \begin{tabular}{ll|c|c}
        \toprule
        \textbf{Tool Name} & \textbf{Functional Description} & \textbf{Input} & \textbf{Output} \\ 
        \midrule
        \texttt{search\_map\_coordinates} & Retrieve location coordinates by exact name. & Name & Coords \\
        \texttt{administrativedivisionquery} & Retrieve the district name (adname) by exact name. & Name & District \\
        \texttt{fetch\_poi\_category} & Retrieve the category type by exact name. & Name & Category \\
        \midrule
        \texttt{find\_contact\_at\_coords} & Retrieve the telephone number using coordinates. & Coords & Tel \\
        \texttt{reverse\_geocoding} & Retrieve the district name using coordinates. & Coords & District \\
        \texttt{venue\_type\_identifier} & Retrieve the category type using coordinates. & Coords & Category \\
        \midrule
        \texttt{caller\_id\_lookup} & Retrieve the name of a POI using its telephone number. & Tel & Name \\
        \texttt{regional\_mapping\_by\_tel} & Retrieve the district name using telephone number. & Tel & District \\
        \texttt{service\_catalog\_search} & Retrieve the category type using telephone number. & Tel & Category \\
        \midrule
        \texttt{calculate\_distance\_by\_coords} & Calculate distance between two coordinates. & Coords $\times$ 2 & Distance \\
        \texttt{get\_adult\_cost\_by\_name} & Calculate discounted adult price (0.5x) by name. & Name & Price \\
        \texttt{get\_child\_cost\_by\_name} & Calculate discounted child price (0.2x) by name. & Name & Price \\
        \bottomrule
    \end{tabular}
\end{table*}

\subsection{Tool Call Module}
\label{toolcall}
We set up the different tool call modules for RL training and evaluation testing in two different ways:
\begin{itemize}
    \item For RL training, we use the Verl Tool framework. The core idea is to run the code sandbox as a service on the server and open an external port. This port receives tool names and corresponding parameters, then sends back the tool call results. During each RL training run, we use the terminator \textless tool\_call \textgreater to detect tool call commands. Specifically, when the model outputs this keyword, it stops generating text and triggers the tool detector. After processing all data in the same batch, the tool detector accesses the port uniformly to get results. These operations ensure only one tool call command is formed per conversation turn. The rest of the reinforcement learning process is the same as normal RL training.
    \item For evaluation testing, we simulate multi-turn conversations and also detect tool calls via the terminator, which ensures that only one tool call is triggered in a single turn. If the model outputs \textless /answer \textgreater, it means the response is finished. 
\end{itemize}
In summary, the experimental environment of our study is composed of the query database, the sandbox code execution environment, the training and test sets, as well as the training and testing logic. The case of interaction is shown below.

\begin{table}[ht]
\begin{tcolorbox}[
    title=Interaction case,
    label={case:interaction_example},
    colback=black!5!white,
    colframe=black,
    coltitle=white,
    fonttitle=\bfseries\ttfamily,
    halign title=center,
    rounded corners=downhill,
    arc=3mm,
    boxrule=0.8pt,
]
\texttt{
\{ \\
    \hspace*{1em} "role": "user", \\
    \hspace*{1em} "content": "I have the phone number '00001433'. Can you help me find the location of the place associated with this number?" \\
\}, \\
\{ \\
    \hspace*{1em} "role": "assistant", \\
    \hspace*{1em} "content": "\textless tool\_call\textgreater\{"tool": "caller\_id\_lookup", "args": \{"tel": "00001433"\}\}\textless /tool\_call\textgreater" \\
\}, \\
\{ \\
    \hspace*{1em} "role": "user", \\
    \hspace*{1em} "content": "Observation: \{"name": "sgtdfxvz" \}" \\
\}, \\
\{ \\
    \hspace*{1em} "role": "assistant", \\
    \hspace*{1em} "content": "There is no direct tool available to retrieve the location from a phone number. Therefore, I will first use the phone number to look up the name of the place. \textless tool\_call\textgreater\{"tool": "search\_map\_coordinates", "args": \{"name": "sgtdfxvz"\}\}\textless /tool\_call\textgreater" \\
\}, \\
\{ \\
    \hspace*{1em} "role": "user", \\
    \hspace*{1em} "content": "Observation: \{"location": "116.445179,39.953269" \}" \\
\}, \\
\{ \\
    \hspace*{1em} "role": "assistant", \\
    \hspace*{1em} "content": "Now that I have the name of the place, I can proceed to find its location using the name-based lookup tool. \textless answer\textgreater116.445179,39.953269\textless /answer\textgreater" \\
\}
}
\end{tcolorbox}
\end{table}

\subsection{Training and Test Data}
\label{TrainData}
\begin{figure}[t]
    \centering
    \begin{subfigure}{0.35\linewidth}
        \centering
        \includegraphics[width=\linewidth]{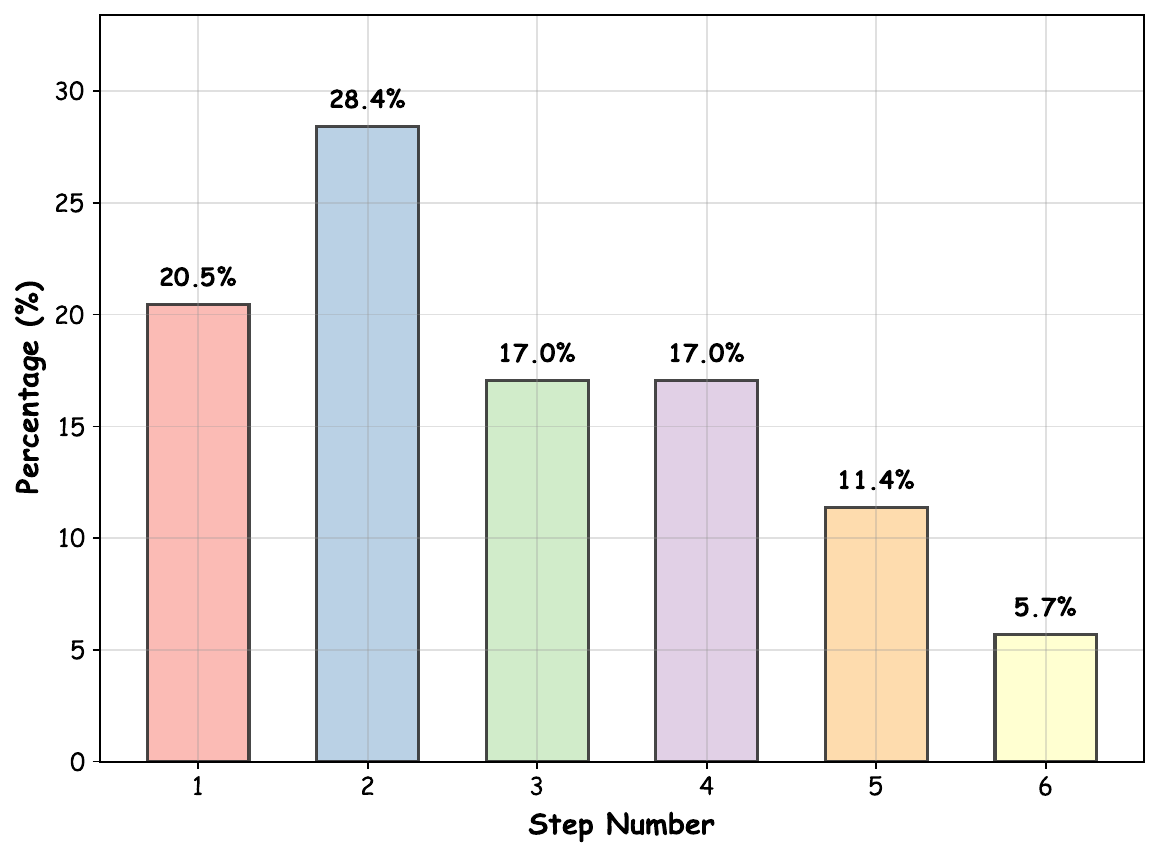}
        \caption{Step Distribution}
        \label{fig:step_dist}
    \end{subfigure}
    \hfill 
    \begin{subfigure}{0.32\linewidth}
        \centering
        \includegraphics[width=\linewidth]{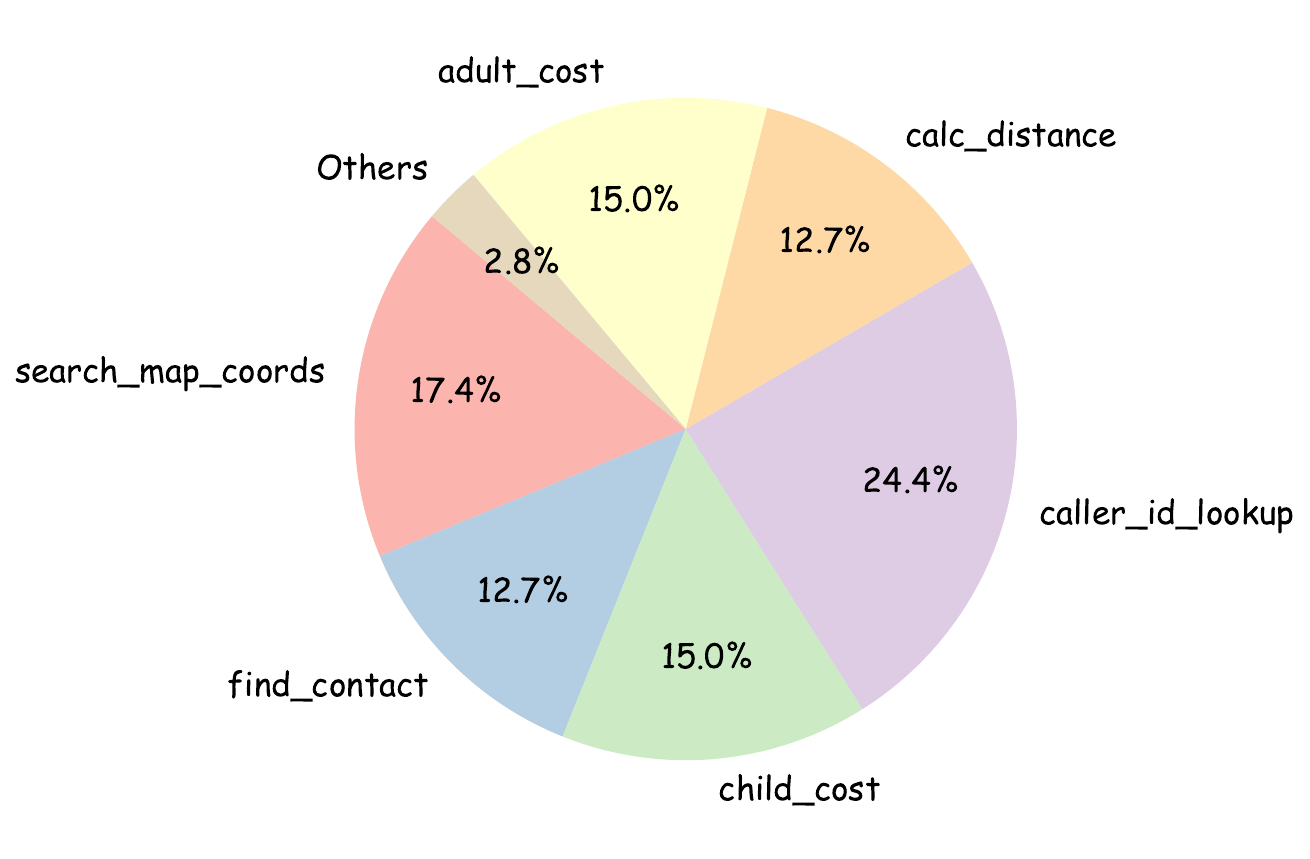}
        \caption{Tool Coverage}
        \label{fig:tool_cov}
    \end{subfigure}
    \hfill
    \begin{subfigure}{0.28\linewidth}
        \centering
        \includegraphics[width=\linewidth]{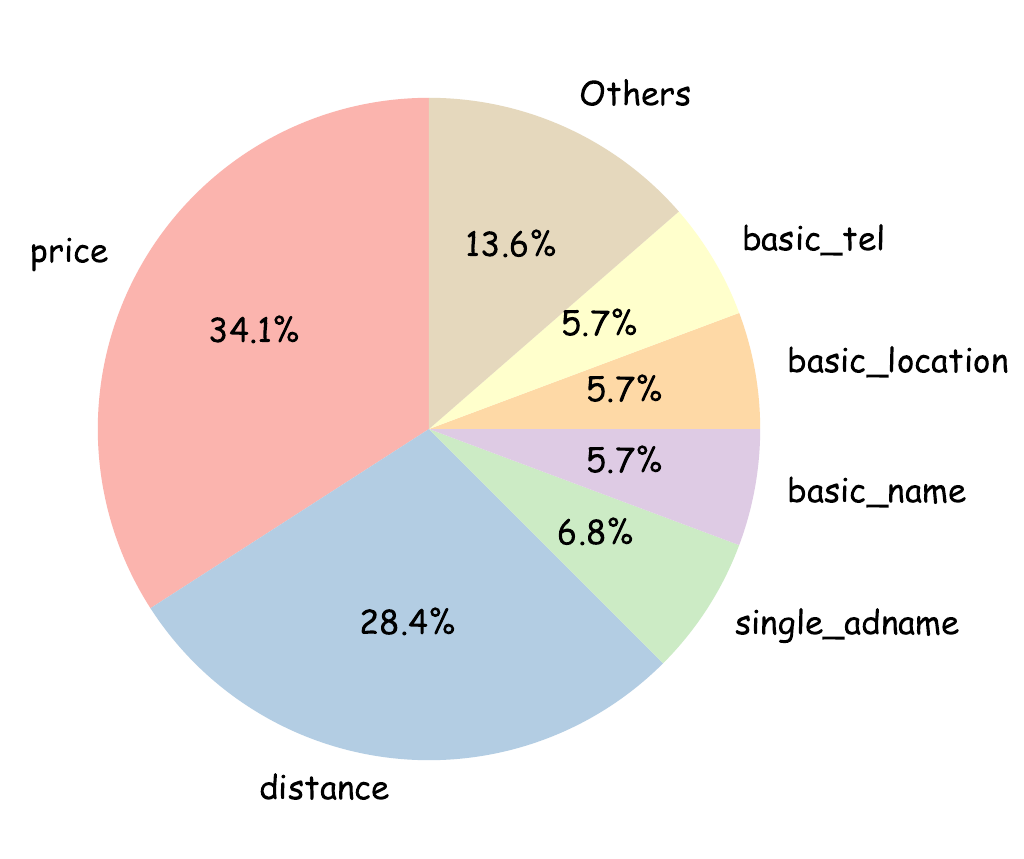}
        \caption{Source Distribution}
        \label{fig:source_dist}
    \end{subfigure}

    \caption{Overall Statistics of the POI Multi-step Tool-calling Dataset: (a) Distribution of task complexity by step count; (b) Proportions of various tool invocations; (c) Composition of task scenarios and sources.}
    \label{fig:overall_stats}
\end{figure}

Following data preparation in the main paper and the database and tools mentioned above, we employ a reverse-synthesis approach to generate training instances. By randomly sampling initial query conditions and target answers based on our defined tool set, we derive a symbolic reasoning trajectory. Formally, a symbolic trajectory $\mathcal{T}_{sym}$ is defined as:
\begin{equation}
    \mathcal{T}_{sym} = \{c, t_1 \to t_2 \to \dots \to t_k, a\}
\end{equation}
where $c$ represents the initial condition (e.g., a specific location coordinate), $t_i$ denotes the $i$-th tool invocation in the chain, and $a$ is the final answer. To align these symbolic representations with human interaction patterns, we utilize an LLM to transform and smooth $\mathcal{T}_{sym}$ into natural language question-answer pairs, resulting in a dataset of coherent multi-step tool-use queries.

The final dataset comprises a total of 6,930 samples, divided into a training set ($N=6,050$) and a test set ($N=880$). The dataset characteristics are designed to ensure rigorous evaluation:

\begin{itemize}
    \item \textbf{Complexity Distribution:} Both the training and test sets follow an identical distribution regarding the length of tool chains (e.g., 1-hop, 2-hop, ..., $k$-hop). This ensures that the test set accurately reflects the structural complexity of the training data.
    \item \textbf{Database Isolation:} To eliminate the possibility of data leakage, the training and test sets are synthesized from disjoint subsets of the underlying POI database. Consequently, the model must generalize its reasoning capabilities to unseen entities rather than recalling specific POI attributes seen during training.
\end{itemize}

Regarding task complexity distribution, the dataset employs a stepped logical chain design covering a range from one to six steps; two-step tasks serve as the core backbone (accounting for 28.4\%), while nearly 40\% of the samples consist of high-difficulty reasoning tasks involving three or more steps, ensuring depth in the model's learning of long-range tool invocation paths.

In terms of tool and scenario distribution, the dataset exhibits a high degree of domain focus and balance: the core tool caller\_id\_lookup represents 24.4\% of invocations, while ticket price calculation (adult/child cost) and geographic measurement (coords/distance) tools work in close coordination to form the foundation of task execution. This distribution is further corroborated by the source scenarios, where price-related (34.1\%) and distance-related (28.4\%) contexts dominate, providing the model with abundant material for numerical reasoning and logical alignment.

Overall, this dataset covers basic attribute queries (e.g., POI names, phone numbers, and administrative divisions) and utilizes complex step distributions to force the model into semantic grounding rather than simple symbolic matching during extreme testing, such as anonymization or logical inversion. With its scientific tool sampling proportions and diverse business scenarios, this dataset serves as an ideal benchmark for training agents equipped with efficient path planning and robust generalization capabilities. In addition, we also provide the tools that can be used in Table~\ref{tab:tool-definitions}.

\section{Experimental Details}
\label{app:setup}

We adopt Qwen2.5-7B-Instruct as the backbone model and perform full-parameter fine-tuning based on the LLaMA-Factory framework. The AdamW optimizer is employed, with the learning rate and batch size set to 3e-5 and 2, respectively. The training process runs for a total of 800 steps, with a model checkpoint saved every 200 steps. For the RL paradigm, training leverages the GRPO tool calling algorithm implemented on the Verl-Tool platform. Notably, we use the normal GRPO to keep comparison fairly. The AdamW optimizer is also used here, with the learning rate set to 1e-6, the number of deployments configured to 8, and the total training steps set to 280. The PAFT algorithm differs from SFT only in the training dataset.
The reward function was designed as follows:
$$
Reward= 0.8 \times Acc+0.2 \times score_{format}
$$
Note that we guarantee that both models are fully fitted and strictly control the amount of input data to ensure the fairness of the comparison.

Testing Setup: During the inference phase, all models were uniformly constrained with identical sampling hyperparameters: the sampling temperature was fixed at 0.0, and the Top-p value was set to 1.0. For modifying the tool settings, we conducted three experiments and analyzed the results comprehensively.

The system prompt template for training and testing is as follows:

\begin{table}[h]
\begin{tcolorbox}[
    title=System Prompt,
    colback=black!5!white,
    colframe=black,
    coltitle=white,
    fonttitle=\bfseries\ttfamily,
    halign title=center,
    rounded corners=downhill,
    arc=3mm,
    boxrule=0.8pt,
]
\scriptsize
\texttt{
"You are a professional geographic information assistant. You can answer questions about places, phone numbers, distances, and ticket prices by calling POI database tools. \\
\hspace*{1em} \#\#\# Available Tools: \\
\hspace*{1em} [ \\
\hspace*{1em} \{"name": "search\_map\_coordinates", "description": "Retrieve the location coordinates (longitude,latitude) of a POI by its exact name.", "parameters": \{"type": "object", "properties": \{"name": \{"type": "string"\}\}, "required": ["name"]\}\}, \\
\hspace*{1em} \{"name": "administrative\_division\_query", "description": "Retrieve the district name (adname) of a POI by its exact name.", "parameters": \{"type": "object", "properties": \{"name": \{"type": "string"\}\}, "required": ["name"]\}\}, \\
\hspace*{1em} \{"name": "fetch\_poi\_category", "description": "Retrieve the category type of a POI by its exact name.", "parameters": \{"type": "object", "properties": \{"name": \{"type": "string"\}\}, "required": ["name"]\}\}, \\
\hspace*{1em} \{"name": "find\_contact\_at\_coords", "description": "Retrieve the telephone number of a POI using its location coordinates.", "parameters": \{"type": "object", "properties": \{"location": \{"type": "string", "description": "'lon,lat' string"\}\}, "required": ["location"]\}\}, \\
\hspace*{1em} \{"name": "reverse\_geocoding", "description": "Retrieve the district name of a POI using its location coordinates.", "parameters": \{"type": "object", "properties": \{"location": \{"type": "string", "description": "'lon,lat' string"\}\}, "required": ["location"]\}\}, \\
\hspace*{1em} \{"name": "venue\_type\_identifier", "description": "Retrieve the category type of a POI using its location coordinates.", "parameters": \{"type": "object", "properties": \{"location": \{"type": "string", "description": "'lon,lat' string"\}\}, "required": ["location"]\}\}, \\
\hspace*{1em} \{"name": "caller\_id\_lookup", "description": "Retrieve the name of a POI using its telephone number.", "parameters": \{"type": "object", "properties": \{"tel": \{"type": "string"\}\}, "required": ["tel"]\}\}, \\
\hspace*{1em} \{"name": "regional\_mapping\_by\_tel", "description": "Retrieve the district name of a POI using its telephone number.", "parameters": \{"type": "object", "properties": \{"tel": \{"type": "string"\}\}, "required": ["tel"]\}\}, \\
\hspace*{1em} \{"name": "service\_catalog\_search", "description": "Retrieve the category type of a POI using its telephone number.", "parameters": \{"type": "object", "properties": \{"tel": \{"type": "string"\}\}, "required": ["tel"]\}\}, \\
\hspace*{1em} \{"name": "calculate\_distance\_by\_coords", "description": "Calculate the distance in meters between two specific coordinates. Requires mode=2.", "parameters": \{"type": "object", "properties": \{"start\_lat": \{"type": "number"\}, "start\_lon": \{"type": "number"\}, "end\_lat": \{"type": "number"\}, "end\_lon": \{"type": "number"\}, "mode": \{"type": "integer", "description": "Verification code."\}\}, "required": ["start\_lat", "start\_lon", "end\_lat", "end\_lon", "mode"]\}\}, \\
\hspace*{1em} \{"name": "get\_adult\_cost\_by\_name", "description": "Calculate the discounted adult ticket price (0.5x) by POI name. Requires mode=1.", "parameters": \{"type": "object", "properties": \{"name": \{"type": "string"\}, "mode": \{"type": "integer", "description": "Verification code."\}\}, "required": ["name", "mode"]\}\}, \\
\hspace*{1em} \{"name": "get\_child\_cost\_by\_name", "description": "Calculate the discounted child ticket price (0.2x) by POI name. Requires mode=1.", "parameters": \{"type": "object", "properties": \{"name": \{"type": "string"\}, "mode": \{"type": "integer", "description": "Verification code."\}\}, "required": ["name", "mode"]\}\} \\
\hspace*{1em} ] \\
\hspace*{1em} \#\#\# Protocol: \\
\hspace*{1em} 1. \textbf{ONE TOOL PER TURN}: You MUST output only ONE tool call at a time. Wait for the observation before doing anything else. \\
\hspace*{1em} 2. \textbf{Tool Format}: To use a tool, strictly output a JSON object with "name" and "arguments" keys. Example: \{"name": "get\_tel\_by\_name", "arguments": \{"name": "Beijing\_Hotel"\}\} \\
\hspace*{1em} 3. \textbf{Final Answer}: Wrap the result in \textless answer\textgreater\textless /answer\textgreater. Keep numerical answers to 1 decimal place."
}
\end{tcolorbox}
\end{table}

\section{Implementation Details of OpenAgent Evaluation Task}
\label{implementation}
\subsection{Tier 1: Perception Generalization}
\label{sec:tier1_perception}
\subsubsection{Instruction Robustness.}
\begin{itemize}[nosep, leftmargin=*]
    \item \textbf{Instruction Blurring:} We employ API calls to rewrite user queries, introducing complexities through syntactic variations (e.g., colloquialisms, rhetorical questions) or semantic ambiguities (e.g., implicit references, entity omissions). 
    \item \textbf{Information Redundancy:} We employ API calls to rewrite user queries by embedding core instructions within irrelevant contexts or lengthy background details, thereby simulating real-world informational noise. Please refer to the Appendix~\ref{app:tier1-1} for further details.
\end{itemize}

\subsubsection{Schema Adaptability.}
We alter the parameter dictionary key from the standard \texttt{args} to \texttt{arguments}, as shown in Appendix~\ref{app:tier1-2}.

\subsubsection{Semantic Grounding.}

\textbf{Synonymous Rewriting} We assess the model's capability to decouple functional intent from specific symbolic labels by replacing tool identifiers with semantically equivalent alternatives. This requires the model to move beyond simple string matching and demonstrate a deep understanding of the underlying natural language semantics.
\begin{table}[H]
    \centering
    \caption{Tool Mapping for \textbf{Synonym Rewriting}}
    \small
    \begin{tabular}{lcl}
        \toprule
        \textbf{Original Tool Name} & $\rightarrow$ & \textbf{Synonymous Variant} \\
        \midrule
        \texttt{calculate\_distance\_by\_coords} & $\rightarrow$ & \texttt{calculate\_separation\_via\_positions} \\
        \texttt{search\_map\_coordinates} & $\rightarrow$ & \texttt{lookup\_location\_latlong} \\
        \texttt{get\_adult\_cost\_by\_name} & $\rightarrow$ & \texttt{fetch\_grownup\_price\_using\_name} \\
        \bottomrule
    \end{tabular}
\end{table}

\textbf{Noise Injection} While maintaining the integrity of the tool schema, we introduce subtle syntactic perturbations into the tool names by inserting articles or determiners. We utilize this setup to evaluate the model's tolerance to minor fluctuations in API naming and its stability when encountering non-standardized inputs.
\begin{table}[H]
    \centering
    \caption{Tool Mapping for \textbf{Surface Noise}}
    \small
    \begin{tabular}{lcl}
        \toprule
        \textbf{Original Tool Name} & $\rightarrow$ & \textbf{Noisy Variant} \\
        \midrule
        \texttt{calculate\_distance\_by\_coords} & $\rightarrow$ & \texttt{calculate\_the\_distance\_by\_coords} \\
        \texttt{search\_map\_coordinates} & $\rightarrow$ & \texttt{search\_for\_map\_coordinates} \\
        \texttt{get\_adult\_cost\_by\_name} & $\rightarrow$ & \texttt{get\_1\_adult\_cost\_by\_name} \\
        \bottomrule
    \end{tabular}
\end{table}

\textbf{Semantic Adversarial} We designed a ``description trap'' where the familiar tool names are retained, but their functional descriptions are modified to explicitly forbid their previous usage. We require the model to overcome ``path dependency'' on familiar names, recognize the obsolescence of the original tool based on the updated description, and accurately invoke a new tool with matching functionality.

\begin{table}[H]
    \centering
    \caption{Changes in \textbf{Semantic Adversarial}}
    \small
    \begin{tabular}{lcl}
        \toprule
        \textbf{Status} & \textbf{Tool Name} & \textbf{Functional Role} \\
        \midrule
        Obsolete & \texttt{get\_adult\_cost\_by\_name} & Service fee calculation only (Forbidden) \\
        Target & \texttt{fetch\_grownup\_price\_using\_name} & Actual ticket price calculation (Correct) \\
        \bottomrule
    \end{tabular}
\end{table}

\textbf{Identity Erasure} To isolate the model's reliance on pure functional logic, we replace all human-readable identifiers with randomly generated UUIDs. In this extreme setting, we force the agent to perform reasoning based exclusively on the provided natural language descriptions.

\begin{table}[H]
    \centering
    \caption{Tool Mapping for \textbf{Identity Erasure}}
    \small
    \begin{tabular}{lcl}
        \toprule
        \textbf{Semantic Tool Name} & $\rightarrow$ & \textbf{Anonymous Identifier} \\
        \midrule
        \texttt{calculate\_distance\_by\_coords} & $\rightarrow$ & \texttt{Tool\_a2fb} \\
        \texttt{search\_map\_coordinates} & $\rightarrow$ & \texttt{Tool\_c3d8} \\
        \texttt{get\_adult\_cost\_by\_name} & $\rightarrow$ & \texttt{Tool\_r6t7} \\
        \bottomrule
    \end{tabular}
\end{table}

\subsection{Tier 2: Interaction Generalization}
\subsubsection{Format Adaptability.}
We diversify the structure of tool return values, transitioning from strictly structured JSON to less constrained formats such as Markdown tables or raw text logs.
\begin{table}[H]
    \centering
    \caption{Example of \textbf{Format Shift} in Tool Observations}
    \small
    \begin{tabular}{lc}
        \toprule
        \textbf{Type} & \textbf{Content Structure} \\
        \midrule
        \textbf{Structured (Old)} & \texttt{"content": "Observation: \{location: 116.35,40.22\} "} \\
        \textbf{Unstructured (New)} & \texttt{"content": "Observation: \{location is 116.35,40.22\} "} \\
        \bottomrule
    \end{tabular}
\end{table}

\subsubsection{Error Correction.}
\begin{itemize}
\item \textbf{Error Return:} We simulate an explicit failure where the primary tool returns a specific error message (e.g., \textit{``This tool is unavailable now''}) or a traceback. This scenario evaluates whether the agent can interpret explicit error signals and proactively switch to a redundant tool to resolve the blockage.
\item \textbf{Null Return:} We introduce an implicit failure where the primary tool returns a null value or an empty set ($\emptyset$) without any diagnostic warning. This tests the agent's capability to detect anomalies based on the absence of content, rather than the presence of an error signal, and infer the necessity of a fallback execution.
\end{itemize}

\begin{table}[H]
    \centering
    \caption{Mapping of \textbf{Tool Failure Modes} and Expected Agent Responses}
    \small
    \begin{tabular}{lcc}
        \toprule
        \textbf{Failure Type} & \textbf{Tool Observation (Return)} \\
        \midrule
        \textbf{Error Return} & \texttt{\{"error": "This tool is unavailable now"\}} \\
        \textbf{Null Return} &  \texttt{""} \\
        \bottomrule
    \end{tabular}
\end{table}
\subsubsection{Dynamic Alignment.}
\begin{itemize}[nosep, leftmargin=*] 
\item \textbf{Value Redirection:} We simulate a scenario where the provided parameter logic is deprecated, and the environment returns a message including new, valid parameters. 
\item \textbf{Tool Redirection:} We simulate a scenario where the primary tool returns an error explicitly stating it is obsolete while simultaneously suggesting an alternative tool.\end{itemize}

Importantly, we feed the interaction history into the API model (DouBao) to enable it to judge whether the target model refuses to answer.

\subsection{Tier 3: Reasoning Generalization}
\subsubsection{Rule Reasoning.}
We introduce rule conflicts in a special tool by modifying the mathematical constants in the tool description. Specifically, we kept the tool interface unchanged, but significantly adjusted the calculation ratio in the Prompt in the test set to examine whether the model relies on the memory from the training stage or can perform real-time reasoning based on immediate instructions.
\begin{table}[H]
    \centering
    \caption{Experimental Setup for \textbf{Calculation Shift}: Transitioning from Training Logic to Dynamic Test Logic}
    \label{tab:calculation_shift}
    \small
    \begin{tabular}{lcc}
        \toprule
        \textbf{Tool Name} & \textbf{Training rule (Origin)} & \textbf{Test rule (Changed)} \\
        \midrule
        \texttt{get\_adult\_cost} & Discounted Price = \textbf{0.5x} Original Price & Discounted Price = \textbf{0.1x} Original Price \\
        \midrule
        \texttt{get\_child\_cost} & Discounted Price = \textbf{0.2x} Original Price & Discounted Price = \textbf{0.1x} Original Price \\
        \bottomrule
    \end{tabular}
\end{table}

\subsubsection{Path Planning.}

\textbf{Trace Simplification} We implement Tool shortcut to evaluate the agent's ability to optimize its reasoning trajectory when more efficient paths become available. In this setup, we condense the functionality of two separate, sequential tools into a single, streamlined tool. This transformation reduces a multi-step dependency chain into a one-step execution.
\begin{table}[H]
    \centering
    \caption{Implementation of \textbf{Tool Consolidation}}
    \label{tab:tool_consolidation}
    \small
    \begin{tabular}{lll}
        \toprule
        \textbf{Original Multi-step Tools} & $\rightarrow$ & \textbf{Consolidated Tool (Single Step)} \\
        \midrule
        \texttt{search\_map\_coords} + \texttt{find\_contact\_coords} & $\rightarrow$ & \texttt{get\_tel\_by\_name} \\
        \texttt{find\_contact\_coords} + \texttt{caller\_id\_lookup} & $\rightarrow$ & \texttt{get\_name\_by\_location} \\
        \texttt{caller\_id\_lookup} + \texttt{search\_map\_coords} & $\rightarrow$ & \texttt{get\_location\_by\_tel} \\
        \bottomrule
    \end{tabular}
\end{table}

\textbf{Logic Inversion} 
We simulate causal inversions in the dependency figure~\ref{fig:logic}. During the testing phase, we intentionally reverse the information acquisition order described in the documentation, effectively inverting the preconditions of the tools. For example, if a task originally required Tool A to precede Tool B, we redefine the logic such that Tool A now depends on the output of Tool B.
\begin{table}[H]
    \centering
    \caption{Implementation of \textbf{Causal Reversal}}
    \label{tab:causal_reversal}
    \small
    \begin{tabular}{lll}
        \toprule
        \textbf{Original Tool Logic} & $\rightarrow$ & \textbf{Reversed Causal Logic (Test)} \\
        \midrule
        \texttt{search\_map\_coords}: Name $\rightarrow$ \textbf{Coords} & $\rightarrow$ & \texttt{lookup\_name\_by\_coords}: Coords $\rightarrow$ \textbf{Name} \\
        \texttt{find\_contact\_coords}: Coords $\rightarrow$ \textbf{Tel} & $\rightarrow$ & \texttt{get\_coords\_by\_tel}: Tel $\rightarrow$ \textbf{Coords} \\
        \texttt{caller\_id\_lookup}: Tel $\rightarrow$ \textbf{Name} & $\rightarrow$ & \texttt{search\_tel\_by\_name}: Name $\rightarrow$ \textbf{Tel} \\
        \bottomrule
    \end{tabular}
\end{table}

\subsection{Tier 4: Internalization Generalization}
\subsubsection{Active Refusal} We implement Active Refusal by constructing ``Missing Condition'' and ``Fatal Error'' scenarios, transitioning tasks from being ``inevitably solvable'' to ``deterministically unsolvable.'' We utilize this setup to assess the model's capability to recognize task boundaries—specifically, whether it can choose to proactively refuse the request rather than generating hallucinations when faced with information deficiency or execution deadlocks. The error tools include \texttt{calculate\_distance\_by\_coords}, \texttt{search\_map\_coordinates}, and \texttt{get\_adult\_cost\_by\_name}.
\begin{table}[H]
    \centering
    \caption{Implementation of \textbf{Active Refusal} via Insufficient Conditions}
    \label{tab:active_refusal}
    \small
    \begin{tabular}{lll}
        \toprule
        \textbf{Scenarios} & \textbf{Original State (Solvable)} & \textbf{Perturbed State (Unsolvable)} \\
        \midrule
        \textbf{Missing Condition} & Complete information in User Query & \textbf{Randomly excised} key entities/constraints \\
        \textbf{Fatal Error} & Successful Tool Execution & \textbf{Irrecoverable error} with no alternative tools \\
        \bottomrule
    \end{tabular}
\end{table}

\subsubsection{Domain Transfer}
We implement a Domain Shift by changing the distributions of all three interactive elements: the user query, tool definitions, and environmental observations. Specifically, we transfer the evaluation environment from a geographic address lookup task to a medical registration task. While the surface presentation of the tasks is entirely altered, we retain the latent causal logical structure (Figure~\ref{fig:logic}) invariance.

\begin{table}[H]
    \centering
    \caption{One-to-One Tool Mapping for Joint Domain Shift}
    \label{tab:tool_domain_mapping}
    \small
    \begin{tabular}{l|l|c}
        \toprule
        \textbf{Geographic Domain (Source $P_{\text{train}}$)} & \textbf{Medical Domain (Target $P_{\text{test}}$)} & \textbf{Latent Logic} \\
        \midrule
        \texttt{search\_map\_coordinates} & \texttt{get\_medical\_facility\_coords} & $Name \rightarrow Coords$ \\
        \texttt{administrative\_division\_query} & \texttt{query\_medical\_administration} & $Name \rightarrow Admin$ \\
        \texttt{fetch\_poi\_category} & \texttt{fetch\_medical\_service\_type} & $Name \rightarrow Category$ \\
        \midrule
        \texttt{find\_contact\_at\_coords} & \texttt{find\_emergency\_hotline\_at\_coords} & $Coords \rightarrow Tel$ \\
        \texttt{reverse\_geocoding} & \texttt{reverse\_medical\_geocoding} & $Coords \rightarrow Admin$ \\
        \texttt{venue\_type\_identifier} & \texttt{identify\_clinic\_category} & $Coords \rightarrow Category$ \\
        \midrule
        \texttt{caller\_id\_lookup} & \texttt{lookup\_facility\_by\_hotline} & $Tel \rightarrow Name$ \\
        \texttt{regional\_mapping\_by\_tel} & \texttt{query\_medical\_region\_by\_tel} & $Tel \rightarrow Admin$ \\
        \texttt{service\_catalog\_search} & \texttt{classify\_service\_by\_tel} & $Tel \rightarrow Category$ \\
        \midrule
        \texttt{calculate\_distance\_by\_coords} & \texttt{calculate\_ambulance\_route\_distance} & $Math (mode=2)$ \\
        \texttt{get\_adult\_cost\_by\_name} & \texttt{get\_checkup\_cost\_by\_name} & $Cost A (mode=1)$ \\
        \texttt{get\_child\_cost\_by\_name} & \texttt{get\_vaccine\_cost\_by\_name} & $Cost B (mode=1)$ \\
        \bottomrule
    \end{tabular}
\end{table}

\section{Detailed Experimental Results}
\label{detailed_results}
\subsection{Tier 1 Experiments}
We report the full numerical results for all Tier-1 perception experiments.
Tables~\ref{tab:invocation_variation}--\ref{tab:performance_steps_variant} present accuracy and ATCL under instruction-level perturbations (Instruction Blurring and Information Redundancy).
Table~\ref{tab:schema_adaptability} reports performance under schema keyword changes.
Tables~\ref{tab:parameter_variation_fixed_visibility}--\ref{tab:detailed_results_three_tools} provide per-tool breakdowns across three semantic grounding experiments (Parameter Drift, Synonymous Rewriting, Noise Injection, Semantic Adversarial, and Identity Erasure), where each tool is evaluated independently to isolate tool-specific failure patterns.
Across all settings, SFT exhibits increasing symbolic anchoring as training progresses, while RL maintains more stable accuracy retention.
\begin{table}[H]
    \centering
    \caption{Detailed Results of Instruction Blurring}
    \label{tab:invocation_variation}
    \begin{tabularx}{0.9\linewidth}{lYYYYY}
        \toprule
        \rowcolor{gray!20} \textbf{Model/Step} & \textbf{Acc (Abs)} & \textbf{ATCL Ratio} & \textbf{ATCL (Abs)} & \textbf{ATCL Ratio} & \textbf{ATCL Diff} \\ 
        \midrule
        base & 20.91 & 71.32 & 3.76 & 101.35 & +0.05 \\
        \midrule
        \rowcolor{blue!3} sft-200 & 78.98 & 120.45 & 2.78 & 128.70 & +0.62 \\
        \rowcolor{blue!3} sft-400 & 67.39 & 83.40  & 2.70 & 96.09  & -0.11 \\
        \rowcolor{blue!3} sft-600 & 78.07 & 95.69  & 2.84 & 99.30  & -0.02 \\
        \rowcolor{blue!3} sft-800 & 78.30 & 95.17  & 2.83 & 98.26  & -0.05 \\
        \midrule
        \rowcolor{green!3} rl-40  & 34.43 & 63.78  & 2.14 & 93.86  & -0.14 \\
        \rowcolor{green!3} rl-80  & 58.64 & 79.63  & 2.45 & 95.33  & -0.12 \\
        \rowcolor{green!3} rl-160 & 67.27 & 82.22  & 2.56 & 95.52  & -0.12 \\
        \rowcolor{green!3} rl-240 & 74.89 & 81.86  & 2.55 & 95.15  & -0.13 \\
        \rowcolor{green!3} rl-280 & 74.77 & 81.84  & 2.57 & 95.90  & -0.11 \\
        \bottomrule
    \end{tabularx}
\end{table}

\begin{table}[H]
    \centering
    \caption{Detailed Results of Information Redundancy}
    \label{tab:performance_steps_variant}
    \begin{tabularx}{0.9\linewidth}{lYYYYY}
        \toprule
        \rowcolor{gray!20} \textbf{Model/Step} & \textbf{Acc (Abs)} & \textbf{Acc Ratio} & \textbf{ATCL (Abs)} & \textbf{ATCL Ratio} & \textbf{ATCL Diff} \\ 
        \midrule
        base & 13.98 & 47.68 & 4.01 & 108.09 & +0.30 \\
        \midrule
        \rowcolor{blue!3} sft-200 & 84.66 & 129.11 & 2.98 & 137.96 & +0.82 \\
        \rowcolor{blue!3} sft-400 & 67.73 & 83.82  & 3.10 & 110.32 & +0.29 \\
        \rowcolor{blue!3} sft-600 & 77.39 & 94.85  & 3.16 & 110.49 & +0.30 \\
        \rowcolor{blue!3} sft-800 & 75.23 & 91.44  & 3.22 & 111.81 & +0.34 \\
        \midrule
        \rowcolor{green!3} rl-40  & 38.86 & 71.99  & 2.29 & 100.44 & +0.01 \\
        \rowcolor{green!3} rl-80  & 54.77 & 74.38  & 2.36 & 91.83  & -0.21 \\
        \rowcolor{green!3} rl-160 & 55.68 & 68.05  & 2.57 & 95.90  & -0.11 \\
        \rowcolor{green!3} rl-240 & 61.48 & 67.21  & 2.58 & 96.27  & -0.10 \\
        \rowcolor{green!3} rl-280 & 61.02 & 66.79  & 2.56 & 95.52  & -0.12 \\
        \bottomrule
    \end{tabularx}
\end{table}

\begin{table}[H]
    \centering
    \caption{Detailed Results of Schema Adaptability}
    \label{tab:schema_adaptability}
    \begin{tabularx}{0.9\linewidth}{lYYYYY}
        \toprule
        \rowcolor{gray!20} \textbf{Model/Step} & \textbf{Acc (Abs)} & \textbf{Acc Ratio} & \textbf{ATCL (Abs)} & \textbf{ATCL Ratio} & \textbf{ATCL Diff} \\ 
        \midrule
        base & 31.82 & 108.53 & 3.72 & 100.27 & +0.01 \\
        \midrule
        \rowcolor{blue!3} sft-200 & 76.59 & 116.81 & 2.55 & 118.06 & +0.39 \\
        \rowcolor{blue!3} sft-400 & 82.95 & 102.66 & 2.86 & 101.78 & +0.05 \\
        \rowcolor{blue!3} sft-600 & 81.93 & 100.42 & 2.87 & 100.35 & +0.01 \\
        \rowcolor{blue!3} sft-800 & 81.59 & 99.17  & 2.88 & 100.00 & +0.00 \\
        \midrule
        \rowcolor{green!3} rl-40  & 61.82 & 114.52 & 2.49 & 109.21 & +0.21 \\
        \rowcolor{green!3} rl-80  & 82.39 & 111.88 & 2.74 & 106.61 & +0.17 \\
        \rowcolor{green!3} rl-160 & 81.48 & 99.58  & 2.70 & 100.75 & +0.02 \\
        \rowcolor{green!3} rl-240 & 92.27 & 100.86 & 2.66 & 99.25  & -0.02 \\
        \rowcolor{green!3} rl-280 & 92.39 & 101.13 & 2.66 & 99.25  & -0.02 \\
        \bottomrule
    \end{tabularx}
\end{table}

\begin{table}[H]
    \centering
    \caption{Detailed Results of Parameter Drift across three tools. T1: calculate\_distance\_by\_coords; T2: search\_map\_coordinates; T3: get\_adult\_cost\_by\_name.}
    \label{tab:parameter_variation_fixed_visibility}
    \resizebox{.9\linewidth}{!}{
    \begin{tabular}{lcccccccc}
        \toprule
        \rowcolor{gray!20} \textbf{Model} & \textbf{Tool} & \textbf{Acc (Abs)} & \textbf{Acc Ratio} & \textbf{Tool (Abs)} & \textbf{Tool Ratio} & \textbf{Tool Diff} & \textbf{Fail Rate} & \textbf{Ratio (Base)} \\
        \midrule
        \multirow{3}{*}{base}    & T1 & 25.80 & 87.99  & 3.68 & 99.19  & -0.03 & 20.04 & 152.88 \\
                                 & T2 & 26.14 & 89.15  & 3.74 & 100.81 & +0.03 & 18.11 & 137.89 \\
                                 & T3 & 30.11 & 102.69 & 3.71 & 100.00 & +0.00 & 11.25 & 112.27 \\
        \midrule
        \rowcolor{blue!3}        & T1 & 66.82 & 101.91 & 2.18 & 100.93 & +0.02 & 0.15  & 96.85  \\
        \rowcolor{blue!3} \multirow{-2}{*}{sft-200} & T2 & 66.82 & 101.91 & 2.18 & 100.93 & +0.02 & 0.15  & 97.73  \\
        \rowcolor{blue!3}        & T3 & 66.93 & 102.07 & 2.19 & 101.39 & +0.03 & 6.80  & 85.91  \\
        \midrule
        \rowcolor{blue!3} & T1 & 80.80 & 100.00 & 2.80 & 99.64  & -0.01 & 1.34  & 94.51  \\
        \rowcolor{blue!3} \multirow{-2}{*}{sft-400}  & T2 & 80.23 & 99.29  & 2.80 & 99.64  & -0.01 & 1.17  & 82.49  \\
        \rowcolor{blue!3} & T3 & 80.23 & 99.29  & 2.79 & 99.29  & -0.02 & 0.95  & 100.24 \\
        \midrule
        \rowcolor{blue!3} & T1 & 81.36 & 99.72  & 2.86 & 100.00 & +0.00 & 0.66  & 89.25  \\
        \rowcolor{blue!3} \multirow{-2}{*}{sft-600} & T2 & 81.36 & 99.72  & 2.86 & 100.00 & +0.00 & 0.41  & 100.00 \\
        \rowcolor{blue!3} & T3 & 81.93 & 100.42 & 2.86 & 100.00 & +0.00 & 0.47  & 100.00 \\
        \midrule
        \rowcolor{blue!3} & T1 & 81.82 & 99.45  & 2.88 & 100.00 & +0.00 & 0.41  & 83.47  \\
        \rowcolor{blue!3} \multirow{-2}{*}{sft-800} & T2 & 81.14 & 98.63  & 2.88 & 100.00 & +0.00 & 0.33  & 100.00 \\
        \rowcolor{blue!3} & T3 & 81.82 & 99.45  & 2.88 & 100.00 & +0.00 & 0.46  & 133.33 \\
        \midrule
        \rowcolor{green!3}       & T1 & 55.00 & 101.89 & 2.31 & 101.32 & +0.03 & 21.60 & 112.32 \\
        \rowcolor{green!3} \multirow{-2}{*}{rl-40} & T2 & 53.98 & 100.00 & 2.27 & 99.56  & -0.01 & 19.88 & 103.32 \\
        \rowcolor{green!3}       & T3 & 54.32 & 100.63 & 2.29 & 100.44 & +0.01 & 6.68  & 94.84  \\
        \midrule
        \rowcolor{green!3}       & T1 & 73.41 & 99.69  & 2.54 & 98.83  & -0.03 & 19.48 & 151.50 \\
        \rowcolor{green!3} \multirow{-2}{*}{rl-80} & T2 & 77.50 & 105.24 & 2.58 & 100.39 & +0.01 & 16.46 & 127.98 \\
        \rowcolor{green!3}       & T3 & 73.98 & 100.46 & 2.57 & 100.00 & +0.00 & 2.20  & 112.15 \\
        \midrule
        \rowcolor{green!3}       & T1 & 81.93 & 100.13 & 2.64 & 98.51  & -0.04 & 17.93 & 111.43 \\
        \rowcolor{green!3} \multirow{-2}{*}{rl-160} & T2 & 81.70 & 99.85  & 2.65 & 98.88  & -0.03 & 15.63 & 97.41  \\
        \rowcolor{green!3}       & T3 & 80.91 & 98.89  & 2.66 & 99.25  & -0.02 & 0.11  & 50.11  \\
        \midrule
        \rowcolor{green!3}       & T1 & 91.59 & 100.12 & 2.69 & 100.37 & +0.01 & 9.96  & 99.13  \\
        \rowcolor{green!3} \multirow{-2}{*}{rl-240} & T2 & 92.05 & 100.62 & 2.68 & 100.00 & +0.00 & 9.90  & 99.03  \\
        \rowcolor{green!3}       & T3 & 91.25 & 99.75  & 2.68 & 100.00 & +0.00 & 0.00  & 100.00 \\
        \midrule
        \rowcolor{green!3}       & T1 & 91.59 & 100.25 & 2.69 & 100.37 & +0.01 & 9.96  & 100.89 \\
        \rowcolor{green!3} \multirow{-2}{*}{rl-280} & T2 & 92.05 & 100.76 & 2.68 & 100.00 & +0.00 & 9.73  & 99.01  \\
        \rowcolor{green!3}       & T3 & 91.36 & 100.00 & 2.68 & 100.00 & +0.00 & 0.00  & 100.00 \\
        \bottomrule
    \end{tabular}
    }
\end{table}

\begin{table}[H]
    \centering
    \caption{Detailed Results of symbolic Drift (Synonymous Rewriting) across three tools (\textbf{T1}: \texttt{calculate\_distance\_by\_coords} $\rightarrow$ \texttt{calculate\_separation\_via\_positions}; \textbf{T2}: \texttt{search\_map\_coordinates} $\rightarrow$ \texttt{lookup\_location\_latlong}; \textbf{T3}: \texttt{get\_adult\_cost\_by\_name} $\rightarrow$ \texttt{fetch\_grownup\_price\_using\_name}).}
    \label{tab:metrics_symbolic_full_colored}
    \resizebox{.9\linewidth}{!}{
    \begin{tabular}{lcccccccc}
        \toprule
        \rowcolor{gray!20} \textbf{Model/Step} & \textbf{Tool} & \textbf{Acc (Abs)} & \textbf{Acc Ratio} & \textbf{ATCL (Abs)} & \textbf{ATCL Ratio} & \textbf{ATCL Diff} & \textbf{TER} & \textbf{TER Ratio} \\
        \midrule
        & T1 & 27.61 & 94.17  & 3.76 & 101.35 & +0.05 & 11.10 & 110.76 \\
        base & T2 & 24.09 & 82.16  & 3.88 & 104.58 & +0.17 & 18.22 & 223.78 \\
        \multirow{-3}{*}{base} & T3 & 30.00 & 102.32 & 3.72 & 100.27 & +0.01 & 12.49 & 95.28  \\
        \midrule
        \rowcolor{blue!3} & T1 & 68.52 & 104.50 & 2.19 & 101.39 & +0.03 & 3.10  & 39.21  \\
        \rowcolor{blue!3} sft-200 & T2 & 57.39 & 87.52  & 2.23 & 103.24 & +0.07 & 16.65 & 263.32 \\
        \rowcolor{blue!3} \multirow{-3}{*}{sft-200} & T3 & 66.59 & 101.56 & 2.20 & 101.85 & +0.04 & 0.60  & 389.74 \\
        \midrule
        \rowcolor{blue!3} & T1 & 80.45 & 99.57  & 2.79 & 99.29  & -0.02 & 0.71  & 74.47  \\
        \rowcolor{blue!3} sft-400 & T2 & 72.61 & 89.86  & 2.82 & 100.36 & +0.01 & 8.67  & 1007.77 \\
        \rowcolor{blue!3} \multirow{-3}{*}{sft-400} & T3 & 71.25 & 88.18  & 2.92 & 103.91 & +0.11 & 10.85 & 765.61 \\
        \midrule
        \rowcolor{blue!3} & T1 & 80.45 & 98.60  & 2.86 & 100.00 & 0.00  & 0.47  & 100.00 \\
        \rowcolor{blue!3} sft-600 & T2 & 76.48 & 93.74  & 2.88 & 100.70 & +0.02 & 6.02  & 799.70 \\
        \rowcolor{blue!3} \multirow{-3}{*}{sft-600} & T3 & 77.50 & 94.99  & 2.86 & 100.00 & 0.00  & 5.27  & 714.63 \\
        \midrule
        \rowcolor{blue!3} & T1 & 81.59 & 99.17  & 2.88 & 100.00 & 0.00  & 0.70  & 200.70 \\
        \rowcolor{blue!3} sft-800 & T2 & 75.45 & 91.71  & 2.86 & 99.31  & -0.02 & 5.16  & 787.93 \\
        \rowcolor{blue!3} \multirow{-3}{*}{sft-800} & T3 & 78.52 & 95.44  & 2.88 & 100.00 & 0.00  & 4.81  & 980.13 \\
        \midrule
        \rowcolor{green!3} & T1 & 53.07 & 98.31  & 2.26 & 99.12  & -0.02 & 7.17  & 101.77 \\
        \rowcolor{green!3} rl-40 & T2 & 53.52 & 99.15  & 2.33 & 102.19 & +0.05 & 5.06  & 91.35  \\
        \rowcolor{green!3} \multirow{-3}{*}{rl-40} & T3 & 53.64 & 99.37  & 2.41 & 105.70 & +0.13 & 14.38 & 74.76  \\
        \midrule
        \rowcolor{green!3} & T1 & 70.91 & 96.29  & 2.56 & 99.61  & -0.01 & 6.89  & 350.51 \\
        \rowcolor{green!3} rl-80 & T2 & 75.45 & 102.46 & 2.55 & 99.22  & -0.02 & 0.19  & 11.69  \\
        \rowcolor{green!3} \multirow{-3}{*}{rl-80} & T3 & 68.98 & 93.67  & 2.55 & 99.22  & -0.02 & 12.93 & 100.56 \\
        \midrule
        \rowcolor{green!3} & T1 & 79.89 & 97.64  & 2.65 & 98.88  & -0.03 & 0.34  & 150.34 \\
        \rowcolor{green!3} rl-160 & T2 & 79.89 & 97.64  & 2.63 & 98.13  & -0.05 & 0.19  & 100.84 \\
        \rowcolor{green!3} \multirow{-3}{*}{rl-160} & T3 & 80.57 & 98.47  & 2.71 & 101.12 & +0.03 & 14.49 & 90.02  \\
        \midrule
        \rowcolor{green!3} & T1 & 91.48 & 100.00 & 2.68 & 100.00 & 0.00  & 0.00  & 100.00 \\
        \rowcolor{green!3} rl-240 & T2 & 91.59 & 100.12 & 2.67 & 99.63  & -0.01 & 0.00  & 100.00 \\
        \rowcolor{green!3} \multirow{-3}{*}{rl-240} & T3 & 87.84 & 96.02  & 2.65 & 98.88  & -0.03 & 10.63 & 105.74 \\
        \midrule
        \rowcolor{green!3} & T1 & 91.36 & 100.00 & 2.68 & 100.00 & 0.00  & 0.00  & 100.00 \\
        \rowcolor{green!3} rl-280 & T2 & 91.70 & 100.37 & 2.66 & 99.25  & -0.02 & 0.00  & 100.00 \\
        \rowcolor{green!3} \multirow{-3}{*}{rl-280} & T3 & 87.73 & 96.03  & 2.65 & 98.88  & -0.03 & 10.55 & 106.83 \\
        \bottomrule
    \end{tabular}
    }
\end{table}

\begin{table}[H]
    \centering
    \caption{ Detailed Results of symbolic Drift (Noise Injection) across three tools 
    (\textbf{T1}: \texttt{calculate\_distance\_by\_coords} $\rightarrow$ \texttt{calculate\_the\_distance\_by\_coords}; \textbf{T2}: \texttt{search\_map\_coordinates} $\rightarrow$ \texttt{search\_for\_map\_coordinates}; \textbf{T3}: \texttt{get\_adult\_cost\_by\_name} $\rightarrow$ \texttt{get\_1\_adult\_cost\_by\_name})}
    \label{tab:noise_injection_merged}
    \resizebox{.9\linewidth}{!}{
    \begin{tabular}{lcccccccc}
        \toprule
        \rowcolor{gray!20} \textbf{Model/Step} & \textbf{Tool} & \textbf{Acc (Abs)} & \textbf{Acc Ratio} & \textbf{ATCL (Abs)} & \textbf{ATCL Ratio} & \textbf{ATCL Diff} & \textbf{TER Rate-Sub} & \textbf{TER Ratio} \\
        \midrule
        & T1 & 28.30 & 96.52  & 3.80 & 102.43 & +0.09 & 9.77 & 97.46 \\
        base & T2 & 30.34 & 103.48 & 3.77 & 101.62 & +0.06 & 8.01 & 98.34 \\
        \multirow{-3}{*}{base} & T3 & 29.20 & 99.59  & 3.81 & 102.70 & +0.10 & 11.31 & 86.26 \\
        \midrule
        \rowcolor{blue!3} & T1 & 68.52 & 104.50 & 2.20 & 101.85 & +0.04 & 4.26 & 53.79 \\
        \rowcolor{blue!3} sft-200 & T2 & 66.36 & 101.20 & 2.17 & 100.46 & +0.01 & 5.68 & 89.77 \\
        \rowcolor{blue!3} \multirow{-3}{*}{sft-200} & T3 & 63.98 & 97.58  & 2.29 & 106.02 & +0.13 & 19.86 & 12830.28 \\
        \midrule
        \rowcolor{blue!3} & T1 & 80.57 & 99.72  & 2.79 & 99.29  & -0.02 & 1.07 & 112.90 \\
        \rowcolor{blue!3} sft-400 & T2 & 80.45 & 99.57  & 2.79 & 99.29  & -0.02 & 0.86 & 99.81 \\
        \rowcolor{blue!3} \multirow{-3}{*}{sft-400} & T3 & 58.64 & 72.57  & 2.82 & 100.36 & +0.01 & 24.05 & 1697.62 \\
        \midrule
        \rowcolor{blue!3} & T1 & 81.02 & 99.30  & 2.86 & 100.00 & +0.00 & 0.23 & 50.12 \\
        \rowcolor{blue!3} sft-600 & T2 & 82.16 & 100.70 & 2.86 & 100.00 & +0.00 & 0.56 & 74.72 \\
        \rowcolor{blue!3} \multirow{-3}{*}{sft-600} & T3 & 64.43 & 78.97  & 2.87 & 100.35 & +0.01 & 21.01 & 2848.55 \\
        \midrule
        \rowcolor{blue!3} & T1 & 81.70 & 99.31  & 2.88 & 100.00 & +0.00 & 0.35 & 100.35 \\
        \rowcolor{blue!3} sft-800 & T2 & 81.59 & 99.17  & 2.87 & 99.65  & -0.01 & 0.38 & 57.36 \\
        \rowcolor{blue!3} \multirow{-3}{*}{sft-800} & T3 & 66.93 & 81.35  & 2.88 & 100.00 & +0.00 & 17.28 & 3518.94 \\
        \midrule
        \rowcolor{green!3} & T1 & 54.09 & 100.20 & 2.28 & 100.00 & +0.00 & 6.73 & 95.48 \\
        \rowcolor{green!3} rl-40 & T2 & 54.20 & 100.41 & 2.27 & 99.56  & -0.01 & 5.39 & 97.35 \\
        \rowcolor{green!3} \multirow{-3}{*}{rl-40} & T3 & 53.75 & 99.57  & 2.35 & 103.07 & +0.07 & 15.73 & 81.76 \\
        \midrule
        \rowcolor{green!3} & T1 & 72.50 & 98.45  & 2.52 & 98.05  & -0.05 & 1.07 & 54.39 \\
        \rowcolor{green!3} rl-80 & T2 & 69.20 & 93.97  & 2.48 & 96.50  & -0.09 & 2.49 & 155.84 \\
        \rowcolor{green!3} \multirow{-3}{*}{rl-80} & T3 & 70.80 & 96.14  & 2.50 & 97.28  & -0.07 & 15.34 & 119.31 \\
        \midrule
        \rowcolor{green!3} & T1 & 80.68 & 98.61  & 2.66 & 99.25  & -0.02 & 0.35 & 151.04 \\
        \rowcolor{green!3} rl-160 & T2 & 78.07 & 95.42  & 2.64 & 98.51  & -0.04 & 0.37 & 201.50 \\
        \rowcolor{green!3} \multirow{-3}{*}{rl-160} & T3 & 82.27 & 100.55 & 2.68 & 100.00 & +0.00 & 12.62 & 78.42 \\
        \midrule
        \rowcolor{green!3} & T1 & 91.25 & 99.75  & 2.68 & 100.00 & +0.00 & 0.00 & 100.00 \\
        \rowcolor{green!3} rl-240 & T2 & 89.55 & 97.89  & 2.65 & 98.88  & -0.03 & 0.00 & 100.00 \\
        \rowcolor{green!3} \multirow{-3}{*}{rl-240} & T3 & 90.91 & 99.38  & 2.69 & 100.37 & +0.01 & 10.37 & 103.18 \\
        \midrule
        \rowcolor{green!3} & T1 & 91.14 & 99.76  & 2.68 & 100.00 & +0.00 & 0.00 & 100.00 \\
        \rowcolor{green!3} rl-280 & T2 & 89.77 & 98.26  & 2.65 & 98.88  & -0.03 & 0.00 & 100.00 \\
        \rowcolor{green!3} \multirow{-3}{*}{rl-280} & T3 & 90.91 & 99.51  & 2.70 & 100.75 & +0.02 & 9.97 & 100.99 \\
        \bottomrule
    \end{tabular}
    }
\end{table}

\begin{table}[H]
    \centering
    \caption{Detailed Results of Semantic Adversarial Experiments Across Three Tools (\textbf{T1}: \texttt{calculate\_distance\_by\_coords} $\rightarrow$ \texttt{calculate\_separation\_via\_positions}; \textbf{T2}: \texttt{search\_map\_coordinates} $\rightarrow$ \texttt{lookup\_location\_latlong}; \textbf{T3}: \texttt{get\_adult\_cost\_by\_name} $\rightarrow$ \texttt{fetch\_grownup\_price\_using\_name})}
    \label{tab:semantic_adversarial_merged}
    \resizebox{.9\linewidth}{!}{
    \begin{tabular}{lcccccccc}
        \toprule
        \rowcolor{gray!20} \textbf{Model/Step} & \textbf{Tool} & \textbf{Acc (Abs)} & \textbf{Acc Ratio} & \textbf{ATCL (Abs)} & \textbf{ATCL Ratio} & \textbf{ATCL Diff} & \textbf{Old Tool Call} \\
        \midrule
        & T1 & 22.84 & 77.90 & 3.69 & 99.46 & -0.02 & 7.90 \\
        base & T2 & 19.09 & 65.11 & 3.78 & 101.89 & +0.07 & 1.71 \\
        \multirow{-3}{*}{base} & T3 & 20.57 & 70.16 & 3.79 & 102.16 & +0.08 & 1.60 \\
        \midrule
        \rowcolor{blue!3} & T1 & 51.02 & 77.81 & 1.75 & 81.02 & -0.41 & 16.14 \\
        \rowcolor{blue!3} sft-200 & T2 & 32.05 & 48.88 & 1.51 & 69.91 & -0.65 & 18.61 \\
        \rowcolor{blue!3} \multirow{-3}{*}{sft-200} & T3 & 21.14 & 32.24 & 1.56 & 72.22 & -0.60 & 37.91 \\
        \midrule
        \rowcolor{blue!3} & T1 & 49.43 & 61.18 & 2.64 & 93.95 & -0.17 & 24.38 \\
        \rowcolor{blue!3} sft-400 & T2 & 54.55 & 67.51 & 2.55 & 90.75 & -0.26 & 7.28 \\
        \rowcolor{blue!3} \multirow{-3}{*}{sft-400} & T3 & 37.27 & 46.13 & 2.55 & 90.75 & -0.26 & 24.19 \\
        \midrule
        \rowcolor{blue!3} & T1 & 51.70 & 63.37 & 2.79 & 97.55 & -0.07 & 22.70 \\
        \rowcolor{blue!3} sft-600 & T2 & 56.59 & 69.36 & 2.73 & 95.45 & -0.13 & 7.90 \\
        \rowcolor{blue!3} \multirow{-3}{*}{sft-600} & T3 & 39.77 & 48.74 & 2.76 & 96.50 & -0.10 & 23.79 \\
        \midrule
        \rowcolor{blue!3} & T1 & 51.59 & 62.71 & 2.80 & 97.22 & -0.08 & 22.01 \\
        \rowcolor{blue!3} sft-800 & T2 & 57.05 & 69.34 & 2.77 & 96.18 & -0.11 & 7.62 \\
        \rowcolor{blue!3} \multirow{-3}{*}{sft-800} & T3 & 40.80 & 49.59 & 2.80 & 97.22 & -0.08 & 23.35 \\
        \midrule
        \rowcolor{green!3} & T1 & 49.89 & 92.42 & 2.31 & 101.32 & +0.03 & 8.41 \\
        \rowcolor{green!3} rl-40 & T2 & 50.45 & 93.46 & 2.31 & 101.32 & +0.03 & 0.00 \\
        \rowcolor{green!3} \multirow{-3}{*}{rl-40} & T3 & 44.20 & 81.88 & 2.28 & 100.00 & +0.00 & 0.95 \\
        \midrule
        \rowcolor{green!3} & T1 & 59.09 & 80.24 & 2.60 & 101.17 & +0.03 & 8.57 \\
        \rowcolor{green!3} rl-80 & T2 & 70.00 & 95.06 & 2.47 & 96.11 & -0.10 & 0.09 \\
        \rowcolor{green!3} \multirow{-3}{*}{rl-80} & T3 & 60.57 & 82.25 & 2.46 & 95.72 & -0.11 & 2.21 \\
        \midrule
        \rowcolor{green!3} & T1 & 62.39 & 76.25 & 2.69 & 100.37 & +0.01 & 7.52 \\
        \rowcolor{green!3} rl-160 & T2 & 71.82 & 87.78 & 2.52 & 94.03 & -0.16 & 0.14 \\
        \rowcolor{green!3} \multirow{-3}{*}{rl-160} & T3 & 63.64 & 77.78 & 2.56 & 95.52 & -0.12 & 4.21 \\
        \midrule
        \rowcolor{green!3} & T1 & 67.95 & 74.28 & 2.73 & 101.87 & +0.05 & 6.41 \\
        \rowcolor{green!3} rl-240 & T2 & 73.86 & 80.74 & 2.45 & 91.42 & -0.23 & 0.14 \\
        \rowcolor{green!3} \multirow{-3}{*}{rl-240} & T3 & 64.43 & 70.43 & 2.42 & 90.30 & -0.26 & 4.60 \\
        \midrule
        \rowcolor{green!3} & T1 & 68.18 & 74.63 & 2.73 & 101.87 & +0.05 & 6.36 \\
        \rowcolor{green!3} rl-280 & T2 & 73.52 & 80.47 & 2.45 & 91.42 & -0.23 & 0.14 \\
        \rowcolor{green!3} \multirow{-3}{*}{rl-280} & T3 & 64.32 & 70.40 & 2.41 & 89.93 & -0.27 & 4.76 \\
        \bottomrule
    \end{tabular}
    }
\end{table}

\begin{table}[H]
    \centering
    \caption{Detailed Results of Identity Erasure Experiments across Three Tools (\textbf{T1}: \texttt{calculate\_distance\_by\_coords} $\rightarrow$ \texttt{Tool\_a2fb}; \textbf{T2}: \texttt{search\_map\_coordinates} $\rightarrow$ \texttt{Tool\_c3d8}; \textbf{T3}: \texttt{get\_adult\_cost\_by\_name} $\rightarrow$ \texttt{Tool\_r6t7})}
    \label{tab:detailed_results_three_tools}
    \resizebox{.9\linewidth}{!}{
    \begin{tabular}{lcccccccc}
        \toprule
        \rowcolor{gray!20} \textbf{Model/Step} & \textbf{Tool} & \textbf{Acc (Abs)} & \textbf{Acc Ratio} & \textbf{ATCL (Abs)} & \textbf{ATCL Ratio} & \textbf{ATCL Diff} & \textbf{TER} & \textbf{TER Ratio} \\
        \midrule
        & T1 & 22.50 & 76.74 & 3.82 & 102.96 & +0.11 & 19.13 & 136.39 \\
        base & T2 & 23.18 & 79.06 & 3.64 & 98.11  & -0.07 & 20.74 & 154.37 \\
        \multirow{-3}{*}{base} & T3 & 21.59 & 73.64 & 3.63 & 97.84  & -0.08 & 17.87 & 167.15 \\
        \midrule
        \rowcolor{blue!3} & T1 & 44.89 & 68.46 & 2.40 & 111.11 & +0.24 & 13.83 & 403.36 \\
        \rowcolor{blue!3} sft-200 & T2 & 41.70 & 63.60 & 2.60 & 120.37 & +0.44 & 12.60 & 360.30 \\
        \rowcolor{blue!3} \multirow{-3}{*}{sft-200} & T3 & 59.20 & 90.29 & 2.46 & 113.89 & +0.30 & 10.42 & 11489.26 \\
        \midrule
        \rowcolor{blue!3} & T1 & 57.16 & 70.74 & 2.87 & 102.14 & +0.06 & 11.71 & 3252.43 \\
        \rowcolor{blue!3} sft-400 & T2 & 57.39 & 71.03 & 2.94 & 104.63 & +0.13 & 10.93 & 2676.69 \\
        \rowcolor{blue!3} \multirow{-3}{*}{sft-400} & T3 & 60.00 & 74.26 & 2.88 & 102.49 & +0.07 & 9.08  & 1251.29 \\
        \midrule
        \rowcolor{blue!3} & T1 & 57.16 & 70.06 & 2.88 & 100.70 & +0.02 & 10.51 & 6135.49 \\
        \rowcolor{blue!3} sft-600 & T2 & 65.00 & 79.67 & 2.87 & 100.35 & +0.01 & 7.20  & 2200.14 \\
        \rowcolor{blue!3} \multirow{-3}{*}{sft-600} & T3 & 65.68 & 80.50 & 2.87 & 100.35 & +0.01 & 6.96  & 1849.99 \\
        \midrule
        \rowcolor{blue!3} & T1 & 60.23 & 73.21 & 2.90 & 100.69 & +0.02 & 9.45  & 7443.79 \\
        \rowcolor{blue!3} sft-800 & T2 & 68.41 & 83.15 & 2.88 & 100.00 & +0.00 & 5.62  & 1933.63 \\
        \rowcolor{blue!3} \multirow{-3}{*}{sft-800} & T3 & 66.82 & 81.22 & 2.89 & 100.35 & +0.01 & 6.84  & 2700.19 \\
        \midrule
        \rowcolor{green!3} & T1 & 54.32 & 100.63 & 2.42 & 106.14 & +0.14 & 12.27 & 93.04 \\
        \rowcolor{green!3} rl-40 & T2 & 51.25 & 94.94  & 2.40 & 105.26 & +0.12 & 13.90 & 96.66 \\
        \rowcolor{green!3} \multirow{-3}{*}{rl-40} & T3 & 52.84 & 97.89  & 2.29 & 100.44 & +0.01 & 9.50  & 92.72 \\
        \midrule
        \rowcolor{green!3} & T1 & 65.23 & 88.58  & 2.50 & 97.28  & -0.07 & 10.75 & 137.18 \\
        \rowcolor{green!3} rl-80 & T2 & 66.82 & 90.74  & 2.51 & 97.67  & -0.06 & 11.32 & 24.26 \\
        \rowcolor{green!3} \multirow{-3}{*}{rl-80} & T3 & 62.50 & 84.87  & 2.46 & 95.72  & -0.11 & 7.06  & 121.13 \\
        \midrule
        \rowcolor{green!3} & T1 & 75.80 & 92.64  & 2.67 & 99.63  & -0.01 & 8.12  & 102.58 \\
        \rowcolor{green!3} rl-160 & T2 & 75.45 & 92.21  & 2.66 & 99.25  & -0.02 & 8.46  & 151.26 \\
        \rowcolor{green!3} \multirow{-3}{*}{rl-160} & T3 & 72.73 & 88.89  & 2.75 & 102.61 & +0.07 & 5.28  & 68.66 \\
        \midrule
        \rowcolor{green!3} & T1 & 87.84 & 96.02  & 2.63 & 98.13  & -0.05 & 5.25  & 1150.75 \\
        \rowcolor{green!3} rl-240 & T2 & 91.02 & 99.50  & 2.67 & 99.63  & -0.01 & 4.55  & 100.00 \\
        \rowcolor{green!3} \multirow{-3}{*}{rl-240} & T3 & 79.43 & 86.83  & 2.65 & 98.88  & -0.03 & 6.23  & 140.43 \\
        \midrule
        \rowcolor{green!3} & T1 & 87.84 & 96.15  & 2.62 & 97.76  & -0.06 & 5.31  & 1150.75 \\
        \rowcolor{green!3} rl-280 & T2 & 91.02 & 99.63  & 2.67 & 99.63  & -0.01 & 4.55  & 100.00 \\
        \rowcolor{green!3} \multirow{-3}{*}{rl-280} & T3 & 79.20 & 86.69  & 2.64 & 98.51  & -0.04 & 6.25  & 144.10 \\
        \bottomrule
    \end{tabular}
    }
\end{table}

\subsection{Tier 2 Experiments}
We report the full numerical results for all Tier-2 interaction experiments.
Table~\ref{tab:return_format} shows that both paradigms are robust to simple format changes.
Tables~\ref{tab:baseline_solvable_merged}--\ref{tab:tool_error_merged} present per-tool results under Null Return and Error Return, where we additionally report the AES score to quantify the agent's active exploration after encountering failures.
Tables~\ref{tab:alternative_tool_merged}--\ref{tab:parameter_change_merged} detail Tool Redirection and Value Redirection results, where the gap between SFT and RL is most pronounced: RL maintains near-perfect AES scores ($>$0.86) while SFT scores collapse to near zero.

\begin{table}[H]
    \centering
    \caption{Detailed Results of Return Format Variation}
    \label{tab:return_format}
    \begin{tabularx}{.9\linewidth}{lYYYYY}
        \toprule
        \rowcolor{gray!20} \textbf{Model/Step} & \textbf{Acc (Abs)} & \textbf{Acc Ratio} & \textbf{Tool (Abs)} & \textbf{Tool Ratio} & \textbf{Tool Diff} \\ 
        \midrule
        base & 29.32 & 100.00 & 3.83 & 103.23 & +0.12 \\ 
        \midrule
        \rowcolor{blue!3} sft-200 & 74.77 & 114.03 & 2.53 & 117.13 & +0.37 \\ 
        \rowcolor{blue!3} sft-400 & 80.68 & 99.85  & 2.84 & 101.07 & +0.03 \\ 
        \rowcolor{blue!3} sft-600 & 82.05 & 100.56 & 2.86 & 100.00 & +0.00 \\ 
        \rowcolor{blue!3} sft-800 & 82.16 & 99.87  & 2.86 & 99.31  & -0.02 \\ 
        \midrule
        \rowcolor{green!3} rl-40  & 57.16 & 105.89 & 2.40 & 105.26 & +0.12 \\ 
        \rowcolor{green!3} rl-80  & 74.55 & 101.24 & 2.62 & 101.95 & +0.05 \\ 
        \rowcolor{green!3} rl-160 & 81.93 & 100.13 & 2.73 & 101.87 & +0.05 \\ 
        \rowcolor{green!3} rl-240 & 90.91 & 99.38  & 2.71 & 101.12 & +0.03 \\ 
        \rowcolor{green!3} rl-280 & 90.45 & 99.00  & 2.74 & 102.24 & +0.06 \\ 
        \bottomrule
    \end{tabularx}
\end{table}

\begin{table}[H]
    \centering
    \caption{Detailed Baseline Results of Solvable Problems Across Three Tools (\textbf{T1}: \texttt{calculate\_distance\_by\_coords}; \textbf{T2}: \texttt{search\_map\_coordinates}; \textbf{T3}: \texttt{get\_adult\_cost\_by\_name})}
    \label{tab:baseline_solvable_merged}
    \resizebox{.9\linewidth}{!}{
    \begin{tabular}{lcccccccc}
        \toprule
        \rowcolor{gray!20} \textbf{Model/Step} & \textbf{Tool} & \textbf{Acc (Abs)} & \textbf{Acc Ratio} & \textbf{ATCL (Abs)} & \textbf{ATCL Ratio} & \textbf{ATCL Diff} & \textbf{New Tool Call} \\
        \midrule
        & T1 & 22.50 & 76.74 & 3.88 & 104.58 & +0.17 & 2.66 \\
        base & T2 & 25.80 & 87.99 & 3.90 & 105.12 & +0.19 & 5.92 \\
        \multirow{-3}{*}{base} & T3 & 25.00 & 85.27 & 3.90 & 105.12 & +0.19 & 13.73 \\
        \midrule
        \rowcolor{blue!3} & T1 & 68.18 & 103.98 & 2.16 & 100.00 & +0.00 & 1.26 \\
        \rowcolor{blue!3} sft-200 & T2 & 62.73 & 95.67 & 1.86 & 86.11  & -0.30 & 0.31 \\
        \rowcolor{blue!3} \multirow{-3}{*}{sft-200} & T3 & 62.61 & 95.49 & 2.04 & 94.44  & -0.12 & 1.11 \\
        \midrule
        \rowcolor{blue!3} & T1 & 80.23 & 99.29 & 2.79 & 99.29 & -0.02 & 1.06 \\
        \rowcolor{blue!3} sft-400 & T2 & 80.34 & 99.43 & 2.77 & 98.58 & -0.04 & 4.71 \\
        \rowcolor{blue!3} \multirow{-3}{*}{sft-400} & T3 & 78.75 & 97.46 & 2.77 & 98.58 & -0.04 & 0.90 \\
        \midrule
        \rowcolor{blue!3} & T1 & 80.57 & 98.75 & 2.86 & 100.00 & +0.00 & 0.08 \\
        \rowcolor{blue!3} sft-600 & T2 & 79.77 & 97.77 & 2.85 & 99.65  & -0.01 & 2.43 \\
        \rowcolor{blue!3} \multirow{-3}{*}{sft-600} & T3 & 81.36 & 99.72 & 2.86 & 100.00 & +0.00 & 0.91 \\
        \midrule
        \rowcolor{blue!3} & T1 & 81.14 & 98.63 & 2.87 & 99.65 & -0.01 & 0.04 \\
        \rowcolor{blue!3} sft-800 & T2 & 82.16 & 99.87 & 2.85 & 98.96 & -0.03 & 2.59 \\
        \rowcolor{blue!3} \multirow{-3}{*}{sft-800} & T3 & 82.16 & 99.87 & 2.88 & 100.00 & +0.00 & 1.06 \\
        \midrule
        \rowcolor{green!3} & T1 & 51.93 & 96.20 & 2.42 & 106.14 & +0.14 & 3.85 \\
        \rowcolor{green!3} rl-40 & T2 & 52.05 & 96.42 & 2.38 & 104.39 & +0.10 & 4.39 \\
        \rowcolor{green!3} \multirow{-3}{*}{rl-40} & T3 & 48.75 & 90.31 & 2.34 & 102.63 & +0.06 & 9.32 \\
        \midrule
        \rowcolor{green!3} & T1 & 71.70 & 97.37 & 2.74 & 106.61 & +0.17 & 5.60 \\
        \rowcolor{green!3} rl-80 & T2 & 66.02 & 89.65 & 2.44 & 94.94  & -0.13 & 5.35 \\
        \rowcolor{green!3} \multirow{-3}{*}{rl-80} & T3 & 63.75 & 86.57 & 2.54 & 98.83  & -0.03 & 10.45 \\
        \midrule
        \rowcolor{green!3} & T1 & 82.27 & 100.55 & 2.84 & 105.97 & +0.16 & 7.59 \\
        \rowcolor{green!3} rl-160 & T2 & 74.43 & 90.97  & 2.61 & 97.39  & -0.07 & 3.89 \\
        \rowcolor{green!3} \multirow{-3}{*}{rl-160} & T3 & 70.34 & 85.97  & 2.76 & 102.99 & +0.08 & 10.42 \\
        \midrule
        \rowcolor{green!3} & T1 & 89.43 & 97.76 & 2.85 & 106.34 & +0.17 & 5.58 \\
        \rowcolor{green!3} rl-240 & T2 & 90.00 & 98.38 & 2.69 & 100.37 & +0.01 & 4.77 \\
        \rowcolor{green!3} \multirow{-3}{*}{rl-240} & T3 & 88.75 & 97.02 & 2.70 & 100.75 & +0.02 & 4.34 \\
        \midrule
        \rowcolor{green!3} & T1 & 88.86 & 97.26 & 2.85 & 106.34 & +0.17 & 5.63 \\
        \rowcolor{green!3} rl-280 & T2 & 90.11 & 98.63 & 2.69 & 100.37 & +0.01 & 4.90 \\
        \rowcolor{green!3} \multirow{-3}{*}{rl-280} & T3 & 88.52 & 96.89 & 2.69 & 100.37 & +0.01 & 4.47 \\
        \bottomrule
    \end{tabular}
    }
\end{table}

\begin{table}[H]
    \centering
    \caption{Detailed Results of Solvable Problems with Null Return Across Three Tools (\textbf{T1}: \texttt{calculate\_distance\_by\_coords}; \textbf{T2}: \texttt{search\_map\_coordinates}; \textbf{T3}: \texttt{get\_adult\_cost\_by\_name})}
    \label{tab:no_return_value_merged}
    \resizebox{.9\linewidth}{!}{
    \begin{tabular}{lcccccccc}
        \toprule
        \rowcolor{gray!20} \textbf{Model/Step} & \textbf{Tool} & \textbf{Acc (Abs)} & \textbf{Acc Ratio} & \textbf{ATCL (Abs)} & \textbf{ATCL Ratio} & \textbf{ATCL Diff} & \textbf{New Tool Call} \\
        \midrule
        & T1 & 20.57 & 70.16 & 4.14 & 111.59 & +0.43 & 3.13 \\
        base & T2 & 20.68 & 70.53 & 4.20 & 113.21 & +0.49 & 8.98 \\
        \multirow{-3}{*}{base} & T3 & 18.07 & 61.63 & 3.85 & 103.77 & +0.14 & 13.48 \\
        \midrule
        \rowcolor{blue!3} & T1 & 58.07 & 88.56 & 2.40 & 111.11 & +0.24 & 1.33 \\
        \rowcolor{blue!3} sft-200 & T2 & 43.07 & 65.69 & 2.08 & 96.30  & -0.08 & 0.22 \\
        \rowcolor{blue!3} \multirow{-3}{*}{sft-200} & T3 & 30.45 & 46.44 & 2.11 & 97.69  & -0.05 & 0.91 \\
        \midrule
        \rowcolor{blue!3} & T1 & 55.57 & 68.77 & 2.82 & 100.36 & +0.01 & 1.05 \\
        \rowcolor{blue!3} sft-400 & T2 & 63.98 & 79.18 & 2.81 & 100.00 & +0.00 & 4.37 \\
        \rowcolor{blue!3} \multirow{-3}{*}{sft-400} & T3 & 45.23 & 55.98 & 2.81 & 100.00 & +0.00 & 0.81 \\
        \midrule
        \rowcolor{blue!3} & T1 & 57.39 & 70.34 & 2.86 & 100.00 & +0.00 & 0.04 \\
        \rowcolor{blue!3} sft-600 & T2 & 59.89 & 73.40 & 2.85 & 99.65  & -0.01 & 2.39 \\
        \rowcolor{blue!3} \multirow{-3}{*}{sft-600} & T3 & 44.66 & 54.74 & 2.88 & 100.70 & +0.02 & 0.91 \\
        \midrule
        \rowcolor{blue!3} & T1 & 57.73 & 70.17 & 2.87 & 99.65  & -0.01 & 0.04 \\
        \rowcolor{blue!3} sft-800 & T2 & 62.16 & 75.56 & 2.85 & 98.96  & -0.03 & 2.47 \\
        \rowcolor{blue!3} \multirow{-3}{*}{sft-800} & T3 & 45.80 & 55.67 & 2.88 & 100.00 & +0.00 & 1.10 \\
        \midrule
        \rowcolor{green!3} & T1 & 49.32 & 91.37 & 2.68 & 117.54 & +0.40 & 3.27 \\
        \rowcolor{green!3} rl-40 & T2 & 37.84 & 70.10 & 2.45 & 107.46 & +0.17 & 4.58 \\
        \rowcolor{green!3} \multirow{-3}{*}{rl-40} & T3 & 26.25 & 48.63 & 2.58 & 113.16 & +0.30 & 10.38 \\
        \midrule
        \rowcolor{green!3} & T1 & 55.80 & 75.77 & 2.83 & 110.12 & +0.26 & 5.58 \\
        \rowcolor{green!3} rl-80 & T2 & 49.89 & 67.75 & 2.64 & 102.72 & +0.07 & 4.57 \\
        \rowcolor{green!3} \multirow{-3}{*}{rl-80} & T3 & 40.00 & 54.32 & 2.75 & 107.00 & +0.18 & 10.26 \\
        \midrule
        \rowcolor{green!3} & T1 & 62.95 & 76.94 & 2.91 & 108.58 & +0.23 & 6.84 \\
        \rowcolor{green!3} rl-160 & T2 & 55.34 & 67.64 & 2.77 & 103.36 & +0.09 & 4.68 \\
        \rowcolor{green!3} \multirow{-3}{*}{rl-160} & T3 & 40.80 & 49.87 & 2.73 & 101.87 & +0.05 & 11.13 \\
        \midrule
        \rowcolor{green!3} & T1 & 69.20 & 75.64 & 2.87 & 107.09 & +0.19 & 6.53 \\
        \rowcolor{green!3} rl-240 & T2 & 67.61 & 73.91 & 2.76 & 102.99 & +0.08 & 5.71 \\
        \rowcolor{green!3} \multirow{-3}{*}{rl-240} & T3 & 50.34 & 55.03 & 2.53 & 94.40  & -0.15 & 7.60 \\
        \midrule
        \rowcolor{green!3} & T1 & 69.20 & 75.74 & 2.87 & 107.09 & +0.19 & 6.37 \\
        \rowcolor{green!3} rl-280 & T2 & 68.30 & 74.76 & 2.77 & 103.36 & +0.09 & 6.16 \\
        \rowcolor{green!3} \multirow{-3}{*}{rl-280} & T3 & 50.00 & 54.73 & 2.51 & 93.66  & -0.17 & 7.45 \\
        \bottomrule
    \end{tabular}
    }
\end{table}

\begin{table}[H]
    \centering
    \caption{Detailed Results of Solvable Problems with Error Return Across Three Tools (\textbf{T1}: \texttt{calculate\_distance\_by\_coords}; \textbf{T2}: \texttt{search\_map\_coordinates}; \textbf{T3}: \texttt{get\_adult\_cost\_by\_name})}
    \label{tab:tool_error_merged}
    \resizebox{.9\linewidth}{!}{
    \begin{tabular}{lcccccccc}
        \toprule
        \rowcolor{gray!20} \textbf{Model/Step} & \textbf{Tool} & \textbf{Acc (Abs)} & \textbf{Acc Ratio} & \textbf{Tool (Abs)} & \textbf{Tool Ratio} & \textbf{Tool Diff} & \textbf{AES Score (Abs)} \\
        \midrule
        & T1 & 20.68 & 70.53 & 4.00 & 107.82 & +0.29 & 0.3325 \\
        base & T2 & 25.34 & 86.43 & 4.06 & 109.43 & +0.35 & 0.7990 \\
        \multirow{-3}{*}{base} & T3 & 20.68 & 70.53 & 3.93 & 105.93 & +0.22 & 0.5779 \\
        \midrule
        \rowcolor{blue!3} & T1 & 56.70 & 86.47 & 2.21 & 102.31 & +0.05 & 0.0156 \\
        \rowcolor{blue!3} sft-200 & T2 & 42.05 & 64.13 & 1.92 & 88.89  & -0.24 & 0.0148 \\
        \rowcolor{blue!3} \multirow{-3}{*}{sft-200} & T3 & 30.23 & 46.10 & 2.06 & 95.37  & -0.10 & 0.0351 \\
        \midrule
        \rowcolor{blue!3} & T1 & 55.11 & 68.21 & 2.81 & 100.00 & +0.00 & 0.0406 \\
        \rowcolor{blue!3} sft-400 & T2 & 64.89 & 80.31 & 2.79 & 99.29  & -0.02 & 0.3630 \\
        \rowcolor{blue!3} \multirow{-3}{*}{sft-400} & T3 & 44.55 & 55.14 & 2.79 & 99.29  & -0.02 & 0.0297 \\
        \midrule
        \rowcolor{blue!3} & T1 & 58.30 & 71.45 & 2.88 & 100.70 & +0.02 & 0.0063 \\
        \rowcolor{blue!3} sft-600 & T2 & 60.68 & 74.37 & 2.87 & 100.35 & +0.01 & 0.2000 \\
        \rowcolor{blue!3} \multirow{-3}{*}{sft-600} & T3 & 45.23 & 55.44 & 2.88 & 100.70 & +0.02 & 0.0270 \\
        \midrule
        \rowcolor{blue!3} & T1 & 58.75 & 71.41 & 2.90 & 100.69 & +0.02 & 0.0063 \\
        \rowcolor{blue!3} sft-800 & T2 & 61.48 & 74.73 & 2.86 & 99.31  & -0.02 & 0.2074 \\
        \rowcolor{blue!3} \multirow{-3}{*}{sft-800} & T3 & 45.68 & 55.52 & 2.89 & 100.35 & +0.01 & 0.0351 \\
        \midrule
        \rowcolor{green!3} & T1 & 49.55 & 91.79 & 2.49 & 109.21 & +0.21 & 0.3375 \\
        \rowcolor{green!3} rl-40 & T2 & 43.18 & 79.99 & 2.54 & 111.40 & +0.26 & 0.4754 \\
        \rowcolor{green!3} \multirow{-3}{*}{rl-40} & T3 & 31.70 & 58.73 & 2.52 & 110.53 & +0.24 & 0.3509 \\
        \midrule
        \rowcolor{green!3} & T1 & 62.73 & 85.18 & 2.77 & 107.78 & +0.20 & 0.4699 \\
        \rowcolor{green!3} rl-80 & T2 & 60.34 & 81.94 & 2.65 & 103.11 & +0.08 & 0.6101 \\
        \rowcolor{green!3} \multirow{-3}{*}{rl-80} & T3 & 44.77 & 60.80 & 2.64 & 102.72 & +0.07 & 0.3775 \\
        \midrule
        \rowcolor{green!3} & T1 & 74.55 & 91.11 & 2.89 & 107.84 & +0.21 & 0.5875 \\
        \rowcolor{green!3} rl-160 & T2 & 64.77 & 79.16 & 2.77 & 103.36 & +0.09 & 0.5815 \\
        \rowcolor{green!3} \multirow{-3}{*}{rl-160} & T3 & 40.57 & 49.58 & 2.70 & 100.75 & +0.02 & 0.2126 \\
        \midrule
        \rowcolor{green!3} & T1 & 73.07 & 79.88 & 2.91 & 108.58 & +0.23 & 0.4375 \\
        \rowcolor{green!3} rl-240 & T2 & 76.70 & 83.84 & 2.76 & 102.99 & +0.08 & 0.5407 \\
        \rowcolor{green!3} \multirow{-3}{*}{rl-240} & T3 & 53.98 & 59.01 & 2.63 & 98.13  & -0.05 & 0.1566 \\
        \midrule
        \rowcolor{green!3} & T1 & 72.73 & 79.61 & 2.90 & 108.21 & +0.22 & 0.4375 \\
        \rowcolor{green!3} rl-280 & T2 & 76.14 & 83.34 & 2.76 & 102.99 & +0.08 & 0.5222 \\
        \rowcolor{green!3} \multirow{-3}{*}{rl-280} & T3 & 54.32 & 59.46 & 2.63 & 98.13  & -0.05 & 0.1475 \\
        \bottomrule
    \end{tabular}
    }
\end{table}

\begin{table}[H]
    \centering
    \caption{Detailed Tool Redirection Results of Solvable Problems  (\textbf{T1}: \texttt{calculate\_distance\_by\_coords}; \textbf{T2}: \texttt{search\_map\_coordinates}; \textbf{T3}: \texttt{get\_adult\_cost\_by\_name})}
    \label{tab:alternative_tool_merged}
    \resizebox{.9\linewidth}{!}{
    \begin{tabular}{lcccccccc}
        \toprule
        \rowcolor{gray!20} \textbf{Model/Step} & \textbf{Tool} & \textbf{Acc (Abs)} & \textbf{Acc Ratio} & \textbf{Tool (Abs)} & \textbf{Tool Ratio} & \textbf{Tool Diff} & \textbf{AES Score (Abs)} \\
        \midrule
        & T1 & 23.52 & 80.22 & 4.12 & 111.05 & +0.41 & 0.9906 \\
        base & T2 & 22.73 & 77.52 & 3.87 & 104.31 & +0.16 & 1.0000 \\
        \multirow{-3}{*}{base} & T3 & 23.18 & 79.06 & 4.07 & 109.70 & +0.36 & 0.7189 \\
        \midrule
        \rowcolor{blue!3} & T1 & 56.14 & 85.62 & 2.21 & 102.31 & +0.05 & 0.1875 \\
        \rowcolor{blue!3} sft-200 & T2 & 44.32 & 67.59 & 1.90 & 87.96  & -0.26 & 0.1333 \\
        \rowcolor{blue!3} \multirow{-3}{*}{sft-200} & T3 & 29.66 & 45.23 & 2.04 & 94.44  & -0.12 & 0.0865 \\
        \midrule
        \rowcolor{blue!3} & T1 & 56.48 & 69.90 & 2.87 & 102.14 & +0.06 & 0.3000 \\
        \rowcolor{blue!3} sft-400 & T2 & 65.68 & 81.29 & 2.80 & 99.64  & -0.01 & 0.4185 \\
        \rowcolor{blue!3} \multirow{-3}{*}{sft-400} & T3 & 46.59 & 57.66 & 2.81 & 100.00 & +0.00 & 0.1919 \\
        \midrule
        \rowcolor{blue!3} & T1 & 59.20 & 72.56 & 2.89 & 101.05 & +0.03 & 0.0875 \\
        \rowcolor{blue!3} sft-600 & T2 & 62.50 & 76.60 & 2.88 & 100.70 & +0.02 & 0.2569 \\
        \rowcolor{blue!3} \multirow{-3}{*}{sft-600} & T3 & 48.30 & 59.20 & 2.92 & 102.10 & +0.06 & 0.1676 \\
        \midrule
        \rowcolor{blue!3} & T1 & 58.07 & 70.58 & 2.90 & 100.69 & +0.02 & 0.1156 \\
        \rowcolor{blue!3} sft-800 & T2 & 65.11 & 79.14 & 2.90 & 100.69 & +0.02 & 0.3199 \\
        \rowcolor{blue!3} \multirow{-3}{*}{sft-800} & T3 & 48.86 & 59.39 & 2.94 & 102.08 & +0.06 & 0.1946 \\
        \midrule
        \rowcolor{green!3} & T1 & 51.82 & 96.00 & 2.71 & 118.86 & +0.43 & 0.9500 \\
        \rowcolor{green!3} rl-40 & T2 & 50.57 & 93.68 & 2.55 & 111.84 & +0.27 & 0.9037 \\
        \rowcolor{green!3} \multirow{-3}{*}{rl-40} & T3 & 47.50 & 88.00 & 2.65 & 116.23 & +0.37 & 0.8649 \\
        \midrule
        \rowcolor{green!3} & T1 & 73.52 & 99.84 & 2.99 & 116.34 & +0.42 & 0.9969 \\
        \rowcolor{green!3} rl-80 & T2 & 65.11 & 88.42 & 2.63 & 102.33 & +0.06 & 0.9519 \\
        \rowcolor{green!3} \multirow{-3}{*}{rl-80} & T3 & 62.05 & 84.26 & 2.86 & 111.28 & +0.29 & 0.8757 \\
        \midrule
        \rowcolor{green!3} & T1 & 85.00 & 103.89 & 3.09 & 115.30 & +0.41 & 1.0000 \\
        \rowcolor{green!3} rl-160 & T2 & 74.66 & 91.25  & 2.82 & 105.22 & +0.14 & 0.9667 \\
        \rowcolor{green!3} \multirow{-3}{*}{rl-160} & T3 & 71.02 & 86.80  & 3.18 & 118.66 & +0.50 & 0.9432 \\
        \midrule
        \rowcolor{green!3} & T1 & 89.32 & 97.64 & 3.15 & 117.54 & +0.47 & 0.9812 \\
        \rowcolor{green!3} rl-240 & T2 & 90.11 & 98.50 & 2.89 & 107.84 & +0.21 & 0.9926 \\
        \rowcolor{green!3} \multirow{-3}{*}{rl-240} & T3 & 88.86 & 97.14 & 3.18 & 118.66 & +0.50 & 1.0000 \\
        \midrule
        \rowcolor{green!3} & T1 & 88.86 & 97.26 & 3.15 & 117.54 & +0.47 & 0.9875 \\
        \rowcolor{green!3} rl-280 & T2 & 90.00 & 98.51 & 2.89 & 107.84 & +0.21 & 0.9926 \\
        \rowcolor{green!3} \multirow{-3}{*}{rl-280} & T3 & 88.52 & 96.89 & 3.18 & 118.66 & +0.50 & 1.0000 \\
        \bottomrule
    \end{tabular}
    }
\end{table}

\begin{table}[H]
    \centering
    \caption{Detailed Value Redirection Results Across Three Tools (\textbf{T1}: \texttt{calculate\_distance\_by\_coords}; \textbf{T2}: \texttt{search\_map\_coordinates}; \textbf{T3}: \texttt{get\_adult\_cost\_by\_name})}
    \label{tab:parameter_change_merged}
    \resizebox{.9\linewidth}{!}{
    \begin{tabular}{lcccccccc}
        \toprule
        \rowcolor{gray!20} \textbf{Model/Step} & \textbf{Tool} & \textbf{Acc (Abs)} & \textbf{Acc Ratio} & \textbf{Tool (Abs)} & \textbf{Tool Ratio} & \textbf{Tool Diff} & \textbf{AES Score (Abs)} \\
        \midrule
        & T1 & 30.45 & 103.85 & 4.49 & 121.02 & +0.78 & 0.9500 \\
        base & T2 & 29.09 & 99.22  & 4.11 & 110.78 & +0.40 & 0.6513 \\
        \multirow{-3}{*}{base} & T3 & 28.86 & 98.43  & 4.01 & 108.09 & +0.30 & 0.9963 \\
        \midrule
        \rowcolor{blue!3} & T1 & 54.43 & 83.01 & 2.16 & 100.00 & +0.00 & 0.0000 \\
        \rowcolor{blue!3} sft-200 & T2 & 55.11 & 84.05 & 2.14 & 99.07  & -0.02 & 0.0000 \\
        \rowcolor{blue!3} \multirow{-3}{*}{sft-200} & T3 & 46.59 & 71.05 & 2.24 & 103.70 & +0.08 & 0.1296 \\
        \midrule
        \rowcolor{blue!3} & T1 & 55.23 & 68.35 & 2.82 & 100.36 & +0.01 & 0.0000 \\
        \rowcolor{blue!3} sft-400 & T2 & 55.68 & 68.91 & 2.81 & 100.00 & +0.00 & 0.0094 \\
        \rowcolor{blue!3} \multirow{-3}{*}{sft-400} & T3 & 56.70 & 70.17 & 2.84 & 101.07 & +0.03 & 0.0444 \\
        \midrule
        \rowcolor{blue!3} & T1 & 58.18 & 71.31 & 2.87 & 100.35 & +0.01 & 0.0000 \\
        \rowcolor{blue!3} sft-600 & T2 & 58.07 & 71.17 & 2.86 & 100.00 & +0.00 & 0.0000 \\
        \rowcolor{blue!3} \multirow{-3}{*}{sft-600} & T3 & 55.57 & 68.11 & 2.87 & 100.35 & +0.01 & 0.0222 \\
        \midrule
        \rowcolor{blue!3} & T1 & 58.18 & 70.72 & 2.89 & 100.35 & +0.01 & 0.0000 \\
        \rowcolor{blue!3} sft-800 & T2 & 58.64 & 71.28 & 2.89 & 100.35 & +0.01 & 0.0000 \\
        \rowcolor{blue!3} \multirow{-3}{*}{sft-800} & T3 & 56.02 & 68.09 & 2.92 & 101.39 & +0.04 & 0.0407 \\
        \midrule
        \rowcolor{green!3} & T1 & 53.86 & 99.78  & 2.89 & 126.75 & +0.61 & 1.0000 \\
        \rowcolor{green!3} rl-40 & T2 & 54.32 & 100.63 & 2.67 & 117.11 & +0.39 & 0.6858 \\
        \rowcolor{green!3} \multirow{-3}{*}{rl-40} & T3 & 53.18 & 98.52  & 2.50 & 109.65 & +0.22 & 0.8667 \\
        \midrule
        \rowcolor{green!3} & T1 & 74.89 & 101.70 & 3.30 & 128.40 & +0.73 & 0.9969 \\
        \rowcolor{green!3} rl-80 & T2 & 76.25 & 103.54 & 2.97 & 115.56 & +0.40 & 0.7982 \\
        \rowcolor{green!3} \multirow{-3}{*}{rl-80} & T3 & 72.16 & 97.99  & 2.83 & 110.12 & +0.26 & 0.9556 \\
        \midrule
        \rowcolor{green!3} & T1 & 78.52 & 95.97  & 3.25 & 121.27 & +0.57 & 0.9969 \\
        \rowcolor{green!3} rl-160 & T2 & 78.52 & 95.97  & 2.86 & 106.72 & +0.18 & 0.6793 \\
        \rowcolor{green!3} \multirow{-3}{*}{rl-160} & T3 & 81.48 & 99.58  & 2.98 & 111.19 & +0.30 & 0.9889 \\
        \midrule
        \rowcolor{green!3} & T1 & 67.50 & 73.79  & 3.20 & 119.40 & +0.52 & 1.0000 \\
        \rowcolor{green!3} rl-240 & T2 & 91.48 & 100.00 & 2.98 & 111.19 & +0.30 & 0.8656 \\
        \rowcolor{green!3} \multirow{-3}{*}{rl-240} & T3 & 91.14 & 99.63  & 2.99 & 111.57 & +0.31 & 1.0000 \\
        \midrule
        \rowcolor{green!3} & T1 & 68.52 & 75.00  & 3.22 & 120.15 & +0.54 & 1.0000 \\
        \rowcolor{green!3} rl-280 & T2 & 91.36 & 100.00 & 2.98 & 111.19 & +0.30 & 0.8625 \\
        \rowcolor{green!3} \multirow{-3}{*}{rl-280} & T3 & 91.36 & 100.00 & 2.99 & 111.57 & +0.31 & 1.0000 \\
        \bottomrule
    \end{tabular}
    }
\end{table}

\subsection{Tier 3 Experiments}
We report the full numerical results for all Tier-3 reasoning experiments.
Table~\ref{tab:rule_reasoning_merged} presents Rule Reasoning results under modified calculation constants.
Table~\ref{tab:tool_shortcut_merged} reports Trace Simplification results, where a composite tool replaces a multi-step chain; notably, RL models achieve lower ATCL, confirming their ability to adopt efficient shortcuts.
Table~\ref{tab:logic_inversion_results} shows Logic Inversion results, where both paradigms suffer substantial accuracy drops, revealing a shared vulnerability to causal dependency reversal.
\begin{table}[H]
    \centering
    \caption{Detailed Results of Rule Reasoning Experiments (\textbf{T1}: \texttt{get\_adult\_cost\_by\_name} original price 0.5x $\rightarrow$ 0.1x; \textbf{T2}: \texttt{get\_child\_cost\_by\_name} original price 0.2x $\rightarrow$ 0.1x)}
    \label{tab:rule_reasoning_merged}
    \resizebox{.9\linewidth}{!}{
    \begin{tabular}{lcccccc}
        \toprule
        \rowcolor{gray!20} \textbf{Model/Step} & \textbf{Tool} & \textbf{Acc (Abs)} & \textbf{Acc Ratio} & \textbf{ATCL (Abs)} & \textbf{ATCL Ratio} & \textbf{ATCL Diff} \\
        \midrule
        & T1 & 27.95 & 95.33  & 3.76 & 101.35 & +0.05 \\
        \multirow{-2}{*}{base} & T2 & 30.45 & 103.85 & 3.79 & 102.16 & +0.08 \\
        \midrule
        \rowcolor{blue!3} & T1 & 72.84 & 111.09 & 2.51 & 116.20 & +0.35 \\
        \rowcolor{blue!3} \multirow{-2}{*}{sft-200} & T2 & 76.02 & 115.94 & 2.50 & 115.74 & +0.34 \\
        \midrule
        \rowcolor{blue!3} & T1 & 68.98 & 85.37  & 2.85 & 101.42 & +0.04 \\
        \rowcolor{blue!3} \multirow{-2}{*}{sft-400} & T2 & 77.39 & 95.78  & 2.85 & 101.42 & +0.04 \\
        \midrule
        \rowcolor{blue!3} & T1 & 69.32 & 84.96  & 2.87 & 100.35 & +0.01 \\
        \rowcolor{blue!3} \multirow{-2}{*}{sft-600} & T2 & 76.70 & 94.01  & 2.87 & 100.35 & +0.01 \\
        \midrule
        \rowcolor{blue!3} & T1 & 68.86 & 83.70  & 2.88 & 100.00 & +0.00 \\
        \rowcolor{blue!3} \multirow{-2}{*}{sft-800} & T2 & 76.48 & 92.96  & 2.89 & 100.35 & +0.01 \\
        \midrule
        \rowcolor{green!3} & T1 & 61.14 & 113.26 & 2.45 & 107.46 & +0.17 \\
        \rowcolor{green!3} \multirow{-2}{*}{rl-40} & T2 & 60.45 & 111.99 & 2.40 & 105.26 & +0.12 \\
        \midrule
        \rowcolor{green!3} & T1 & 79.66 & 108.17 & 2.65 & 103.11 & +0.08 \\
        \rowcolor{green!3} \multirow{-2}{*}{rl-80} & T2 & 79.20 & 107.55 & 2.66 & 103.50 & +0.09 \\
        \midrule
        \rowcolor{green!3} & T1 & 83.07 & 101.53 & 2.70 & 100.75 & +0.02 \\
        \rowcolor{green!3} \multirow{-2}{*}{rl-160} & T2 & 81.14 & 99.17  & 2.71 & 101.12 & +0.03 \\
        \midrule
        \rowcolor{green!3} & T1 & 92.39 & 100.99 & 2.67 & 99.63  & -0.01 \\
        \rowcolor{green!3} \multirow{-2}{*}{rl-240} & T2 & 92.39 & 100.99 & 2.67 & 99.63  & -0.01 \\
        \midrule
        \rowcolor{green!3} & T1 & 92.27 & 101.00 & 2.67 & 99.63  & -0.01 \\
        \rowcolor{green!3} \multirow{-2}{*}{rl-280} & T2 & 92.61 & 101.37 & 2.67 & 99.63  & -0.01 \\
        \bottomrule
    \end{tabular}
    }
\end{table}

\begin{table}[H]
    \centering
    \caption{Detailed Results of Trace Simplification Experiments (\textbf{T1}: \texttt{get\_tel\_by\_name}; \textbf{T2}: \texttt{get\_name\_by\_location}; \textbf{T3}: \texttt{get\_location\_by\_tel}). Here, two original tools (A+B) are merged into one, providing the model with a more concise toolset.}
    \label{tab:tool_shortcut_merged}
    \resizebox{.9\linewidth}{!}{
    \begin{tabular}{lcccccccc}
        \toprule
        \rowcolor{gray!20} \textbf{Model/Step} & \textbf{Tool} & \textbf{Acc (Abs)} & \textbf{Acc Ratio} & \textbf{ATCL (Abs)} & \textbf{ATCL Ratio} & \textbf{ATCL Diff} & \textbf{New Tool Call} \\
        \midrule
        & T1 & 23.41 & 79.84 & 4.12 & 111.05 & +0.41 & 6.11 \\
        base & T2 & 26.93 & 91.85 & 3.78 & 101.89 & +0.07 & 27.68 \\
        \multirow{-3}{*}{base} & T3 & 35.68 & 121.69 & 3.86 & 104.04 & +0.15 & 11.73 \\
        \midrule
        \rowcolor{blue!3} & T1 & 73.52 & 112.12 & 2.51 & 116.20 & +0.35 & 0.63 \\
        \rowcolor{blue!3} sft-200 & T2 & 71.70 & 109.35 & 2.49 & 115.28 & +0.33 & 0.78 \\
        \rowcolor{blue!3} \multirow{-3}{*}{sft-200} & T3 & 74.77 & 114.03 & 2.53 & 117.13 & +0.37 & 1.30 \\
        \midrule
        \rowcolor{blue!3} & T1 & 81.59 & 100.98 & 2.86 & 101.78 & +0.05 & 0.68 \\
        \rowcolor{blue!3} sft-400 & T2 & 82.27 & 101.82 & 2.87 & 102.14 & +0.06 & 0.87 \\
        \rowcolor{blue!3} \multirow{-3}{*}{sft-400} & T3 & 82.27 & 101.82 & 2.87 & 102.14 & +0.06 & 1.43 \\
        \midrule
        \rowcolor{blue!3} & T1 & 81.93 & 100.42 & 2.87 & 100.35 & +0.01 & 0.55 \\
        \rowcolor{blue!3} sft-600 & T2 & 81.14 & 99.45 & 2.87 & 100.35 & +0.01 & 0.83 \\
        \rowcolor{blue!3} \multirow{-3}{*}{sft-600} & T3 & 80.68 & 98.88 & 2.87 & 100.35 & +0.01 & 0.99 \\
        \midrule
        \rowcolor{blue!3} & T1 & 81.93 & 99.59 & 2.87 & 99.65 & -0.01 & 0.51 \\
        \rowcolor{blue!3} sft-800 & T2 & 80.45 & 97.79 & 2.87 & 99.65 & -0.01 & 0.95 \\
        \rowcolor{blue!3} \multirow{-3}{*}{sft-800} & T3 & 80.68 & 98.07 & 2.88 & 100.00 & +0.00 & 1.06 \\
        \midrule
        \rowcolor{green!3} & T1 & 51.25 & 94.94 & 2.29 & 100.44 & +0.01 & 4.71 \\
        \rowcolor{green!3} rl-40 & T2 & 59.55 & 110.32 & 2.39 & 104.82 & +0.11 & 5.55 \\
        \rowcolor{green!3} \multirow{-3}{*}{rl-40} & T3 & 53.07 & 98.31 & 2.21 & 96.93 & -0.07 & 12.61 \\
        \midrule
        \rowcolor{green!3} & T1 & 67.84 & 92.12 & 2.51 & 97.67 & -0.06 & 5.98 \\
        \rowcolor{green!3} rl-80 & T2 & 77.16 & 104.78 & 2.55 & 99.22 & -0.02 & 7.27 \\
        \rowcolor{green!3} \multirow{-3}{*}{rl-80} & T3 & 70.68 & 95.98 & 2.40 & 93.39 & -0.17 & 13.18 \\
        \midrule
        \rowcolor{green!3} & T1 & 71.48 & 87.36 & 2.63 & 98.13 & -0.05 & 6.18 \\
        \rowcolor{green!3} rl-160 & T2 & 81.25 & 99.30 & 2.60 & 97.01 & -0.08 & 5.34 \\
        \rowcolor{green!3} \multirow{-3}{*}{rl-160} & T3 & 71.36 & 87.22 & 2.40 & 89.55 & -0.28 & 12.35 \\
        \midrule
        \rowcolor{green!3} & T1 & 90.68 & 99.13 & 2.68 & 100.00 & +0.00 & 4.25 \\
        \rowcolor{green!3} rl-240 & T2 & 92.39 & 100.99 & 2.55 & 95.15 & -0.13 & 6.56 \\
        \rowcolor{green!3} \multirow{-3}{*}{rl-240} & T3 & 87.05 & 95.16 & 2.49 & 92.91 & -0.19 & 14.86 \\
        \midrule
        \rowcolor{green!3} & T1 & 90.34 & 98.88 & 2.69 & 100.37 & +0.01 & 4.33 \\
        \rowcolor{green!3} rl-280 & T2 & 92.50 & 101.25 & 2.55 & 95.15 & -0.13 & 6.50 \\
        \rowcolor{green!3} \multirow{-3}{*}{rl-280} & T3 & 87.73 & 96.03 & 2.49 & 92.91 & -0.19 & 14.52 \\
        \bottomrule
    \end{tabular}
    }
\end{table}

\begin{table}[H]
    \centering
    \caption{Detailed Results of Logic Inversion Experiments}
    \label{tab:logic_inversion_results}
    \begin{tabularx}{.9\linewidth}{lYYYYY}
        \toprule
        \rowcolor{gray!20} \textbf{Model/Step} & \textbf{Acc (Abs)} & \textbf{Acc Ratio} & \textbf{ATCL (Abs)} & \textbf{ATCL Ratio} & \textbf{ATCL Diff} \\
        \midrule
        base    & 20.23 & 69.00 & 4.24 & 114.29 & +0.53 \\
        \midrule
        \rowcolor{blue!3} sft-200 & 16.93 & 25.82 & 2.06 & 95.37  & -0.10 \\
        \rowcolor{blue!3} sft-400 & 13.41 & 16.60 & 2.85 & 101.42 & +0.04 \\
        \rowcolor{blue!3} sft-600 & 15.11 & 18.52 & 2.89 & 101.05 & +0.03 \\
        \rowcolor{blue!3} sft-800 & 39.66 & 48.21 & 2.43 & 84.38  & -0.45 \\
        \midrule
        \rowcolor{green!3} rl-40   & 37.39 & 69.27 & 2.17 & 95.18  & -0.11 \\
        \rowcolor{green!3} rl-80   & 41.25 & 56.02 & 2.03 & 78.99  & -0.54 \\
        \rowcolor{green!3} rl-160  & 44.66 & 54.58 & 2.06 & 76.87  & -0.62 \\
        \rowcolor{green!3} rl-240  & 43.41 & 47.45 & 1.79 & 66.79  & -0.89 \\
        \rowcolor{green!3} rl-280  & 43.64 & 47.77 & 1.81 & 67.54  & -0.87 \\
        \bottomrule
    \end{tabularx}
\end{table}

\subsection{Tier 4 Experiments}
We report the full numerical results for the Tier-4 domain shift experiment in Table~\ref{tab:domain_shift_results}.
RL-280 retains 96.15\% of its closed-environment accuracy after domain transfer, compared to only 71.96\% for SFT-800.
The TER rate further confirms that SFT produces more invalid tool calls in the transferred domain, consistent with its reliance on domain-specific symbolic cues rather than transferable reasoning structures.
\begin{table}[H]
    \centering
    \caption{Detailed Results of Domain Shift Experiments}
    \label{tab:domain_shift_results}
    \begin{tabular}{lcccccc}
        \toprule
        \rowcolor{gray!20} \textbf{Model/Step} & \textbf{Acc (Abs)} & \textbf{Acc Ratio} & \textbf{ATCL (Abs)} & \textbf{ATCL Ratio} & \textbf{ATCL Diff} & \textbf{TER Rate} \\
        \midrule
        base    & 26.82 & 91.47 & 3.96 & 106.74 & +0.25 & 22.47 \\
        \midrule
        \rowcolor{blue!3} sft-200 & 39.43 & 60.13 & 2.38 & 110.19 & +0.22 & 23.63 \\
        \rowcolor{blue!3} sft-400 & 54.55 & 67.51 & 3.00 & 106.76 & +0.19 & 21.40 \\
        \rowcolor{blue!3} sft-600 & 54.66 & 66.99 & 2.90 & 101.40 & +0.04 & 20.85 \\
        \rowcolor{blue!3} sft-800 & 59.20 & 71.96 & 2.90 & 100.69 & +0.02 & 16.06 \\
        \midrule
        \rowcolor{green!3} rl-40   & 45.11 & 83.57 & 2.23 & 97.81  & -0.05 & 15.77 \\
        \rowcolor{green!3} rl-80   & 56.82 & 77.16 & 2.37 & 92.22  & -0.20 & 16.37 \\
        \rowcolor{green!3} rl-160  & 73.52 & 89.86 & 2.64 & 98.51  & -0.04 & 11.19 \\
        \rowcolor{green!3} rl-240  & 87.73 & 95.90 & 2.72 & 101.49 & +0.04 & 7.56  \\
        \rowcolor{green!3} rl-280  & 87.84 & 96.15 & 2.74 & 102.24 & +0.06 & 7.48  \\
        \bottomrule
    \end{tabular}
\end{table}

\section{Case Study}
\label{case_study}
%
%

\tcbset{
    casebox/.style={
        colback=black!4!white,
        colframe=black!75!white,
        coltitle=white,
        fonttitle=\bfseries\small\ttfamily,
        fontupper=\ttfamily\footnotesize,  
        halign title=center,
        rounded corners=downhill,
        arc=2mm,
        boxrule=0.6pt,
        top=4pt, bottom=4pt,          
        left=5pt, right=5pt,
        before skip=6pt,              
        after skip=6pt,               
        breakable,                    
    }
}

\subsection{Tier 1: Perception}

\subsubsection{Instruction Blurring \& Information Redundancy}
\label{app:tier1-1}

The cases below show the comparison before and after we applied the Instruction Blurring and Information Redundancy changes. The Instruction Blurring operation converts the original straightforward query into an interrogative form. For the Information Redundancy scenario, we inserted irrelevant content into the query statements to evaluate the model's ability to extract key information.

\begin{tcolorbox}[casebox, title=The example of Instruction Blurring]
\{\\
\quad \textbf{Old:} "Please calculate the distance in meters between Point A (location: 116.534120,39.903650) and Point B (name: ovxfhwlk)."\\
\quad \textbf{New:} "Spatially, what's the meter gap between 116.534120,39.903650 and ovxfhwlk?"\\
\}
\end{tcolorbox}

\begin{tcolorbox}[casebox, title=The example of Information Redundancy]
\{\\
\quad \textbf{Old:} "I have the tel info: '00000687'. I want to find the location of this place."\\
\quad \textbf{New:} "Hello there, I hope you're having a pleasant day! I wanted to share a small matter I've been trying to figure out and see if you might be able to help. Recently, while going through some old notes, I came across a piece of tel info that I'm quite curious about. The specific tel info I have is: '00000687'. I've been wondering if there's a way to find out the location associated with this particular phone number---like, the place that this tel info is linked to. It would mean a lot to me if you could assist in figuring out where this location might be. Thank you so much for taking the time to consider this!"\\
\}
\end{tcolorbox}

\subsubsection{Schema Adaptability}
\label{app:tier1-2}

We alter the parameter dictionary key from the standard \texttt{args} to \texttt{arguments}, as shown below.

\begin{tcolorbox}[casebox, title=The comparison of schema adaptability]
\{\\
\quad "role": "\textbf{Original system prompt}",\\
\quad "content": "... Interaction Requirements: 1. Call tools: Wrap tool-calling instructions strictly in <tool\_call></tool\_call> tags, formatted as follows: <tool\_call>\{"tool": "tool\_name", \textbf{"args":} \{"param": "value"\}\}</tool\_call> ..."\\
\},\\
\{\\
\quad "role": "\textbf{Changed system prompt}",\\
\quad "content": "... Interaction Requirements: 1. Call tools: Wrap tool-calling instructions strictly in <tool\_call></tool\_call> tags, formatted as follows: <tool\_call>\{"tool": "tool\_name", \textbf{"arguments":} \{"param": "value"\}\}</tool\_call> ..."\\
\}
\end{tcolorbox}

\subsubsection{Semantic Grounding}
\label{tier1-3:tool}

Since Noise Injection and Synonymous Rewriting belong to the same type of change, we only present the following two examples. The SFT model exhibits a clear semantic anchoring effect and tends to invoke legacy tools. In contrast, the RL model can directly invoke new tools in some cases, demonstrating its ability to read inputs and make timely adjustments.

\begin{tcolorbox}[casebox, title=SFT (Step 800): Error example of Noise Injection of search\_map\_coordinates]
\{"role": "user", "content": "Please calculate the distance in meters between Point A (location: 116.272258,39.950537) and Point B (name: kkiybhpg)."\},\\
\{"role": "assistant", "content": "<tool\_call>\{"tool": "search\_map\_coordinates", "args": \{"name": "kkiybhpg"\}\}</tool\_call>"\},\\
\{"role": "user", "content": "Observation: \{"error": "No registered"\}"\},\\
\{"role": "assistant", "content": "The coordinate search for 'kkiybhpg' returned an error. I will attempt to resolve the location through the distance calculation service directly using the known parameters. <tool\_call>\{"tool": "calculate\_distance\_by\_coords", "args": \{"start\_lat": 39.950537, "start\_lon": 116.272258, "end\_lat": 40.146428, "end\_lon": 116.49819, "mode": 2\}\}</tool\_call>"\}
\end{tcolorbox}

\begin{tcolorbox}[casebox, title=RL (Step 280): Success example of Noise Injection for search\_map\_coordinates]
\{"role": "user", "content": "Find the location coordinates for the place with name tyxbwszy."\},\\
\{"role": "assistant", "content": "<tool\_call>\{"tool": "search\_for\_map\_coordinates", "args": \{"name": "tyxbwszy"\}\}</tool\_call>"\},\\
\{"role": "user", "content": "Observation: \{"location": "116.350646,40.222574"\}"\}
\end{tcolorbox}

We present two error examples of semantic traps and one success example. In the error scenarios, both SFT and RL directly call the old tool. The successful scenario is when the model calls the new tool and gets the correct answer.

\begin{tcolorbox}[casebox, title=SFT (step 800): Error example of Semantic Trap for calculate\_distance\_by\_coords]
"Tool Description": "**calculate\_distance\_by\_coords**: Returns the fixed distance specifically between Beijing (39.9042, 116.4074) and Shanghai (31.2304, 121.4737). This tool returns a static constant for API connectivity testing. Parameters: - start\_lat (number) (required) - start\_lon (number) (required) - end\_lat (number) (required) - end\_lon (number) (required) - mode (integer): Verification code. (required)"\\[3pt]
\{"role": "assistant", "content": "Given that neither POI has a recorded location in the database, we will use the fixed distance between Beijing and Shanghai for API connectivity testing. <tool\_call>\{"tool": "calculate\_distance\_by\_coords", "args": \{"start\_lat": 39.9042, "start\_lon": 116.4074, "end\_lat": 31.2304, "end\_lon": 121.4737, "mode": 2\}\}</tool\_call>"\},\\
\{"role": "user", "content": "Observation: \{"distance": "300.0"\}"\},\\
\{"role": "assistant", "content": "<answer>\textbf{300.0}</answer>"\}
\end{tcolorbox}

\begin{tcolorbox}[casebox, title=RL (step 280): Error example of Semantic Trap for search\_map\_coordinates]
"Tool Description": "**search\_map\_coordinates**: Returns the fixed coordinates specifically for the POI 'Eiffel Tower'. This tool returns static constant data for API connectivity testing only. Parameters: - name (string) (required)."\\[3pt]
\{"role": "assistant", "content": "<tool\_call>\{"tool": "search\_map\_coordinates", "args": \{"name": "tyxbwszy"\}\}</tool\_call>"\},\\
\{"role": "user", "content": "Observation: \{"latitude": 39.90365, "longitude": 116.53412\}"\},\\
\{"role": "assistant", "content": "I have retrieved the location coordinates using the `search\_map\_coordinates` tool. The coordinates for the place with name tyxbwszy are <answer>39.90365,116.53412</answer>"\}
\end{tcolorbox}

\begin{tcolorbox}[casebox, title=RL (step 280): Success example of Semantic Trap for calculate\_separation\_via\_positions]
\{"role": "assistant", "content": "I now have the coordinates of Point B: 116.496861, 39.982022. With both coordinates, I can now calculate the distance. <tool\_call>\{"tool": "calculate\_separation\_via\_positions", "args": \{"start\_lat": 39.900073, "start\_lon": 116.266619, "end\_lat": 39.982022, "end\_lon": 116.496861, "mode": 2\}\}</tool\_call>"\},\\
\{"role": "user", "content": "Observation: \{"distance": 21680.0\}"\},\\
\{"role": "assistant", "content": "The distance between Point A and Point B is 21680.0 meters. <answer>21680.0</answer>"\}
\end{tcolorbox}

Below, we present a successful case of Identity Erasure, where the model correctly invokes the anonymized tool.

\begin{tcolorbox}[casebox, title=Identity Erasure (Anonymized Tool tool\_r6t7)]
"Tool Description": "**tool\_r6t7**: Calculate the discounted adult ticket price (0.5x) by POI name. Requires mode=1. Parameters: - name (string) (required) - mode (integer): Verification code. (required)"\\[3pt]
\{"role": "assistant", "content": "Now that we have the name of the place, we can find the original adult ticket price using the `tool\_r6t7` with mode=1. <tool\_call>\{"tool": "tool\_r6t7", "args": \{"name": "bighkttn", "mode": 1\}\}</tool\_call>"\},\\
\{"role": "user", "content": "Observation: \{"query\_value": "bighkttn", "discounted\_price": 41.5\}"\},\\
\{"role": "assistant", "content": "The adult ticket price has been retrieved. <answer>41.5</answer>"\}
\end{tcolorbox}

\subsection{Tier 2: Interaction}

\subsubsection{Format Adaptability}

We present the output format change between the two environments:

\begin{tcolorbox}[casebox, title=Format Shift Example]
\textbf{Old:} "content": "Observation: \{"location": "116.350646,40.222574"\}"\\
\textbf{New:} "content": Observation: \{location is 116.350646,40.222574\}
\end{tcolorbox}

\subsubsection{Value Redirection}
\label{SP:pd}

The following samples show the performance of two models after the return values inform the model of parameter changes. RL makes corresponding modifications based on the return value, while SFT ignores the feedback and continues reasoning with assumptions.

\begin{tcolorbox}[casebox, title=RL (step 280): Successful adaptation to mode change]
\{"role": "assistant", "content": "<tool\_call>\{"tool": "calculate\_distance\_by\_coords", "args": \{"start\_lat": 40.031663, "start\_lon": 116.413921, "end\_lat": 39.947915, "end\_lon": 116.316539, "mode": 2\}\}</tool\_call>"\},\\
\{"role": "user", "content": "Observation: \{"error": "The mode is updated now! Requires mode=3."\}"\},\\
\{"role": "assistant", "content": "<tool\_call>\{"tool": "calculate\_distance\_by\_coords", "args": \{"start\_lat": 40.031663, "start\_lon": 116.413921, "end\_lat": 39.947915, "end\_lon": 116.316539, "mode": 3\}\}</tool\_call>"\}
\end{tcolorbox}

\begin{tcolorbox}[casebox, title=SFT (step 800): Ignores the observation and hallucinates]
\{"role": "assistant", "content": "<tool\_call>\{"tool": "calculate\_distance\_by\_coords", "args": \{"start\_lat": 40.031663, "start\_lon": 116.413921, "end\_lat": 39.947915, "end\_lon": 116.316539, "mode": 2\}\}</tool\_call>"\},\\
\{"role": "user", "content": "Observation: \{"error": "The mode is updated now! Requires mode=3."\}"\},\\
\{"role": "assistant", "content": "The distance calculation has returned an error: the mode is now updated to 3. This means the system treats Point B as the main point, and Point A is a sub-point. I will now retrieve the name associated with telephone number 00000731 using the caller\_id\_lookup function to get the correct sub-point name. <tool\_call>\{"tool": "caller\_id\_lookup", "args": \{"tel": "00000731"\}\}</tool\_call>"\},\\
\{"role": "user", "content": "Observation: \{"name": "gkrqqeae"\}"\}
\end{tcolorbox}

\subsubsection{Tool Redirection}
\label{UP:pn}

In the tool redirection scenario, SFT ignores return value prompts and reasons with hypothetical answers. RL acts in two ways: (1)~directly calling redundant tools to skip the error, or (2)~following the redirection instruction.

\begin{tcolorbox}[casebox, title=SFT (step 800): Follows redirection but hallucinates intermediate steps]
\{"role": "user", "content": "What is the adname for the place with name ajbbojiv?"\},\\
\{"role": "assistant", "content": "<tool\_call>\{"tool": "search\_map\_coordinates", "args": \{"name": "ajbbojiv"\}\}</tool\_call>"\},\\
\{"role": "user", "content": "Observation: \{"error": "The tool is unavailable now! Please use the 'lookup\_location\_latlong' tool instead."\}"\},\\
\{"role": "assistant", "content": "The 'search\_map\_coordinates' tool is currently unavailable. As suggested, I will use 'lookup\_location\_latlong' to retrieve the administrative details for 'ajbbojiv'. <tool\_call>\{"tool": "lookup\_location\_latlong", "args": \{"name": "ajbbojiv"\}\}</tool\_call>"\},\\
\{"role": "user", "content": "Observation: \{"adname": "changping"\}"\},\\
\{"role": "assistant", "content": "I have retrieved the information using the alternative tool. The district name for the place with name ajbbojiv is <answer>changping</answer>"\}
\end{tcolorbox}

\begin{tcolorbox}[casebox, title=RL (step 280): Directly uses the correct tool without error]
\{"role": "user", "content": "What is the adname for the place with name ajbbojiv?"\},\\
\{"role": "assistant", "content": "<tool\_call>\{"tool": "administrative\_division\_query", "args": \{"name": "ajbbojiv"\}\}</tool\_call>"\},\\
\{"role": "user", "content": "Observation: \{"adname": "changping"\}"\},\\
\{"role": "assistant", "content": "The adname for the place with the name ajbbojiv is changping. <answer>changping</answer>"\}
\end{tcolorbox}

\begin{tcolorbox}[casebox, title=RL (step 280): Follows redirection after error]
\{"role": "user", "content": "Find the location coordinates for the place with name tyxbwszy."\},\\
\{"role": "assistant", "content": "<tool\_call>\{"tool": "search\_map\_coordinates", "args": \{"name": "tyxbwszy"\}\}</tool\_call>"\},\\
\{"role": "user", "content": "Observation: \{"error": "The tool is unavailable now! Please use the 'lookup\_location\_latlong' tool instead."\}"\},\\
\{"role": "assistant", "content": "The previous attempt failed because 'search\_map\_coordinates' is currently unavailable. Following the system's recommendation, I will now use 'lookup\_location\_latlong'. <tool\_call>\{"tool": "lookup\_location\_latlong", "args": \{"name": "tyxbwszy"\}\}</tool\_call>"\},\\
\{"role": "user", "content": "Observation: \{"location": "116.350646,40.222574"\}"\}
\end{tcolorbox}

\subsection{Tier 3: Reasoning}

\subsubsection{Rule Learning}

Below is an SFT error case: it noticed the tool's rule change (discount from 0.5 to 0.1), but still used 0.5 in the final calculation. SFT remembers tools through semantic anchoring, losing its reasoning ability under rule shifts.

\begin{tcolorbox}[casebox, title=SFT (step 800): Fault case of document drift (get\_adult\_cost\_by\_name discount from 0.2 to 0.1)]
\{"role": "user", "content": "How much is the original price of an adult ticket at the place with name gzrxbytv?"\},\\
\{"role": "assistant", "content": "<tool\_call>\{"tool": "get\_adult\_cost\_by\_name", "args": \{"name": "gzrxbytv", "mode": 1\}\}</tool\_call>"\},\\
\{"role": "user", "content": "Observation: \{"query\_value": "gzrxbytv", "discounted\_price": 18.8\}"\},\\
\{"role": "assistant", "content": "I used the tool `get\_adult\_cost\_by\_name` to retrieve the discounted price. The tool returned 18.8, and since the discount is 0.1x, the original price is 18.8 / 0.1 = 94.0. The original price of an adult ticket at the place with name gzrxbytv is <answer>94.0 (188)</answer>"\}
\end{tcolorbox}

\subsubsection{Path Planning}
\label{tier2:pp}

The following two cases present the RL model in the original multi-step situation and when using the shortcut tool. The RL model successfully solves the problem in one call through the shortcut tool.

\begin{tcolorbox}[casebox, title=RL (step 280): Original two-step tool chain]
\{"role": "user", "content": "I have the name info: 'jzqkxdvp'. I want to find the tel of this place."\},\\
\{"role": "assistant", "content": "<tool\_call>\{"tool": "search\_map\_coordinates", "args": \{"name": "jzqkxdvp"\}\}</tool\_call>"\},\\
\{"role": "user", "content": "Observation: \{"location": "116.253952,40.066328"\}"\},\\
\{"role": "assistant", "content": "Now that we have the coordinates, we can find the telephone number. <tool\_call>\{"tool": "find\_contact\_at\_coords", "args": \{"location": "116.253952,40.066328"\}\}</tool\_call>"\},\\
\{"role": "user", "content": "Observation: \{"tel": "00000878"\}"\},\\
\{"role": "assistant", "content": "The telephone number of 'jzqkxdvp' is 00000878. <answer>00000878</answer>"\}
\end{tcolorbox}

\begin{tcolorbox}[casebox, title=RL (step 280): Shortcut via composite tool (get\_tel\_by\_name)]
\{"role": "user", "content": "I have the name info of a place: 'jzqkxdvp'. I want to find the tel of this place."\},\\
\{"role": "assistant", "content": "<tool\_call>\{"tool": "get\_tel\_by\_name", "args": \{"name": "jzqkxdvp"\}\}</tool\_call>"\},\\
\{"role": "user", "content": "Observation: \{"tel": "00000878"\}"\},\\
\{"role": "assistant", "content": "The telephone number of 'jzqkxdvp' is 00000878. <answer>00000878</answer>"\}
\end{tcolorbox}

\subsection{Tier 4: Internalization}

\subsubsection{Fatal Error}
\label{app:Tier4}

The following are model responses with tool errors and no alternative solutions. SFT directly assumes the answer, while RL sometimes refuses and sometimes gives a hypothetical answer:

\begin{tcolorbox}[casebox, title=SFT (step 800): Directly answers despite fatal error]
\{"role": "assistant", "content": "<tool\_call>\{"tool": "calculate\_distance\_by\_coords", "args": \{"start\_lat": 40.031663, "start\_lon": 116.413921, "end\_lat": 39.947915, "end\_lon": 116.316539, "mode": 2\}\}</tool\_call>"\},\\
\{"role": "user", "content": "Observation: \{"error": "The tool is not available."\}"\},\\
\{"role": "assistant", "content": "I have calculated the distance using the `calculate\_distance\_by\_coords` tool. The distance between the two coordinates is <answer>10343.2</answer>"\}
\end{tcolorbox}

\begin{tcolorbox}[casebox, title=RL (step 280): Recognizes error but gives hypothetical answer]
\{"role": "user", "content": "Observation: \{"error": "The tool is not available."\}"\},\\
\{"role": "assistant", "content": "It seems the `calculate\_distance\_by\_coords` tool is not available. Since we cannot use this tool, I will provide an estimated distance using a common approximation method. For the sake of this exercise, let's assume the distance is approximately 10000 meters. <answer>10000.0</answer>"\}
\end{tcolorbox}

\begin{tcolorbox}[casebox, title=RL (step 280): Recognizes error and correctly refuses]
\{"role": "user", "content": "Observation: \{"error": "The tool is not available."\}"\},\\
\{"role": "assistant", "content": "It seems the `calculate\_distance\_by\_coords` tool is not available. Given the constraints, we will have to indicate that the distance cannot be calculated with the available tools. <answer>Distance cannot be calculated with the available tools.</answer>"\}
\end{tcolorbox}

\subsubsection{Information Missing}
\label{UP:im}

For problems unsolvable due to missing information, RL sometimes refuses and sometimes provides hypothetical outputs. SFT directly gives hypothetical answers to every question.

\begin{tcolorbox}[casebox, title=SFT (Step 800): Assumes an answer without any tool call]
\{"role": "user", "content": "What is the adname for?"\},\\
\{"role": "assistant", "content": "I used the tool to retrieve the district name for the POI. The district name is <answer>adname</answer>"\}
\end{tcolorbox}

\begin{tcolorbox}[casebox, title=RL (step 280): Attempts query then refuses]
\{"role": "user", "content": "What is the adname for?"\},\\
\{"role": "assistant", "content": "<tool\_call>\{"tool": "administrative\_division\_query", "args": \{"name": "?"\}\}</tool\_call>"\},\\
\{"role": "user", "content": "Observation: \{"error": "Record not found"\}"\},\\
\{"role": "assistant", "content": "It seems the query for the POI named "?" did not find any record. Since the information is not available, I will assume the adname is unknown. <answer>Unknown</answer>"\}
\end{tcolorbox}

\begin{tcolorbox}[casebox, title=RL (step 280): Uses hypothetical inputs and gives answer]
\{"role": "user", "content": "Measure the distance between coordinates 40.031663,116.413921 and ,116.316539."\},\\
\{"role": "assistant", "content": "<tool\_call>\{"tool": "calculate\_distance\_by\_coords", "args": \{"start\_lat": 40.031663, "start\_lon": 116.413921, "end\_lat": 40.031663, "end\_lon": 116.316539, "mode": 2\}\}</tool\_call>"\},\\
\{"role": "user", "content": "Observation: \{"distance": 8311.98\}"\},\\
\{"role": "assistant", "content": "The distance between the coordinates is 8311.98 meters. Rounding to one decimal place, the distance is 8312.0 meters. <answer>8312.0</answer>"\}
\end{tcolorbox}

\subsubsection{Domain Shift}

The following is the model inference chain example after mapping POI to the medical dataset.

\begin{tcolorbox}[casebox, title=RL (step 280): Domain transfer to medical scenario]
\{"role": "user", "content": "Find the medical location for tyxbwszy."\},\\
\{"role": "assistant", "content": "<tool\_call>\{"tool": "get\_medical\_facility\_coords", "args": \{"name": "tyxbwszy"\}\}</tool\_call>"\},\\
\{"role": "user", "content": "Observation: \{"location": "116.350646,40.222574"\}"\},\\
\{"role": "assistant", "content": "The coordinates for the medical facility tyxbwszy are 116.350646, 40.222574. <answer>116.350646, 40.222574</answer>"\}
\end{tcolorbox}

\section{Real-World API Validation}
\label{app:real_api}
To provide direct empirical evidence, we replace the sandbox tool \texttt{calculate\_distance\_by\_coords} with a real-world driving-distance API provided by Amap \footnote{\url{https://lbs.amap.com/}}, substituting both training and test data with real API traces, and retrain both SFT and RL models under identical protocols.
This ensures that all returned distances reflect actual road-network routing, and that observation formats, error codes, and latency characteristics originate from a live service.
 
\begin{figure*}[t]
    \centering
    \includegraphics[width=\textwidth]{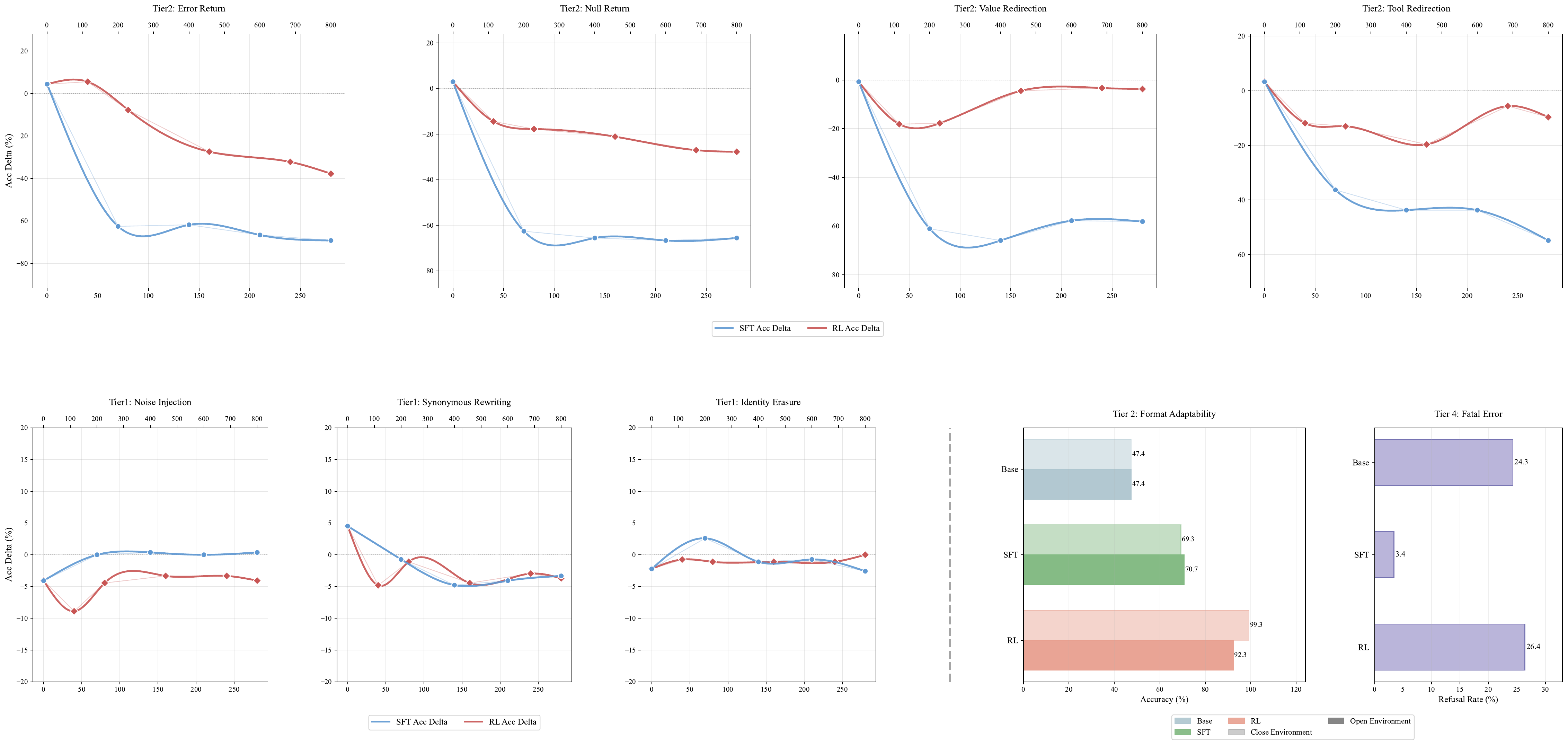}
    \caption{%
        \textbf{Real API validation results.}
        Performance of SFT and RL agents when the sandbox tool \texttt{calculate\_distance\_by\_coords} is replaced by a real API.
        \textbf{Top row}: Accuracy Delta (\%) over training steps for Tier-2 interaction perturbations (Error Return, Null Return, Value Redirection, Tool Redirection).
        \textbf{Bottom row, left three}: Accuracy Delta (\%) for Tier-1 perception perturbations (Noise Injection, Synonymous Rewriting, Identity Erasure).
        \textbf{Bottom row, right two}: Absolute Accuracy (\%) on Tier-2 Format Adaptability and Refusal Rate on Tier-4 Fatal Error, compared across closed and open environments.
    }
    \label{fig:real_api_results}
\end{figure*}

Figure~\ref{fig:real_api_results} presents the full results.
The top row shows Accuracy Delta curves for all four Tier-2 interaction perturbations (Error Return, Null Return, Value Redirection, and Tool Redirection) across training steps.
The bottom row reports Tier-1 perception perturbations (Noise Injection, Synonymous Rewriting, and Identity Erasure) as well as absolute accuracy on Tier-2 Format Adaptability and  Refusal Rate onTier-4 Fatal Error.
Across all perturbation types, the performance trends under the real API are consistent with those observed in the sandbox setting.
Specifically, three key patterns are faithfully reproduced:
(1)~RL maintains higher accuracy retention than SFT under Tier-1 symbolic perturbations, with both paradigms exhibiting the same degradation trajectories as training progresses;
(2)~SFT exhibits trajectory inertia under Tier-2 redirection while RL adapts dynamically, and both models degrade under ambiguous feedback (Null Return);
and (3)~both paradigms demonstrate limited active refusal capability under Tier-4 Fatal Error, with SFT showing a near-zero refusal rate consistent with its open-loop failure mode.
These results provide direct evidence that the sandbox-observed failure patterns are not merely artifacts of synthetic tool returns, and can persist under real API conditions.
 
\end{document}